\documentclass[opendfm]{xlance/xlance}

\usepackage{colortbl}                   
\definecolor{cellbest}{HTML}{DCC7EC}    
\definecolor{cellsecond}{HTML}{F4ECFA}  

\lstdefinestyle{llmoutput}{
    basicstyle=\ttfamily\footnotesize,
    breaklines=true,
    breakatwhitespace=true,
    columns=fullflexible,
    keepspaces=true,
    showstringspaces=false,
    frame=none
}

\newtcblisting{dsoutput}[2][]{
    breakable,
    enhanced,
    colback=gray!5,
    colframe=gray!30,
    boxrule=0.3pt,
    fonttitle=\color{black},
    arc=1mm,
    title=#2,
    listing only,
    listing options={
        style=llmoutput,
    },
    #1
}

\title{DeepSurvey: Agent-Oriented Automated Survey Generation with Analytical Depth and Citation Reliability}

\small{
\author[1]{\equalcontribution{Ziyue Yang}}
\author[1]{\equalcontribution{Da Ma}}
\author[1,4]{Hanqi Li}
\author[1]{Zijian Wang}
\author[1]{Tiancheng Huang}
\author[1]{Zijian Hu}
\author[1,2]{Chenrun Wang}
\author[1]{Yunzhe Zhang}
\author[1]{Xiaobao Wu}
\author[1,3,4]{Kai Yu}
\author[1,2,3,4]{\correspondence{chenlusz@sjtu.edu.cn}{Lu Chen}}
}

\small{
\affiliation[1]{X-LANCE Lab, School of Computer Science, Shanghai Jiao Tong University, Shanghai, China}
\affiliation[2]{Shanghai Innovation Institution, Shanghai, China}
\affiliation[3]{Jiangsu Key Lab of Language Computing, Suzhou, China}
\affiliation[4]{Suzhou Laboratory, Suzhou, China}
}

\abstract{
As scientific literature grows rapidly and research increasingly involves AI agents, automated survey generation has become a key capability for both agents and human researchers. For such agents, a survey serves as a primary knowledge source of a field prior to research. Since pretrained models already encode broad knowledge of established work, these consumers benefit more from analytical depth than from breadth alone; moreover, unsupported claims, once ingested as knowledge, can propagate into downstream research. However, existing systems tend to overemphasize coverage and presentation; they suffer from limited analytical depth due to reliance on abstracts and isolated paper processing, and from unreliable citations due to imprecise retrieval and post-hoc grounding. We present DeepSurvey, an agentic generation system that addresses both limitations. To enhance depth, DeepSurvey extracts structured keynotes, models cross-paper relationships through clustering and comparative analysis, and integrates a code-agent subsystem to recover implementation-level details. To fortify reliability, it combines citation-graph expansion with hybrid filtering for topic-focused retrieval, enforces evidence-constrained analysis and writing, and deploys multi-granularity agentic refinement to validate citation--claim alignment. Experiments show that DeepSurvey achieves the highest content score (8.34/10) and citation quality (recall and precision gains of 25.3\% and 35.2\% over the strongest baseline), generalizes more robustly across domains (a CS-to-non-CS score drop of 0.16 vs.\ 0.28--0.46 for prior systems), and is preferred by domain experts over human-written surveys (83.3\% in overall quality, 100\% in content depth). Moreover, when a coding agent uses a survey as its only literature source, the agent equipped with DeepSurvey achieves the best performance among human-written and baseline-generated surveys.
}

\metadatainfo{
    \small\makebox[\linewidth][c]{
        \href{https://drunk-mooon.github.io/}{\web~\textbf{Homepage}}
    }
}

\begin{document}
\publishdate{\today}
\maketitle


\section{Introduction}\label{sec:intro}
As scientific literature continues to grow rapidly, AI scientists are increasingly expected to autonomously survey a field, identify open questions, and design experiments~\citep{Lu2024TheAS, Yamada2025TheAS}, and recent work has demonstrated the feasibility of end-to-end automated survey generation~\citep{Wang2024AutoSurveyLL}. At the same time, the readership of surveys is itself broadening: LLM-based agents increasingly rely on surveys as a primary knowledge source. For these consumers, pretrained models already command broad knowledge of established work, so breadth alone offers relatively marginal value, while analytical depth and reliable attribution matter most. Depth lies in finer-granularity analysis, in implementation-level knowledge that is often omitted or heavily condensed in paper text and recoverable from code. Such depth also lies in connections among papers, from pairwise contrasts to the structure of the whole field. Reliability lies in the correspondence between claims and the evidence they cite. A survey without depth probably only restates what agents already know, whereas a survey without reliability is likely harmful to downstream agents.



\begin{figure}[!htp]
  \centering
  \includegraphics[width=1.0\linewidth]{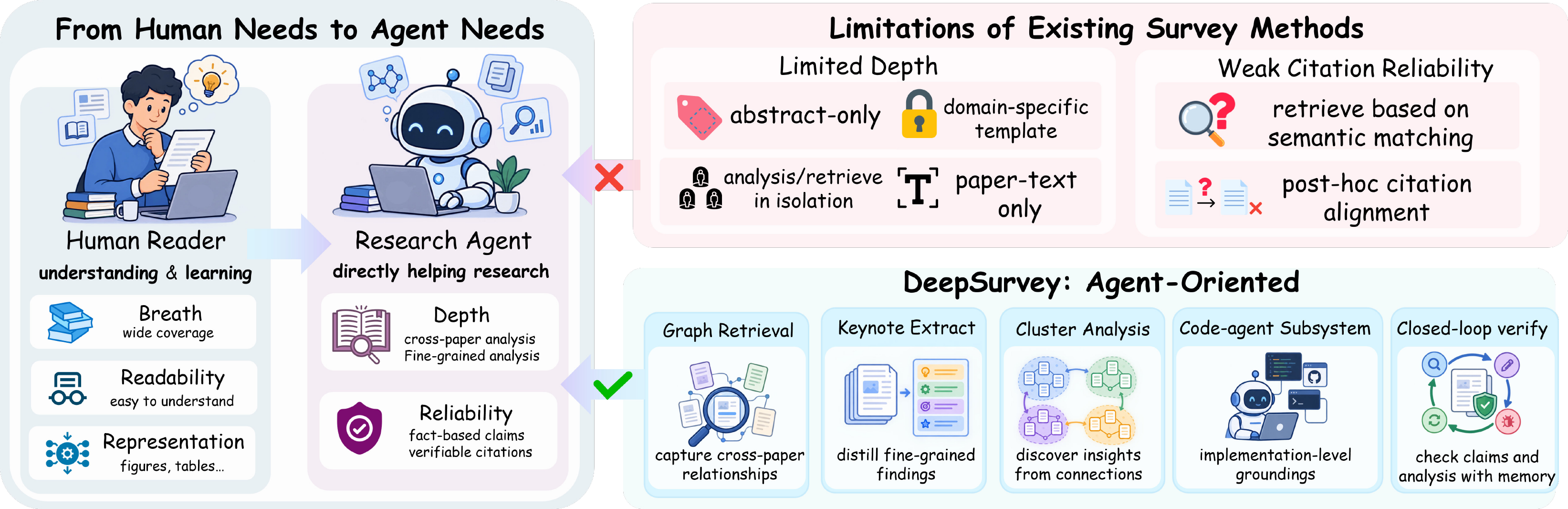}
  \caption{Survey consumers from human to agent. Research agents require analytical depth and verifiable reliability more than breadth and presentation. DeepSurvey addresses the corresponding limitations of existing methods in depth and citation reliability through graph-based retrieval, keynote extraction, cluster analysis, a code-agent subsystem, and closed-loop verification.}
  \label{fig:limitations}
\end{figure}

However, existing automatic survey generation systems tend to overemphasize coverage and presentation~\citep{Wang2024AutoSurveyLL}, and suffer from three fundamental limitations~(Figure~\ref{fig:limitations}).
\emph{1) Limited survey depth:} Most methods rely solely on abstracts~\citep{Wang2024AutoSurveyLL, Yan2025SurveyForgeOT} or domain-specific templates~\citep{Liang2025SurveyXAS} that fail to generalize across diverse fields. Furthermore, they typically process papers in isolation without fully modeling cross-paper relationships~\citep{Go2025LiRAAM,Chao2025LLMMapReduceV3EI}. Crucially, by overlooking non-textual artifacts such as code repositories~\citep{Nguye2025SurveyGAM}, these systems fail to capture key technological details and implementation-level insights that are often omitted or heavily condensed in the main text.
\emph{2) Weak citation reliability:} On the retrieval front, dense semantic matching often introduces tangentially related papers that dilute the survey content~\citep{Wu2025AutoSurvey2ER}. On the grounding front, citations are typically attached via post-hoc alignment rather than derived directly from cited evidence, leaving substantial claims without verifiable support~\citep{Wang2024AutoSurveyLL}.
\emph{3) Unconvincing evaluation.}
Existing evaluations are narrowly restricted to computer science~\citep{Shi2025SciSageAM, Su2025SurGEAB, Nguye2025SurveyGAM}, adopt outdated metrics~\citep{bao-etal-2025-surveygen} or metrics with weak discriminative power~\citep{Wang2024AutoSurveyLL}, and compare against only a handful of baselines~\citep{Wu2025AutoSurvey2ER, Zhou2024LLMMapReduceSL, Liu2025AgenticAL}, making it difficult to reliably assess and differentiate methods.

To address these limitations, we propose {DeepSurvey}, an agent-oriented system optimized along three core axes:
\emph{1) Enhancing survey depth:} To achieve scalable, domain-agnostic analysis, \textbf{DeepSurvey} extracts structured, informative keynotes from full-text literature~(\S\ref{sec:understanding}). Moving beyond isolated paper processing, we explicitly model cross-paper relationships by clustering the literature and conducting multi-level comparative analysis (e.g., via structural relation graphs and joint question-answering reasoning). Furthermore, we integrate a dedicated code-agent subsystem to cross-reference official repositories; this allows DeepSurvey to retrieve and complement critical technological dimensions that are missing from pure text, ensuring a more comprehensive and holistic literature synthesis~(\S\ref{sec:analysis}).
\emph{2) Fortifying citation reliability:} We combine citation-graph expansion with LLM-based re-ranking to maintain strict topical consistency and filter out irrelevant papers~(\S\ref{sec:retrieval}). Instead of relying on vulnerable post-hoc alignment, we enforce strict source attribution throughout the entire generation lifecycle~(\S\ref{sec:analysis}): directly anchoring citations during multi-stage outline planning~(\S\ref{sec:writing}), and deploying an agentic refinement loop to rigorously validate citation-claim pairs~(\S\ref{sec:refinement}).
\emph{3) Strengthening evaluation:} we construct a cross-domain evaluation protocol grounded in existing influential benchmarks, redesign the evaluation protocol with more discriminative and agent-oriented criteria, and include a broader set of competitive baselines to enable more rigorous and informative comparisons~(\S\ref{ssec:DSEval}).

Extensive experiments demonstrate that DeepSurvey achieves an overall content score of 8.341/10 with a depth score of 7.897/10, substantially surpassing prior methods, and reaches 0.630 recall and 0.569 precision in citation quality, outperforming the strongest baseline by 25.3\% and 35.2\%~(\S\ref{ssec:main_result}). DeepSurvey also generalizes more robustly across domains, exhibiting a much smaller performance drop in non-CS fields than existing methods (0.16 vs.\ 0.28--0.46)~(\S\ref{ssec:main_result}). Notably, in downstream validation, a coding agent working from a DeepSurvey survey autonomously outperforms the best published result on PINNacle (relative $L_2$ error $1.53\times10^{-2}$ vs.\ $3.81\times10^{-2}$), a result no other literature condition reaches~(\S\ref{ssec:downstream_agent}). In pairwise human evaluation, domain experts prefer DeepSurvey over human-written surveys, especially in overall quality (83.3\% win rate) and content depth (100\% win rate)~(\S\ref{ssec:human_evaluation}). Our evaluation protocol is also validated through meta-evaluation: the LLM judge achieves a low coefficient of variation of 0.244\%, and its Cohen's $\kappa$ with human experts (0.6463) approaches human--human agreement (0.6958)~(\S\ref{ssec:meta_eval}).


Overall, DeepSurvey provides an end-to-end system for generating in-depth and reliable surveys across domains, designed to support literature synthesis especially for AI scientists.

Our main contributions are as follows:





\begin{itemize}[leftmargin=*, nosep]
    \item We propose \textbf{DeepSurvey}, the first agent-oriented survey generation system that prioritizes the \emph{analytical depth} agents need over breadth, building a reusable analysis substrate via keynote extraction, cross-paper relation modeling, and \textbf{a code-agent subsystem}.

    \item We design closed-loop \emph{citation reliability} mechanisms across the entire generation pipeline, including graph-based retrieval, evidence-constrained writing, and multi-granularity agentic refinement.

    \item Extensive experiments show that DeepSurvey achieves the best content and citation quality across domains and human expert preference over human-written surveys. A coding agent working from a DeepSurvey survey also outperforms the best published result on PINNacle.
\end{itemize}




\section{DeepSurvey}

\begin{figure*}[t]
  \centering
  \includegraphics[width=1.0\textwidth]{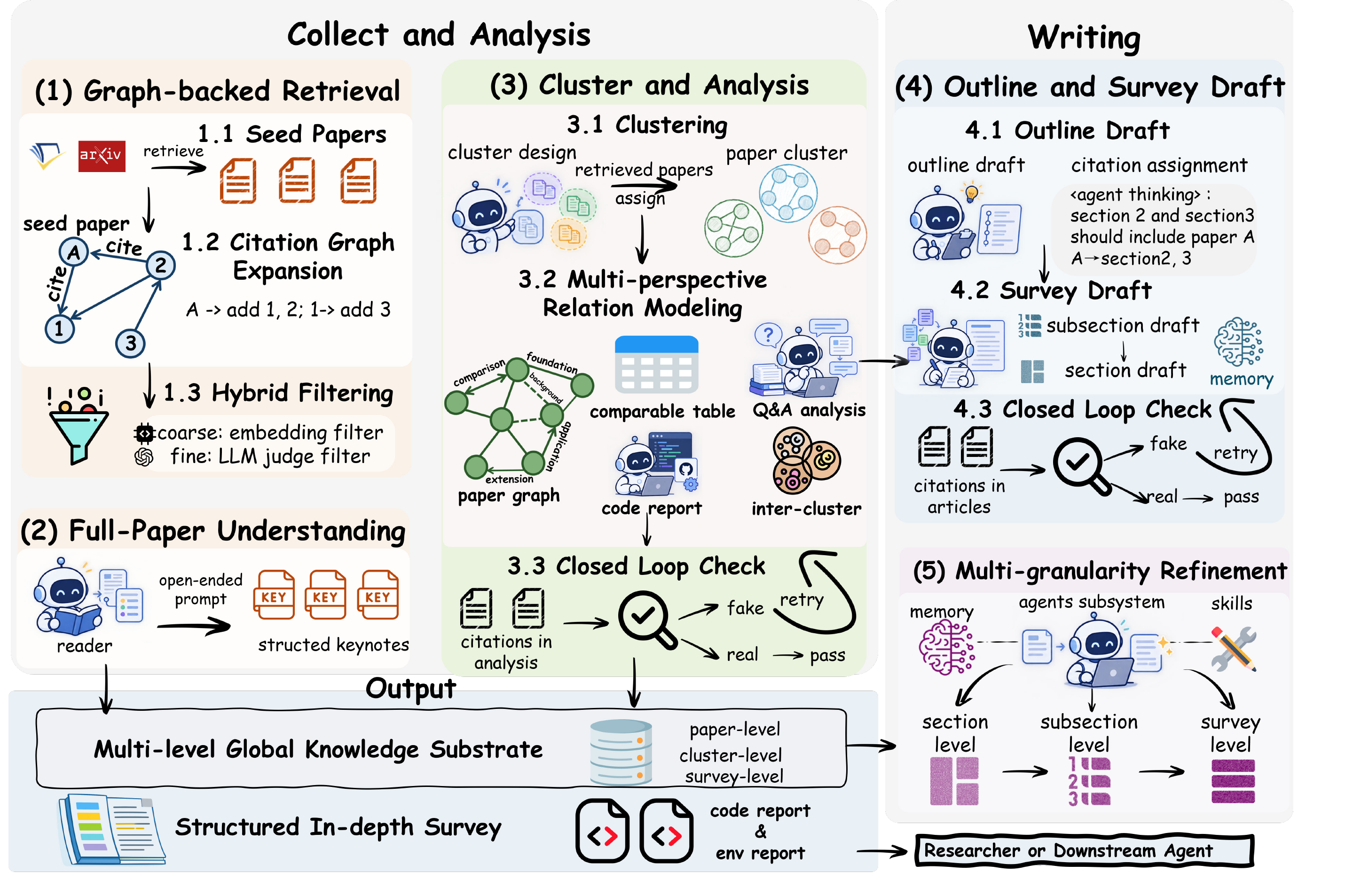}
  \caption{Overview of DeepSurvey. Stages 1--3 build the analysis substrate via graph-based retrieval, keynote extraction, clustering with multi-perspective relation modeling and code-agent subsystem. Stage 4 generates the survey. Stage 5 applies multi-granularity agentic refinement.}
  \label{fig:DeepSurvey}
\end{figure*}

In this section, we propose DeepSurvey, an agentic system that produces an agent-oriented survey together with a structured, reusable analysis knowledge substrate (Figure~\ref{fig:DeepSurvey}): Stages~1--3 build the analysis substrate through graph-based retrieval, full-text keynotes, code-agent analyses and clustering with multi-perspective relation analysis; Stage~4 generates the survey from it with outline-driven planning and citation assignment; Stage~5 applies multi-granularity agentic refinement. Engineering mechanisms for robustness are in Appendix~\ref{app:engineering}.

\subsection{Graph-based Literature Retrieval}
\label{sec:retrieval}

Given a research topic, DeepSurvey retrieves seed papers from Semantic Scholar~\citep{Wade2022TheSS,lo-etal-2020-s2orc,2017Explicit}, providing broader coverage than static local databases. A judge agent filters initial candidates by topical alignment to reduce seed-stage drift. Starting from the seed set, DeepSurvey expands papers through the edges in citation graph in both directions, capturing inter-paper relations that flat semantic search misses. A two-stage filtering strategy controls expansion noise: a coarse semantic filter (encode abstracts and titles) preserves high-recall candidates, while a fine-grained LLM re-ranking removes topically irrelevant papers.


\subsection{Full-text Understanding}
\label{sec:understanding}

For each paper, DeepSurvey reads the full text and generates a structured \textbf{keynote} covering contributions, methods, experiments, assumptions, and limitations. Since key technological details and implementation-level insights are often omitted or heavily condensed in abstracts, full-text reading is essential for capturing the fine-granularity information needed for meaningful cross-paper comparison~\citep{cohan-goharian-2015-scientific}. The open-ended structure avoids fixed templates, enabling domain-adaptive representation that supports cross-domain summarization without sacrificing specificity.

\subsection{Clustering and Relation Analysis}
\label{sec:analysis}

After keynotes generation, DeepSurvey designs cluster themes based on the complete keynote collection and assigns papers to clusters through a clustering agent and a partitioning agent. Papers spanning multiple cluster topics may be assigned to multiple clusters. A verification-retry loop corrects missing or hallucinated assignments.

\begin{figure*}[!htp]
  \centering
  \includegraphics[width=1.0\textwidth]{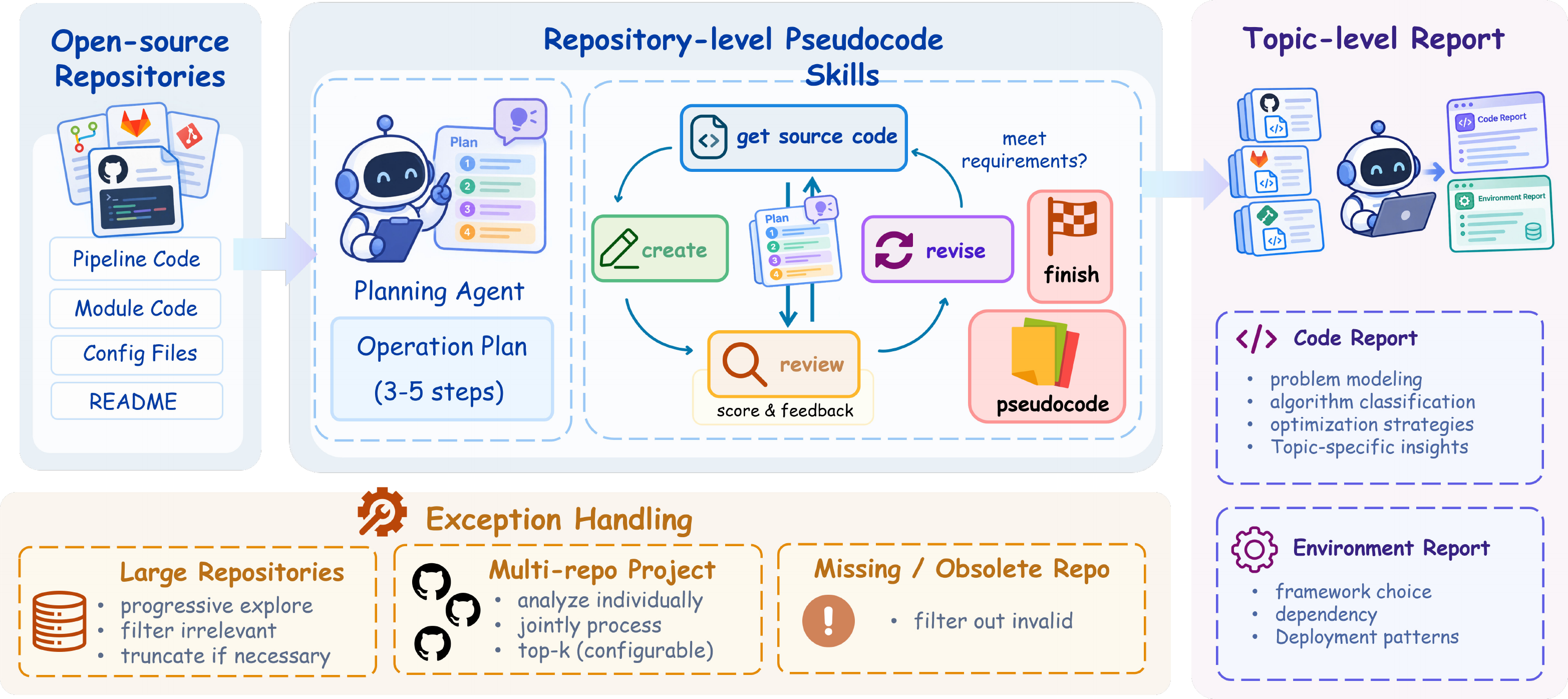}
  \caption{The code-agent subsystem. A centralized planning agent orchestrates specialized skills to iteratively explore open-source repositories and distill repository-level pseudocode with exception handling. The pseudocode is synthesized into topic-level reports.}
  \label{fig:codeagent}
\end{figure*}

Within each cluster, three complementary perspectives are used to model cross-paper analysis: \textbf{relation graphs} model typed citation relations (e.g., foundation, extension, substitution) for technical lineage, \textbf{comparison tables} compare papers along key dimensions, and \textbf{guided Q\&A} extracts cross-paper patterns. \textbf{Inter-cluster} analysis is also performed to capture cross-cluster differences. A \textbf{code-agent subsystem} (Figure~\ref{fig:codeagent}) explores the associated repositories to recover implementation-level details overlooked by purely textual analysis: a planning agent orchestrates specialized skills to distill repository pseudocode, which is synthesized into repository-level reports and further integrated into a topic-level report; these serve as implementation-level evidence for the writing and refinement agents, indexed in the knowledge substrate (details in Appendix~\ref{app:code_analysis}, cases in Appendix~\ref{app:code_analysis_case}).


Analyses require explicit source paper attribution, with monitoring to detect and correct hallucinated sources. All outputs are stored in the global knowledge substrate (details in Appendix~\ref{app:knowledge_base}). Cases are in Appendix~\ref{app:code_analysis_case} and~\ref{app:cluster_analysis_case}.

\subsection{Outline Draft and Scoped Writing}
\label{sec:writing}

DeepSurvey first generates a hierarchical outline from the clustering results with structured evidence in the analysis knowledge substrate, and iteratively refines it against retrieved paper keynotes. The outline specifies the content scope and constraints of each section and subsection.

Following outline generation, papers are \emph{explicitly assigned} to appropriate sections based on keynotes and cluster analyses in knowledge substrate. This citation-anchoring process creates a localized evidence constraint: each writing unit depends only on its assigned paper subset. This mechanism can improve the model's overall performance and reduce hallucination risk by limiting the context~\citep{Liu2023LostIT, Du2025ContextLA, Liu2025TowardsLC}. 

Writing proceeds bottom-up. \textbf{Subsection drafts} receive the outline, subsection description, relevant cluster-based analyses, and keynotes for the assigned papers from knowledge substrate. \textbf{Section drafts} aggregate subsection content with overview-level analysis to form higher-level narratives.

After each paragraph is generated, a \textbf{citation verification} mechanism checks for consistency and authenticity against the assigned paper set. Detected citation errors trigger retry generation with error memory, ensuring strict alignment between the generated text and the evidence collection. Compared with previous methods that simply hard-match hallucinated citations against a database, our approach better aligns claims with citations.

\subsection{Multi-granularity Refinement}\label{sec:refinement}

After initial draft generation, DeepSurvey applies multi-granularity refinement at section, subsection, and survey levels to improve content quality and validate citation--claim 
pairs. A \textbf{centralized planning agent} inspects the draft, outline, revision history, and relevant evidence, then coordinates specialized skills: readers retrieve keynotes and source passages; the reviewer assesses quality and provides suggestions; and the reviser applies edits. The planning agent iteratively invokes these skills until the planner considers the draft satisfactory or a maximum round count is reached. 

Two mechanisms improve refinement quality. First, \textbf{global memory} preserves reviewer feedback, score changes, and revision states across rounds. Second, \textbf{multi-granularity optimization} performs section-level refinement for organization and logic, subsection-level refinement for local analysis, readability and citation--claim validation, and survey-level refinement for coherence. We also include a simplified skill-loop fallback for weaker backbone models.

\section{Experiments}
Reliable evaluation of survey generation requires benchmarks that are both diverse and discriminative. However, existing protocols remain limited: they are either restricted to computer science~\citep{Yan2025SurveyForgeOT} or lack sufficient discriminative power (Table~\ref{tab:coverage_structure_relevance_main_tex}, details in Appendix~\ref{app:eval_autosurvey}), they often compare against only a narrow set of baselines~\citep{Liang2025SurveyXAS}, and they rarely assess whether a survey can directly support downstream agents in conducting research. These limitations motivate a more rigorous, multi-dimensional, cross-domain and agent-oriented evaluation setup.

\subsection{Experiment Setup}\label{ssec:DSEval}
To address these limitations, we construct a multi-domain evaluation setup by combining the complementary strengths of prior benchmark efforts. For topic selection, we draw a high-quality subset from SurveyBench~\citep{Yan2025SurveyForgeOT} and AutoSurvey~\citep{Wang2024AutoSurveyLL} in computer science, focusing on well-defined topics in LLMs and machine learning. Moreover, as automated research expands to a growing range of disciplines, research agents will increasingly need surveys well beyond computer science; we therefore further select a subset of SurveyGen~\citep{bao-etal-2025-surveygen} covering high-interest topics in Biology, Medicine, Psychology, etc. (details in Appendix~\ref{app:eval_introduction_input}). Note that we use the MiMo-series model (MiMo-V2.5)~\footnote{https://mimo.xiaomi.com/mimo-v2-5} as the evaluation model (reasons in Appendix~\ref{app:eval_stability}); DeepSurvey remains the top method under judges from other model families as well, confirming that the results are not an artifact of judge--generator family affinity (Appendix~\ref{app:other_model_result}).



\subsubsection{Evaluation Metrics}\label{ssec:evaluation_metric}

We evaluate generated surveys along two dimensions: content quality measures whether generated surveys achieve sufficient depth and research value, while citation quality assesses the reliability and accuracy of references (detailed metrics are presented in Appendix~\ref{app:eval_introduction_standard}).

\textbf{Content Quality} is assessed using an LLM rubric judge on a 1--10 scale. Building on AutoSurvey's influential framework~\citep{Wang2024AutoSurveyLL} and Agentic AutoSurvey's discriminative standard~\citep{Liu2025AgenticAL}, we further adapt the protocol to the needs of agent consumers (detailed analysis in Appendix~\ref{app:detailed_standard}). We evaluate three aggregated dimensions: \textbf{Core Quality} (synthesis, organization, comprehensiveness, relevance), \textbf{Writing Quality} (readability, academic rigor, clarity and coherence), and \textbf{Content Depth} (critical analysis, novelty and insights, specificity, future directions). The final score is computed as 0.4 $\times$ Core + 0.2 $\times$ Writing + 0.4 $\times$ Depth. Further explanation on the formula and the dimensions is in Appendix~\ref{app:eval_introduction_standard}.



\textbf{Citation Quality} follows AutoSurvey~\citep{Wang2024AutoSurveyLL} and is measured by Citation Recall (whether claims are supported by their citations) and Citation Precision (whether cited papers are truly relevant to the claims); the NLI judgments are rendered by the same judge model (MiMo-V2.5). Details are presented in Appendix~\ref{app:eval_introduction_standard}.




\subsubsection{Baselines}

We compare DeepSurvey against four representative automated survey generation systems spanning diverse methodological approaches. \textbf{AutoSurvey}~\citep{Wang2024AutoSurveyLL} pioneered the paradigm of semantic retrieval with parallel generation. \textbf{SurveyX}~\citep{Liang2025SurveyXAS} leverages AttributeTree for template-based literature understanding. \textbf{SurveyForge}~\citep{Yan2025SurveyForgeOT} introduces SANA for high-quality retrieval. \textbf{LiRA}~\citep{Go2025LiRAAM} employs a multi-agent workflow with specialized roles\footnote{LiRA does not include a retrieval module so we use AutoSurvey's retrieval module as a substitute.}. All baselines use the same generation backbone as DeepSurvey (MiMo-V2.5). Complete parameter settings are presented in Appendix~\ref{app:eval_eval_params} and the specific topics for evaluation are presented in Appendix~\ref{app:eval_introduction_input}.



\subsection{Main Results}\label{ssec:main_result}

\begin{table*}[t]
\centering
\scriptsize
\setlength{\tabcolsep}{1.5pt}
\renewcommand{\arraystretch}{1.15}
\caption{Performance comparison of different survey generation methods on overall, fine-grained content metrics, and citation quality (precision and recall). The best and second-best results are highlighted in \colorbox{cellbest}{purple} and \colorbox{cellsecond}{light purple}, respectively.}
\label{tab:all_results}
\begin{tabular}{l c ccccc cccc ccccc cc}
\toprule
\multirow{3}{*}{Method}
& \multirow{3}{*}{Total Score}
& \multicolumn{14}{c}{Content Quality}
& \multicolumn{2}{c}{\multirow{2}{*}{Citation Quality}} \\
\cmidrule(lr){3-16}
&
& \multicolumn{5}{c}{Core Quality}
& \multicolumn{4}{c}{Writing Quality}
& \multicolumn{5}{c}{Content Depth}
&
& \\
\cmidrule(lr){3-7}
\cmidrule(lr){8-11}
\cmidrule(lr){12-16}
\cmidrule(lr){17-18}
&
& Total
& Synth.
& Org.
& Comp.
& Rel.
& Total
& Read.
& Rigor
& Clarity
& Total
& Crit.
& Novel.
& Spec.
& Future
& Precision
& Recall \\
\midrule

AutoSurvey
& 7.958
& \cellcolor{cellsecond}8.646
& \cellcolor{cellsecond}7.983
& 8.608
& \cellcolor{cellsecond}8.718
& 9.276
& \cellcolor{cellsecond}8.153
& 7.606
& \cellcolor{cellsecond}8.500
& \cellcolor{cellsecond}8.353
& 7.171
& 7.049
& 6.847
& \cellcolor{cellsecond}7.991
& 6.798
& 0.342
& 0.440 \\

SurveyX
& 6.993
& 8.187
& 7.283
& 8.133
& 8.000
& \cellcolor{cellbest}\textbf{9.333}
& 7.067
& 7.053
& 6.313
& 7.833
& 5.763
& 5.733
& 5.133
& 5.983
& 6.200
& 0.125
& 0.194 \\

SurveyForge
& \cellcolor{cellsecond}8.071
& 8.616
& 7.975
& \cellcolor{cellsecond}8.700
& 8.690
& 9.100
& \cellcolor{cellsecond}8.153
& \cellcolor{cellbest}\textbf{7.793}
& 8.283
& \cellcolor{cellbest}\textbf{8.382}
& \cellcolor{cellsecond}7.485
& \cellcolor{cellsecond}7.350
& \cellcolor{cellsecond}7.290
& 7.895
& \cellcolor{cellsecond}7.405
& \cellcolor{cellsecond}0.421
& \cellcolor{cellsecond}0.503 \\

LiRA
& 5.687
& 5.049
& 7.850
& 8.425
& 2.565
& 1.355
& 6.805
& 7.695
& 5.155
& 7.565
& 5.766
& 5.775
& 5.045
& 6.095
& 6.150
& 0.227
& 0.442 \\

DeepSurvey
& \cellcolor{cellbest}\textbf{8.341}
& \cellcolor{cellbest}\textbf{8.865}
& \cellcolor{cellbest}\textbf{8.334}
& \cellcolor{cellbest}\textbf{8.935}
& \cellcolor{cellbest}\textbf{8.889}
& \cellcolor{cellsecond}9.300
& \cellcolor{cellbest}\textbf{8.183}
& \cellcolor{cellsecond}7.711
& \cellcolor{cellbest}\textbf{8.506}
& 8.333
& \cellcolor{cellbest}\textbf{7.897}
& \cellcolor{cellbest}\textbf{7.389}
& \cellcolor{cellbest}\textbf{7.866}
& \cellcolor{cellbest}\textbf{8.194}
& \cellcolor{cellbest}\textbf{8.139}
& \cellcolor{cellbest}\textbf{0.569}
& \cellcolor{cellbest}\textbf{0.630} \\

\bottomrule
\end{tabular}
\end{table*}




\textbf{Content and Citation Quality.} The results are shown in Table~\ref{tab:all_results}. For \textbf{content quality}, DeepSurvey achieves the highest overall content score (8.341), with Content Depth at 7.897, which is 0.412 above the second-highest baseline. For \textbf{citation quality}, DeepSurvey achieves the best scores, with 0.630 recall and 0.569 precision, outperforming the strongest baseline SurveyForge by 25.3\% and 35.2\%, respectively, highlighting the effect of the systematic reliability mechanism. Overall, the DeepSurvey system improves both survey depth and citation reliability. A \textbf{case study} versus the most influential baseline AutoSurvey (Figure~\ref{fig:case_study}) further confirms DeepSurvey’s superiority: it stays topic focused and provides human-expert-aligned insights, while AutoSurvey shows topic drift and superficial analysis (details in Appendix~\ref{app:AS_DS_compare_case}).

\textbf{Meta-Analysis.} All results are averaged over repeated runs. The evaluation proves stable: aggregated CVs are at most 6.3\% across methods, and fine-grained sub-dimensions mostly stay below 10\% (Appendix~\ref{app:eval_results_supp}). DeepSurvey itself is highly stable (aggregated CVs of 0.8\%--2.2\%), with its gains over all baselines reaching large effect sizes (Hedges' $g \geq 0.8$) on Total Score, Content Depth, and citation metrics.

\begin{figure}[!htp]
  \centering
  \includegraphics[width=0.7\linewidth]{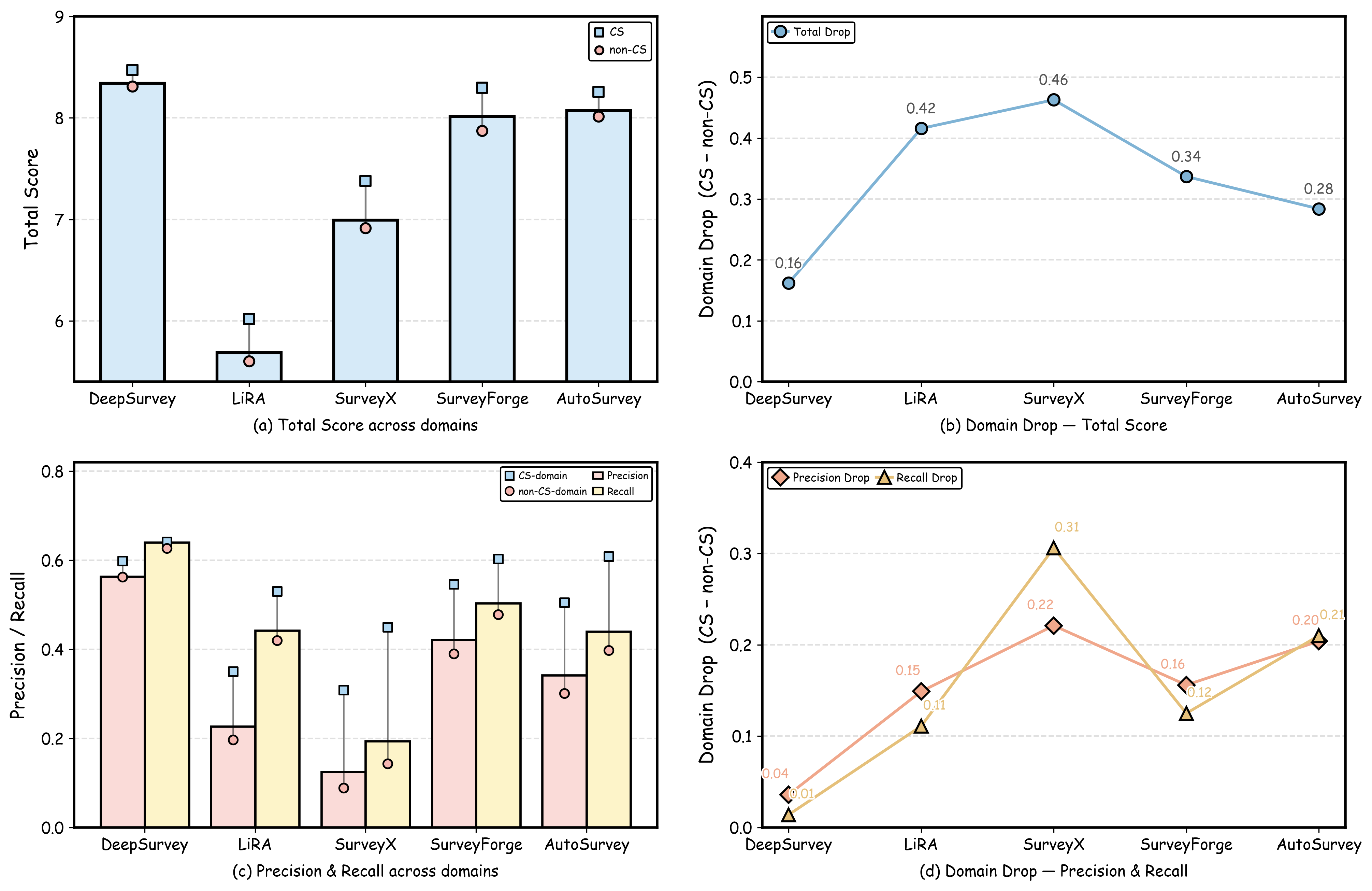}
  \caption{
    Cross-domain generalization across CS and non-CS domains.
    The left panels (a, c) compare overall, CS, and non-CS scores for each method; the right panels (b, d) show the absolute total-score gap between CS and non-CS domains.
    }
  \label{fig:cross-domain-differences}
\end{figure}

\textbf{Cross-domain Evaluation.} Figure~\ref{fig:cross-domain-differences} separates fields in Table~\ref{tab:all_results}: \textbf{DeepSurvey} is consistently the strongest method across CS and non-CS domain settings, surpassing \textbf{AutoSurvey} and \textbf{SurveyForge}, the two strongest prior baselines, and maintains a much smaller drop when shifting to the non-CS domain than all other methods (0.16 vs. 0.28--0.46), confirming substantially stronger domain generalization.

\subsection{Downstream Validation with a Coding Research Agent}\label{ssec:downstream_agent}
To examine whether a DeepSurvey survey improves the research capability of downstream agents, we task an identical coding agent (codex) with optimizing vanilla PINNs on PINNacle~\citep{hao2023pinnaclecomprehensivebenchmarkphysicsinformed}, whose published optimization methods are stripped, under four literature conditions (none, human-written survey, SurveyForge, and DeepSurvey) that serve as its only knowledge source, within a 480-minute budget; each condition is run three times, and the best run is reported (details in Appendix~\ref{app:downstream_agent}). The DeepSurvey-condition transcript shows its mechanism-first survey, which maps each method to the targeted failure mode and its applicability conditions, translated directly into design decisions: the agent measures a per-axis spectral fingerprint from the PDE operator, completes a hard-constraint output transform guided by the survey's analysis, and, finding no dedicated techniques for hyperbolic problems, autonomously invents a symmetric d'Alembert ansatz that yields the Wave-1D-C breakthrough. The delivered method attains an overall relative $L_2$ error of $1.53\times10^{-2}$ (Table~\ref{tab:pinnacle_downstream}), the best of all conditions: $2.5\times$ below the per-case-best published PINNacle reference and the highest novelty score judged by GPT-6 Astra, whereas all other conditions remain above the published reference. This contrast suggests that the implementation-level understanding conveyed by a DeepSurvey survey, rather than the enumeration of methods, is what transfers to downstream agents.

\subsection{Human Expert Evaluation}\label{ssec:human_evaluation}
\label{subsec:human_expert_evaluation}
To verify that DeepSurvey is also applicable to human researchers, we further conduct blinded pairwise evaluations where domain experts compare DeepSurvey against human-written surveys on the same topic set, without knowing each response’s origin (detailed settings in Appendix~\ref{app:human_exp}). Figure~\ref{fig:human_llm_comparison} summarizes the results: DeepSurvey is preferred in overall quality (winning rate 83.3\%), with the strongest advantage (winning rate 100\%) in Content Depth. Writing Quality is comparatively weaker: DeepSurvey wins 12 of the 24 annotations (50.0\%), and 12 of the 22 non-tie annotations (54.5\%) once the 2 ties are excluded, which is consistent with the automatic evaluation results.

\begin{table}[t]
  \centering
  \begin{minipage}[t]{0.42\textwidth}
    \centering
    \captionsetup{type=figure}
    \includegraphics[width=\linewidth]{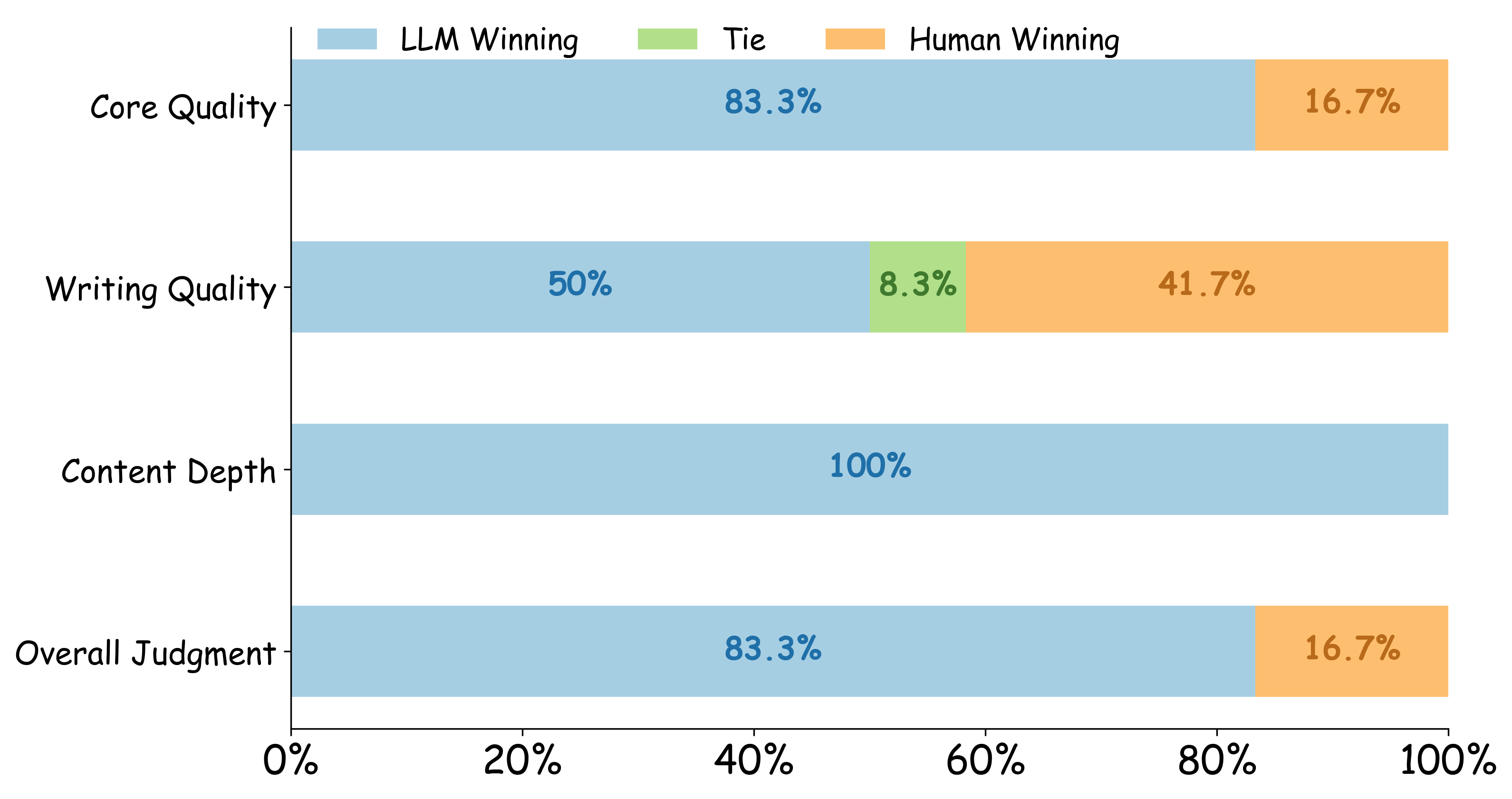}
    \caption{Comparison between DeepSurvey generated and human written surveys on the same topic set judged by human experts.
}
    \label{fig:human_llm_comparison}
  \end{minipage}
  \hfill
  \begin{minipage}[t]{0.54\textwidth}
    \centering
    \captionsetup{type=table}
    \scriptsize
    \setlength{\tabcolsep}{2pt}
    \renewcommand{\arraystretch}{1.1}
    \resizebox{\linewidth}{!}{%
      \begin{tabular}{lrrrrrcc}
        \toprule
        \multirow{2}{*}{Condition} & Burgers- & Heat- & Heat- & Wave- & \multirow{2}{*}{Overall} & \multirow{2}{*}{SOTA} & \multirow{2}{*}{Novelty} \\
        & 1D-C & 2D-CG & 2D-MS & 1D-C & & & \\
        \midrule
        PINN-base (pilot) & 2.10e-02 & 2.58e-02 & 1.53e-01 & 4.86e-01 & 1.71e-01 & -- & -- \\
        PINNacle SOTA & 1.33e-02 & 2.39e-02 & 1.73e-02 & 9.79e-02 & 3.81e-02 & -- & -- \\
        \midrule
        none & \cellcolor{cellbest}\textbf{1.33e-02} & \cellcolor{cellbest}\textbf{1.78e-02} & \cellcolor{cellsecond}1.85e-02 & 4.68e-01 & 1.29e-01 & 1 & 5 \\
        human & 2.20e-02 & \cellcolor{cellsecond}2.59e-02 & 4.80e-02 & \cellcolor{cellsecond}1.63e-01 & \cellcolor{cellsecond}6.48e-02 & 0 & 4 \\
        surveyforge & 1.88e-02 & 2.60e-02 & 9.73e-02 & 5.10e-01 & 1.63e-01 & 0 & 5 \\
        deepsurvey & \cellcolor{cellsecond}1.36e-02 & 2.64e-02 & \cellcolor{cellbest}\textbf{1.18e-02} & \cellcolor{cellbest}\textbf{9.54e-03} & \cellcolor{cellbest}\textbf{1.53e-02} & 2 & 7 \\
        \bottomrule
      \end{tabular}}
    \caption{Downstream validation on PINNacle: relative $L_2$ error (lower is better). ``SOTA'' indicates number of results that surpass the best published PINNacle result. ``Novelty'' is scored by GPT-6 Astra (details in Appendix~\ref{app:downstream_agent}).}
    \label{tab:pinnacle_downstream}
  \end{minipage}
\end{table}

\subsection{Ablation Studies}\label{ssec:ablation}

\begin{table}[t]
  \centering
  \begin{minipage}[t]{0.48\textwidth}
    \centering
    \captionsetup{type=table}          
    \scriptsize
    \setlength{\tabcolsep}{2pt}
    \begin{tabular}{lcccc}
      \toprule
      Method & Total & Core & Writing & Depth \\
      \midrule
      w/o collector & 6.031 & 5.082 & 5.889 & 7.052 \\
      w/o keynote   & 7.957 & 8.634 & 8.081 & 7.218 \\
      w/o analyzer  & 7.933 & 8.514 & 7.943 & 7.346 \\
      w/o writer    & 7.154 & 7.983 & 7.033 & 6.384 \\
      w/o refinement & 7.885 & 8.498 & 7.704 & 7.362 \\
      w/o agentic      & 8.153 & 8.752 & 8.016 & 7.622 \\
      w/o code & \cellcolor{cellsecond}8.217 & \cellcolor{cellsecond}8.833 & \cellcolor{cellsecond}8.110 & \cellcolor{cellsecond}7.655 \\
      DeepSurvey     & \cellcolor{cellbest}\textbf{8.341} & \cellcolor{cellbest}\textbf{8.865} & \cellcolor{cellbest}\textbf{8.183} & \cellcolor{cellbest}\textbf{7.897} \\
      \bottomrule
    \end{tabular}
    \caption{Ablation study of content quality when removing each core module (denoted as ``w/o'').}
    \label{tab:ablation_results}
  \end{minipage}
  \hfill
  \begin{minipage}[t]{0.48\textwidth}
    \centering
    \captionsetup{type=table}
    \scriptsize
    \setlength{\tabcolsep}{2pt}
    \begin{tabular}{lccc}
      \toprule
      Method & Recall & Precision & Valid Ratio \\
      \midrule
      w/o collector & 0.447 & 0.406 & 0.970 \\
      w/o keynote   & 0.600 & 0.529 & 0.989 \\
      w/o analyzer  & \cellcolor{cellbest}\textbf{0.636} & \cellcolor{cellbest}\textbf{0.574} & \cellcolor{cellsecond}0.998 \\
      w/o writer    & 0.593 & 0.509 & 0.610 \\
      w/o refinement & 0.613 & 0.537 & 0.994 \\
      w/o agentic    & 0.591 & 0.511 & \cellcolor{cellbest}\textbf{1.0} \\
      w/o code & 0.617 & 0.567 & 0.995 \\
      DeepSurvey     & \cellcolor{cellsecond}0.630 & \cellcolor{cellsecond}0.569 & 0.996 \\
      \bottomrule
    \end{tabular}
    \caption{Ablation study of citation quality when removing each core module (denoted as ``w/o'').}
    \label{tab:retrieval_precision_validratio}
  \end{minipage}
\end{table}

We evaluate the contribution of each core module by removing it from the full system:
\textbf{w/o collector} replaces graph-based retrieval with static database lookup;
\textbf{w/o keynote} uses only abstracts instead of full-text keynotes;
\textbf{w/o analyzer} skips clustering and relation modeling;
\textbf{w/o writer} generates directly from retrieval results without outline-driven writing or citation checks;
\textbf{w/o refinement} disables iterative agentic optimization.
\textbf{w/o agentic} replaces the central-planner agentic refinement system with a skill-loop (iteratively calling review and revise).
\textbf{w/o code} skips code-agent subsystem's analysis.
In addition, we add Valid Citation Ratio to measure citation authenticity and formatting correctness: citations that cannot be parsed by paper APIs are treated as false references\footnote{Due to the timeliness of the content in the paper's API and issues with API response, the value may contain errors (within 1\%).}.

Tables~\ref{tab:ablation_results} and~\ref{tab:retrieval_precision_validratio} show that DeepSurvey's gains come from complementary components rather than any single module. The \textbf{collector} is the most critical: its removal causes the largest degradation (total: 8.341 to 6.031; recall/precision: 0.630/0.569 to 0.447/0.406). The \textbf{writer} is essential for reliability: its removal collapses valid citation ratio to 0.610. The \textbf{keynote} and \textbf{analyzer} modules mainly contribute to depth. Removing the analyzer slightly improves precision (0.569 to 0.574) but substantially reduces depth, revealing a mild depth–reliability trade-off that the full system balances via coordinated design. The \textbf{refinement} stage brings consistent gains to both content and citation metrics. The simple skill-loop variant (\textbf{w/o agentic}) consistently underperforms the central planning agent on both metrics, indicating that globally coordinating review and revision with draft state, evidence, and revision history is more effective than locally iterating without shared planning context. The scores of detailed dimensions in the ablation experiment are in Appendix~\ref{app:detail_score_in_module_ablation}. Other ablation experiments are supplemented in Appendix~\ref{app:ablation_experiment_supp}.

\textbf{Code Agent Subsystem's Gain.} 

\begin{figure*}[!htp]
  \centering
  \includegraphics[width=0.7\textwidth]{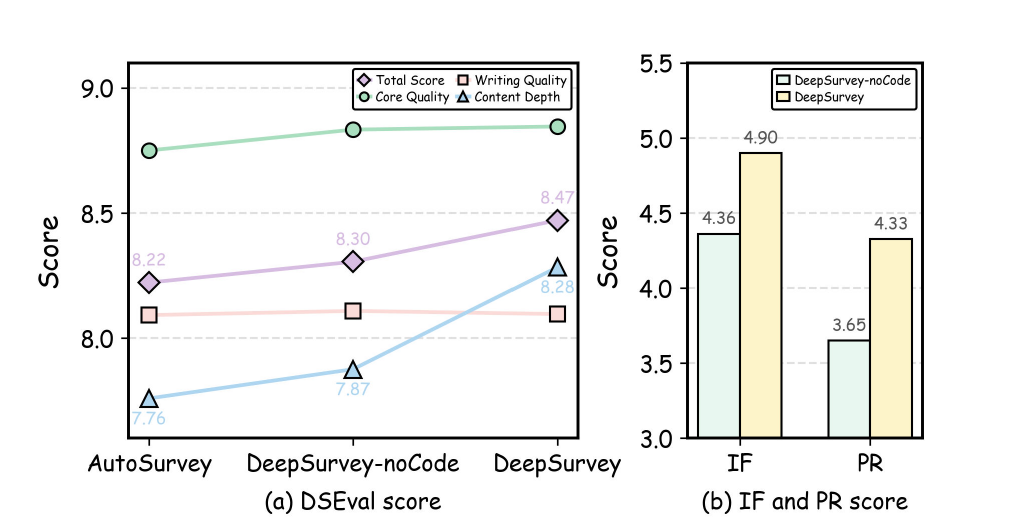}
  \caption{Comparison of AutoSurvey, DeepSurvey-noCode, and DeepSurvey in the CS domain: (a) DSEval scores; (b) Implementation Fidelity (IF) and Practical Reproducibility (PR) scores.}
  \label{fig:code_no_code}
\end{figure*}
Since code repositories are sparse in non-CS fields, we further compare DeepSurvey with DeepSurvey-noCode in the CS domain on DSEval, together with two advanced dimensions tailored to CS surveys: \textbf{Implementation Fidelity} (reflecting implementation details) and \textbf{Practical Reproducibility} (guiding the reproduction and improvement of baselines); detailed criteria are in Appendix~\ref{app:detailed_standard}. As shown in Figure~\ref{fig:code_no_code}, the code agent's gain concentrates in the CS domain ($+0.18$/$+0.41$ in Total Score/Content Depth versus $+0.12$/$+0.24$ across all domains in Table~\ref{tab:ablation_results}), with pronounced improvements on the advanced dimensions (IF and PR improve by $12.4\%$ and $18.5\%$).

\subsection{Meta-evaluation}\label{ssec:meta_eval}

Meta-evaluation is performed to examine the reliability of the evaluation protocol. First, for reliability of LLM-judge, repeated evaluations exhibit low coefficients of variation (CV)~\citep{PearsonMathematicalCT, Reed2002UseOC, Shoukri2006BmcMR}, with \textit{MiMo-no-exp} reaching 0.244\%, well below the 5\% threshold (Table~\ref{tab:stability_stats_main_tex}; Appendix~\ref{app:eval_stability}). These results confirm that the LLM judge is a reliable component of the evaluation protocol. Second, for LLM--human judge agreement, the LLM judge achieves a mean Cohen's $\kappa$~\citep{Cohen1960ACO} of $0.646$, close to the human--human agreement ($0.696$). At the group level, replacing one human with the LLM yields a Fleiss' $\kappa$~\citep{Fleiss1971MeasuringNS} of $0.640$, comparable to the all-human panel ($0.658$) and far above the random-replacement baseline ($0.246$) (Figure~\ref{fig:human_llm_coherence}; details in Appendix~\ref{app:eval_coherence}).

\section{Related Work}

\subsection{Traditional Literature Summarization}

Traditional literature summarization builds on content selection and citation-based relation modeling. Early extractive methods progressed from frequency-based heuristics~\citep{Luhn1958The} and trainable models~\citep{1995A} to graph-based centrality such as LexRank~\citep{DBLP:journals/corr/abs-1109-2128}. Citation indexing~\citep{doi:10.1126/science.122.3159.108} and co-citation analysis~\citep{repec:bla:jamest:v:24:y:1973:i:4:p:265-269} revealed that citation patterns encode field structure, and citation-aware summarization~\citep{cohan-goharian-2015-scientific,CitationAS} combined content understanding with inter-paper relation modeling.

\subsection{LLM-Based Survey Generation}

Recent LLM-based methods decompose survey generation into retrieval, planning, and writing stages. AutoSurvey~\citep{Wang2024AutoSurveyLL} established the paradigm of semantic retrieval with parallel generation, extended by AutoSurvey2~\citep{Wu2025AutoSurvey2ER} and LLM\(\times\)MapReduce~\citep{Zhou2024LLMMapReduceSL}. Subsequent work addressed specific dimensions: SurveyForge~\citep{Yan2025SurveyForgeOT} introduced memory-augmented retrieval; SurveyX~\citep{Liang2025SurveyXAS} proposed AttributeTree for structured representation; SurveyGen~\citep{bao-etal-2025-surveygen} incorporated quality-aware feedback; and multi-agent systems~\citep{Go2025LiRAAM,Liu2025AgenticAL} explored collaborative generation.

\section{Conclusion}


We present \textbf{DeepSurvey}, an agentic survey generation system built for agent consumers, for whom analytical depth and citation reliability matter more than breadth. It constructs a reusable analysis substrate through graph-based retrieval, full-text keynotes, cross-paper relation modeling, and code analysis, and enforces reliability through hybrid filtering, evidence-constrained analysis/writing, and multi-granularity refinement. Across multiple domains, DeepSurvey achieves the best content quality (8.341/10) and citation quality (0.630 recall, 0.569 precision, 25.3\%/35.2\% above the strongest baseline), and is preferred over human-written surveys by domain experts. Most notably, a coding agent reading a DeepSurvey survey autonomously outperforms other baselines, suggesting that the depth such surveys convey is what ultimately transfers to research agents.



\section*{AI Use Statement}
In this work, we used generative AI tools for designing or providing feedback on research methodology or experiments, supporting qualitative and thematic data analysis, assisting with translation, and interpreting results. We have not used generative AI tools for helping develop theoretical models or conceptual frameworks, proposing or refining hypotheses, assisting in the writing of proofs, implementing methods, or for cleaning and reformatting datasets, and the remaining required-disclosure tasks (generating synthetic data sets, formulating mathematical claims and providing critical ingredients for proving mathematical claims) are not applicable to this work. We have reviewed all AI-assisted work. We take responsibility for the final content of this work, including text, claims or artifacts produced with the aid of generative AI.


\section*{Ethics Statement}
Our human expert evaluation (Section~\ref{ssec:human_evaluation}, Appendix~\ref{app:human_exp}) involved only voluntary participants recruited from students and researchers with relevant academic experience; no crowdworkers were used. All annotations were performed in a blinded setting. Participants were informed that the data would be used exclusively for research and reported only in aggregate, and no personally identifying information was collected or will be released. All papers used in retrieval and evaluation are obtained through public academic services (Semantic Scholar, arXiv) in accordance with their terms of service. Human-written surveys are used solely for evaluation and are excluded from generation to prevent data leakage.

DeepSurvey is intended as an assistive tool for literature synthesis, not a means of producing deceptive academic content. Generated surveys are grounded in verifiable, source-attributed evidence, and we explicitly discourage using the system to generate publications or sharing its output without disclosing that it is not human-written. To further support responsible use, we plan to release an annotation tool that flags potentially unsupported claims and citations. We do not foresee other direct risks of harm, as the system neither collects personal data nor makes consequential automated decisions.

\section*{Reproducibility Statement}
The DeepSurvey pipeline is described in Section~\ref{sec:retrieval}--\ref{sec:refinement}, and the key prompts for clustering, outline planning, and drafting are provided in Appendix~\ref{sec:prompt_engineer}, which together allow the system to be re-implemented in full. All major models in our system are open-source: MiMo-V2.5 serves as the generation backbone (identical across DeepSurvey and all baselines), MiMo-V2.5 as the primary judge for both content and citation evaluation, and MiniLMv2 as the embedding model. Re-running the pipeline is therefore inexpensive (Appendix~\ref{app:cost}) and does not depend on proprietary APIs. The evaluation protocol is specified in detail, including evaluation topics across five domains (Appendix~\ref{app:eval_introduction_input}), metric definitions and aggregation weights (Appendix~\ref{app:eval_introduction_standard}), system and decoding parameters (Appendix~\ref{app:eval_eval_params}, Table~\ref{tab:system_params}), and judge-model validation against human experts (Appendix~\ref{app:eval_stability}, \ref{app:eval_coherence}). All baselines are run with their official implementations under documented settings. Main results are reported as mean and standard deviation over repeated runs, with effect sizes given in Appendix~\ref{app:eval_results_supp}. The complete source code is available in an anonymous repository (\url{https://anonymous.4open.science/r/DeepSurvey-anonymous}) and is also submitted with the supplementary material; the full repository, together with the prompts and the evaluation annotation tool, will be publicly released upon publication.

\newpage
\bibliographystyle{xlance/IEEEtran2}
\setcitestyle{numbers}
\bibliography{venues,reference}

@article{
doi:10.1126/science.122.3159.108,
author = {Eugene Garfield },
title = {Citation Indexes for Science},
journal = {Science},
volume = {122},
number = {3159},
pages = {108-111},
year = {1955},
doi = {10.1126/science.122.3159.108},
URL = {https://www.science.org/doi/abs/10.1126/science.122.3159.108},
eprint = {https://www.science.org/doi/pdf/10.1126/science.122.3159.108}}

@article{repec:bla:jamest:v:24:y:1973:i:4:p:265-269,
journal={Journal of the American Society for Information Science},
author={Henry Small},
title={Co-citation in the scientific literature: A new measure of the relationship between two documents},
year={1973},
month={July},
pages={265-269},
volume={24},
number={4},
doi={10.1002/asi.4630240406},
url={https://ideas.repec.org/a/bla/jamest/v24y1973i4p265-269.html},
}

@article{CitationAS,
author = {Wang, Jie and Zhang, Chengzhi and Zhang, Mengying and Deng, Sanhong},
year = {2018},
month = {06},
pages = {20-37},
title = {CitationAS: A Tool of Automatic Survey Generation Based on Citation Content},
volume = {3},
journal = {Journal of Data and Information Science},
doi = {10.2478/jdis-2018-0007}
}

@article{DBLP:journals/corr/abs-1109-2128,
  author       = {G{\"{u}}nes Erkan and
                  Dragomir R. Radev},
  title        = {LexRank: Graph-based Lexical Centrality as Salience in Text Summarization},
  journal      = {CoRR},
  volume       = {abs/1109.2128},
  year         = {2011},
  url          = {http://arxiv.org/abs/1109.2128},
  eprinttype   = {arXiv},
  eprint       = {1109.2128},
  bibsource    = {dblp computer science bibliography, https://dblp.org}
}

@inproceedings{cohan-goharian-2015-scientific,
    title = "Scientific Article Summarization Using Citation-Context and Article{'}s Discourse Structure",
    author = "Cohan, Arman  and
      Goharian, Nazli",
    editor = "M{\`a}rquez, Llu{\'i}s  and
      Callison-Burch, Chris  and
      Su, Jian",
    booktitle = "Proceedings of the 2015 Conference on Empirical Methods in Natural Language Processing",
    month = sep,
    year = "2015",
    address = "Lisbon, Portugal",
    publisher = "Association for Computational Linguistics",
    url = "https://aclanthology.org/D15-1045/",
    doi = "10.18653/v1/D15-1045",
    pages = "390--400"
}

@article{Luhn1958The,
  title={The Automatic Creation of Literature Abstracts},
  author={Hans Peter Luhn},
  journal={IBM J. Res. Dev.},
  year={1958},
  volume={2},
  pages={159-165},
  url={https://api.semanticscholar.org/CorpusID:15475171}
}

@inproceedings{1995A,
  title={A trainable document summarizer},
  author={Julian Kupiec and Jan O. Pedersen and Francine R. Chen},
  booktitle={Annual International ACM SIGIR Conference on Research and Development in Information Retrieval},
  year={1995},
  url={https://api.semanticscholar.org/CorpusID:5775833}
}

@article{Wang2024AutoSurveyLL,
  title={AutoSurvey: Large Language Models Can Automatically Write Surveys},
  author={Yidong Wang and Qi Guo and Wenjin Yao and Hongbo Zhang and Xin Zhang and Zhen Wu and Meishan Zhang and Xinyu Dai and Min Zhang and Qingsong Wen and Wei Ye and Shikun Zhang and Yue Zhang},
  journal={ArXiv},
  year={2024},
  volume={abs/2406.10252},
  url={https://api.semanticscholar.org/CorpusID:270560509}
}

@article{Wu2025AutoSurvey2ER,
  title={AutoSurvey2: Empowering Researchers with Next Level Automated Literature Surveys},
  author={Siyi Wu and Chia Xin Liang and Ziqian Bi and Leyi Zhao and Tianyang Wang and Jun-Jie Song and Yichao Zhang and Keyu Chen and Xinyuan Song},
  journal={ArXiv},
  year={2025},
  volume={abs/2510.26012},
  url={https://api.semanticscholar.org/CorpusID:282592838}
}

@inproceedings{bao-etal-2025-surveygen,
    title = "{S}urvey{G}en: Quality-Aware Scientific Survey Generation with Large Language Models",
    author = "Bao, Tong  and
      Nayeem, Mir Tafseer  and
      Rafiei, Davood  and
      Zhang, Chengzhi",
    editor = "Christodoulopoulos, Christos  and
      Chakraborty, Tanmoy  and
      Rose, Carolyn  and
      Peng, Violet",
    booktitle = "Proceedings of the 2025 Conference on Empirical Methods in Natural Language Processing",
    month = nov,
    year = "2025",
    address = "Suzhou, China",
    publisher = "Association for Computational Linguistics",
    url = "https://aclanthology.org/2025.emnlp-main.136/",
    doi = "10.18653/v1/2025.emnlp-main.136",
    pages = "2712--2736",
    ISBN = "979-8-89176-332-6"
}

@inproceedings{Zhou2024LLMMapReduceSL,
    title = "{LLM}$\times${M}ap{R}educe: Simplified Long-Sequence Processing using Large Language Models",
    author = "Zhou, Zihan  and
      Li, Chong  and
      Chen, Xinyi  and
      Wang, Shuo  and
      Chao, Yu  and
      Li, Zhili  and
      Wang, Haoyu  and
      Shi, Qi  and
      Tan, Zhixing  and
      Han, Xu  and
      Shi, Xiaodong  and
      Liu, Zhiyuan  and
      Sun, Maosong",
    editor = "Che, Wanxiang  and
      Nabende, Joyce  and
      Shutova, Ekaterina  and
      Pilehvar, Mohammad Taher",
    booktitle = "Proceedings of the 63rd Annual Meeting of the Association for Computational Linguistics (Volume 1: Long Papers)",
    month = jul,
    year = "2025",
    address = "Vienna, Austria",
    publisher = "Association for Computational Linguistics",
    url = "https://aclanthology.org/2025.acl-long.1341/",
    doi = "10.18653/v1/2025.acl-long.1341",
    pages = "27664--27678",
    ISBN = "979-8-89176-251-0"
}

@misc{Wang2025LLMtimesMapReduceV2EC,
      title={LLM$\times$MapReduce-V2: Entropy-Driven Convolutional Test-Time Scaling for Generating Long-Form Articles from Extremely Long Resources}, 
      author={Haoyu Wang and Yujia Fu and Zhu Zhang and Shuo Wang and Zirui Ren and Xiaorong Wang and Zhili Li and Chaoqun He and Bo An and Zhiyuan Liu and Maosong Sun},
      year={2025},
      eprint={2504.05732},
      archivePrefix={arXiv},
      primaryClass={cs.CL},
      url={https://arxiv.org/abs/2504.05732}, 
}

@article{Shi2025SciSageAM,
  title={SciSage: A Multi-Agent Framework for High-Quality Scientific Survey Generation},
  author={Xiaofeng Shi and Qian Kou and Yuduo Li and Ning Tang and Jinxin Xie and Longbin Yu and Songjing Wang and Hua Zhou},
  journal={ArXiv},
  year={2025},
  volume={abs/2506.12689},
  url={https://api.semanticscholar.org/CorpusID:279402998}
}

@article{Liang2025SurveyXAS,
  title={SurveyX: Academic Survey Automation via Large Language Models},
  author={Xun Liang and Jiawei Yang and Yezhaohui Wang and Chen Tang and Zifan Zheng and Simin Niu and Shichao Song and Zehao Lin and Hanyu Wang and Bo Tang and Feiyu Xiong and Keming Mao and Zhiyu Li},
  journal={ArXiv},
  year={2025},
  volume={abs/2502.14776},
  url={https://api.semanticscholar.org/CorpusID:276482768}
}

@inproceedings{Yan2025SurveyForgeOT,
  title={SurveyForge: On the Outline Heuristics, Memory-Driven Generation, and Multi-dimensional Evaluation for Automated Survey Writing},
  author={Xiangchao Yan and Shiyang Feng and Jiakang Yuan and Renqiu Xia and Bin Wang and Bo Zhang and Lei Bai},
  booktitle={Annual Meeting of the Association for Computational Linguistics},
  year={2025},
  url={https://api.semanticscholar.org/CorpusID:276813240}
}

@inproceedings{Go2025LiRAAM,
  title={LiRA: A Multi-Agent Framework for Reliable and Readable Literature Review Generation},
  author={Gregory Hok Tjoan Go and Khang Ly and Anders Sogaard and Amin Reza Tabatabaei and Maarten de Rijke and Xinyi Chen},
  booktitle={AAAI Conference on Artificial Intelligence},
  year={2025},
  url={https://api.semanticscholar.org/CorpusID:281886200}
}

@article{Liu2025AgenticAL,
  title={Agentic AutoSurvey: Let LLMs Survey LLMs},
  author={Yixin Liu and Yonghui Wu and Denghui Zhang and Lichao Sun},
  journal={ArXiv},
  year={2025},
  volume={abs/2509.18661},
  url={https://api.semanticscholar.org/CorpusID:281495800}
}

@article{2017Explicit,
  title={Explicit Semantic Ranking for Academic Search via Knowledge Graph Embedding},
  author={Chenyan Xiong and Russell Power and Jamie Callan},
  journal={Proceedings of the 26th International Conference on World Wide Web},
  year={2017},
  url={https://api.semanticscholar.org/CorpusID:1644335}
}

@article{Wade2022TheSS,
  title={The Semantic Scholar Academic Graph (S2AG)},
  author={Alex D Wade},
  journal={Companion Proceedings of the Web Conference 2022},
  year={2022},
  url={https://api.semanticscholar.org/CorpusID:251597885}
}

@inproceedings{lo-etal-2020-s2orc,
    title = "{S}2{ORC}: The Semantic Scholar Open Research Corpus",
    author = "Lo, Kyle  and
      Wang, Lucy Lu  and
      Neumann, Mark  and
      Kinney, Rodney  and
      Weld, Daniel",
    editor = "Jurafsky, Dan  and
      Chai, Joyce  and
      Schluter, Natalie  and
      Tetreault, Joel",
    booktitle = "Proceedings of the 58th Annual Meeting of the Association for Computational Linguistics",
    month = jul,
    year = "2020",
    address = "Online",
    publisher = "Association for Computational Linguistics",
    url = "https://aclanthology.org/2020.acl-main.447/",
    doi = "10.18653/v1/2020.acl-main.447",
    pages = "4969--4983"
}

@article{Xiao2026MiMoV2FlashTR,
  title={MiMo-V2-Flash Technical Report},
  author={Xi Xiao and Bing Xia and Bo Yang and Bofei Gao and Bowen Shen and Chen Zhang and Chenhong He and Chiheng Lou and Fuli Luo and Gang Wang and Gang Xie and Hailin Zhang and Hanglong Lv and Hanyu Li and Heyu Chen and Hong-Mei Xu and Houbin Zhang and Huaqiu Liu and Jiangshan Duo and Jianyu Wei and Jiebao Xiao and Jinhao Dong and Jun-Miao Shi and Jun Jacob Hu and Kainan Bao and Kang Zhou and Lei Li and Liang Zhao and Linghao Zhang and Peidian Li and Qian Chen and Shao-yang Liu and Shi-liang Yu and Shijie Cao and Shimao Chen and Shouqiu Yu and Shuo Liu and Tian-Yu Zhou and Wei Su and Weikun Wang and Wenhan Ma and Xia Deng and Bo Mao and Bowen Ye and Can Cai and Chenghua Wang and Chengxuan Zhu and Chong Ma and Chun Chen and Chunan Li and Dawei Zhu and Deshan Xiao and Dong Zhang and Duo Zhang and Fang Liu and Feiyu Yang and Feng Shi and Guoan Wang and Hao Tian and Hao Wu and Hengxu Qu and Hong Yi and Hongxu An and Hongyi Guan and Xing Zhang and Yi-Hao Song and Yi-Tong Yan and Yihao Zhao and Ying Lai and Yizhao Gao and Yu Cheng and Yu Tian and Yudong Wang and Zheng-Yu Tang and Zheng-Yu Tang and Zheng Wen and Zhichao Song and Zhixian Zheng and Zi-Ang Jiang and Jiantao Wen and Jiarui Sun and Jiawei Li and Jinlong Xue and Jun Xia and Kai Fang and Menghang Zhu and Nuo Chen and Qian Tu and Qihao Zhang and Qiying Wang and Rang Li and Rui Ma and Shao-Qiang Zhang and Shengfan Wang and Shicheng Li and Shuhao Gu and Shu-Yue Ren and Sirui Deng and Tao Guo and Tianyang Lu and Weiji Zhuang and Weikang Zhang and Weimin Xiong and Wen-Jie Huang and Wenyu Yang and Xin Zhang and Xing Yong and Xu Wang and Xueyang Xie and Yilin Jiang and Yixin Yang and Yongzhe He and Yuanyu Tu and Yu-Jie Dong and Yuchen Liu and Yue Ma and Yue Yu and Yu-Cui Xiang and Zhaojun Huang and Zhenrui Lin and Zhipeng Xu and Zhiyang Chen and Zhonghua Deng and Zihan Zhang and Zihao Yue},
  journal={ArXiv},
  year={2026},
  volume={abs/2601.02780},
  url={https://api.semanticscholar.org/CorpusID:284513060}
}

@misc{Singh2025OpenAIGS,
      title={OpenAI GPT-5 System Card}, 
      author={Aaditya Singh and Adam Fry and Adam Perelman and Adam Tart and Adi Ganesh and Ahmed El-Kishky and Aidan McLaughlin and Aiden Low and AJ Ostrow and Akhila Ananthram and Akshay Nathan and Alan Luo and Alec Helyar and Aleksander Madry and Aleksandr Efremov and Aleksandra Spyra and Alex Baker-Whitcomb and Alex Beutel and Alex Karpenko and Alex Makelov and Alex Neitz and Alex Wei and Alexandra Barr and Alexandre Kirchmeyer and Alexey Ivanov and Alexi Christakis and Alistair Gillespie and Allison Tam and Ally Bennett and Alvin Wan and Alyssa Huang and Amy McDonald Sandjideh and Amy Yang and Ananya Kumar and Andre Saraiva and Andrea Vallone and Andrei Gheorghe and Andres Garcia Garcia and Andrew Braunstein and Andrew Liu and Andrew Schmidt and Andrey Mereskin and Andrey Mishchenko and Andy Applebaum and Andy Rogerson and Ann Rajan and Annie Wei and Anoop Kotha and Anubha Srivastava and Anushree Agrawal and Arun Vijayvergiya and Ashley Tyra and Ashvin Nair and Avi Nayak and Ben Eggers and Bessie Ji and Beth Hoover and Bill Chen and Blair Chen and Boaz Barak and Borys Minaiev and Botao Hao and Bowen Baker and Brad Lightcap and Brandon McKinzie and Brandon Wang and Brendan Quinn and Brian Fioca and Brian Hsu and Brian Yang and Brian Yu and Brian Zhang and Brittany Brenner and Callie Riggins Zetino and Cameron Raymond and Camillo Lugaresi and Carolina Paz and Cary Hudson and Cedric Whitney and Chak Li and Charles Chen and Charlotte Cole and Chelsea Voss and Chen Ding and Chen Shen and Chengdu Huang and Chris Colby and Chris Hallacy and Chris Koch and Chris Lu and Christina Kaplan and Christina Kim and CJ Minott-Henriques and Cliff Frey and Cody Yu and Coley Czarnecki and Colin Reid and Colin Wei and Cory Decareaux and Cristina Scheau and Cyril Zhang and Cyrus Forbes and Da Tang and Dakota Goldberg and Dan Roberts and Dana Palmie and Daniel Kappler and Daniel Levine and Daniel Wright and Dave Leo and David Lin and David Robinson and Declan Grabb and Derek Chen and Derek Lim and Derek Salama and Dibya Bhattacharjee and Dimitris Tsipras and Dinghua Li and Dingli Yu and DJ Strouse and Drew Williams and Dylan Hunn and Ed Bayes and Edwin Arbus and Ekin Akyurek and Elaine Ya Le and Elana Widmann and Eli Yani and Elizabeth Proehl and Enis Sert and Enoch Cheung and Eri Schwartz and Eric Han and Eric Jiang and Eric Mitchell and Eric Sigler and Eric Wallace and Erik Ritter and Erin Kavanaugh and Evan Mays and Evgenii Nikishin and Fangyuan Li and Felipe Petroski Such and Filipe de Avila Belbute Peres and Filippo Raso and Florent Bekerman and Foivos Tsimpourlas and Fotis Chantzis and Francis Song and Francis Zhang and Gaby Raila and Garrett McGrath and Gary Briggs and Gary Yang and Giambattista Parascandolo and Gildas Chabot and Grace Kim and Grace Zhao and Gregory Valiant and Guillaume Leclerc and Hadi Salman and Hanson Wang and Hao Sheng and Haoming Jiang and Haoyu Wang and Haozhun Jin and Harshit Sikchi and Heather Schmidt and Henry Aspegren and Honglin Chen and Huida Qiu and Hunter Lightman and Ian Covert and Ian Kivlichan and Ian Silber and Ian Sohl and Ibrahim Hammoud and Ignasi Clavera and Ikai Lan and Ilge Akkaya and Ilya Kostrikov and Irina Kofman and Isak Etinger and Ishaan Singal and Jackie Hehir and Jacob Huh and Jacqueline Pan and Jake Wilczynski and Jakub Pachocki and James Lee and James Quinn and Jamie Kiros and Janvi Kalra and Jasmyn Samaroo and Jason Wang and Jason Wolfe and Jay Chen and Jay Wang and Jean Harb and Jeffrey Han and Jeffrey Wang and Jennifer Zhao and Jeremy Chen and Jerene Yang and Jerry Tworek and Jesse Chand and Jessica Landon and Jessica Liang and Ji Lin and Jiancheng Liu and Jianfeng Wang and Jie Tang and Jihan Yin and Joanne Jang and Joel Morris and Joey Flynn and Johannes Ferstad and Johannes Heidecke and John Fishbein and John Hallman and Jonah Grant and Jonathan Chien and Jonathan Gordon and Jongsoo Park and Jordan Liss and Jos Kraaijeveld and Joseph Guay and Joseph Mo and Josh Lawson and Josh McGrath and Joshua Vendrow and Joy Jiao and Julian Lee and Julie Steele and Julie Wang and Junhua Mao and Kai Chen and Kai Hayashi and Kai Xiao and Kamyar Salahi and Kan Wu and Karan Sekhri and Karan Sharma and Karan Singhal and Karen Li and Kenny Nguyen and Keren Gu-Lemberg and Kevin King and Kevin Liu and Kevin Stone and Kevin Yu and Kristen Ying and Kristian Georgiev and Kristie Lim and Kushal Tirumala and Kyle Miller and Lama Ahmad and Larry Lv and Laura Clare and Laurance Fauconnet and Lauren Itow and Lauren Yang and Laurentia Romaniuk and Leah Anise and Lee Byron and Leher Pathak and Leon Maksin and Leyan Lo and Leyton Ho and Li Jing and Liang Wu and Liang Xiong and Lien Mamitsuka and Lin Yang and Lindsay McCallum and Lindsey Held and Liz Bourgeois and Logan Engstrom and Lorenz Kuhn and Louis Feuvrier and Lu Zhang and Lucas Switzer and Lukas Kondraciuk and Lukasz Kaiser and Manas Joglekar and Mandeep Singh and Mandip Shah and Manuka Stratta and Marcus Williams and Mark Chen and Mark Sun and Marselus Cayton and Martin Li and Marvin Zhang and Marwan Aljubeh and Matt Nichols and Matthew Haines and Max Schwarzer and Mayank Gupta and Meghan Shah and Melody Y. Guan and Melody Huang and Meng Dong and Mengqing Wang and Mia Glaese and Micah Carroll and Michael Lampe and Michael Malek and Michael Sharman and Michael Zhang and Michele Wang and Michelle Pokrass and Mihai Florian and Mikhail Pavlov and Miles Wang and Ming Chen and Mingxuan Wang and Minnia Feng and Mo Bavarian and Molly Lin and Moose Abdool and Mostafa Rohaninejad and Nacho Soto and Natalie Staudacher and Natan LaFontaine and Nathan Marwell and Nelson Liu and Nick Preston and Nick Turley and Nicklas Ansman and Nicole Blades and Nikil Pancha and Nikita Mikhaylin and Niko Felix and Nikunj Handa and Nishant Rai and Nitish Keskar and Noam Brown and Ofir Nachum and Oleg Boiko and Oleg Murk and Olivia Watkins and Oona Gleeson and Pamela Mishkin and Patryk Lesiewicz and Paul Baltescu and Pavel Belov and Peter Zhokhov and Philip Pronin and Phillip Guo and Phoebe Thacker and Qi Liu and Qiming Yuan and Qinghua Liu and Rachel Dias and Rachel Puckett and Rahul Arora and Ravi Teja Mullapudi and Raz Gaon and Reah Miyara and Rennie Song and Rishabh Aggarwal and RJ Marsan and Robel Yemiru and Robert Xiong and Rohan Kshirsagar and Rohan Nuttall and Roman Tsiupa and Ronen Eldan and Rose Wang and Roshan James and Roy Ziv and Rui Shu and Ruslan Nigmatullin and Saachi Jain and Saam Talaie and Sam Altman and Sam Arnesen and Sam Toizer and Sam Toyer and Samuel Miserendino and Sandhini Agarwal and Sarah Yoo and Savannah Heon and Scott Ethersmith and Sean Grove and Sean Taylor and Sebastien Bubeck and Sever Banesiu and Shaokyi Amdo and Shengjia Zhao and Sherwin Wu and Shibani Santurkar and Shiyu Zhao and Shraman Ray Chaudhuri and Shreyas Krishnaswamy and Shuaiqi and Xia and Shuyang Cheng and Shyamal Anadkat and Simón Posada Fishman and Simon Tobin and Siyuan Fu and Somay Jain and Song Mei and Sonya Egoian and Spencer Kim and Spug Golden and SQ Mah and Steph Lin and Stephen Imm and Steve Sharpe and Steve Yadlowsky and Sulman Choudhry and Sungwon Eum and Suvansh Sanjeev and Tabarak Khan and Tal Stramer and Tao Wang and Tao Xin and Tarun Gogineni and Taya Christianson and Ted Sanders and Tejal Patwardhan and Thomas Degry and Thomas Shadwell and Tianfu Fu and Tianshi Gao and Timur Garipov and Tina Sriskandarajah and Toki Sherbakov and Tomek Korbak and Tomer Kaftan and Tomo Hiratsuka and Tongzhou Wang and Tony Song and Tony Zhao and Troy Peterson and Val Kharitonov and Victoria Chernova and Vineet Kosaraju and Vishal Kuo and Vitchyr Pong and Vivek Verma and Vlad Petrov and Wanning Jiang and Weixing Zhang and Wenda Zhou and Wenlei Xie and Wenting Zhan and Wes McCabe and Will DePue and Will Ellsworth and Wulfie Bain and Wyatt Thompson and Xiangning Chen and Xiangyu Qi and Xin Xiang and Xinwei Shi and Yann Dubois and Yaodong Yu and Yara Khakbaz and Yifan Wu and Yilei Qian and Yin Tat Lee and Yinbo Chen and Yizhen Zhang and Yizhong Xiong and Yonglong Tian and Young Cha and Yu Bai and Yu Yang and Yuan Yuan and Yuanzhi Li and Yufeng Zhang and Yuguang Yang and Yujia Jin and Yun Jiang and Yunyun Wang and Yushi Wang and Yutian Liu and Zach Stubenvoll and Zehao Dou and Zheng Wu and Zhigang Wang},
      year={2026},
      eprint={2601.03267},
      archivePrefix={arXiv},
      primaryClass={cs.CL},
      url={https://arxiv.org/abs/2601.03267}, 
}

@article{Wang2020MiniLMv2MS,
  title={MiniLMv2: Multi-Head Self-Attention Relation Distillation for Compressing Pretrained Transformers},
  author={Wenhui Wang and Hangbo Bao and Shaohan Huang and Li Dong and Furu Wei},
  journal={ArXiv},
  year={2020},
  volume={abs/2012.15828},
  url={https://api.semanticscholar.org/CorpusID:229923069}
}

@misc{Wang2024MinerUAO,
      title={MinerU: An Open-Source Solution for Precise Document Content Extraction}, 
      author={Bin Wang and Chao Xu and Xiaomeng Zhao and Linke Ouyang and Fan Wu and Zhiyuan Zhao and Rui Xu and Kaiwen Liu and Yuan Qu and Fukai Shang and Bo Zhang and Liqun Wei and Zhihao Sui and Wei Li and Botian Shi and Yu Qiao and Dahua Lin and Conghui He},
      year={2024},
      eprint={2409.18839},
      archivePrefix={arXiv},
      primaryClass={cs.CV},
      url={https://arxiv.org/abs/2409.18839}, 
}

@article{Zheng2023JudgingLW,
  title={Judging LLM-as-a-judge with MT-Bench and Chatbot Arena},
  author={Lianmin Zheng and Wei-Lin Chiang and Ying Sheng and Siyuan Zhuang and Zhanghao Wu and Yonghao Zhuang and Zi Lin and Zhuohan Li and Dacheng Li and Eric P. Xing and Haotong Zhang and Joseph E. Gonzalez and Ion Stoica},
  journal={ArXiv},
  year={2023},
  volume={abs/2306.05685},
  url={https://api.semanticscholar.org/CorpusID:259129398}
}

@inproceedings{Liu2023GEvalNE,
    title = "{G}-Eval: {NLG} Evaluation using Gpt-4 with Better Human Alignment",
    author = "Liu, Yang  and
      Iter, Dan  and
      Xu, Yichong  and
      Wang, Shuohang  and
      Xu, Ruochen  and
      Zhu, Chenguang",
    editor = "Bouamor, Houda  and
      Pino, Juan  and
      Bali, Kalika",
    booktitle = "Proceedings of the 2023 Conference on Empirical Methods in Natural Language Processing",
    month = dec,
    year = "2023",
    address = "Singapore",
    publisher = "Association for Computational Linguistics",
    url = "https://aclanthology.org/2023.emnlp-main.153/",
    doi = "10.18653/v1/2023.emnlp-main.153",
    pages = "2511--2522"
}

@article{Zhu2023JudgeLMFL,
  title={JudgeLM: Fine-tuned Large Language Models are Scalable Judges},
  author={Lianghui Zhu and Xinggang Wang and Xinlong Wang},
  journal={ArXiv},
  year={2023},
  volume={abs/2310.17631},
  url={https://api.semanticscholar.org/CorpusID:264490588}
}

@article{Kim2024Prometheus2A,
  title={Prometheus 2: An Open Source Language Model Specialized in Evaluating Other Language Models},
  author={Seungone Kim and Juyoung Suk and Shayne Longpre and Bill Yuchen Lin and Jamin Shin and Sean Welleck and Graham Neubig and Moontae Lee and Kyungjae Lee and Minjoon Seo},
  journal={ArXiv},
  year={2024},
  volume={abs/2405.01535},
  url={https://api.semanticscholar.org/CorpusID:269502688}
}

@article{Liu2023LostIT,
  title={Lost in the Middle: How Language Models Use Long Contexts},
  author={Nelson F. Liu and Kevin Lin and John Hewitt and Ashwin Paranjape and Michele Bevilacqua and Fabio Petroni and Percy Liang},
  journal={Transactions of the Association for Computational Linguistics},
  year={2023},
  volume={12},
  pages={157-173},
  url={https://api.semanticscholar.org/CorpusID:259360665}
}

@article{Du2025ContextLA,
  title={Context Length Alone Hurts LLM Performance Despite Perfect Retrieval},
  author={Yufeng Du and Minyang Tian and S. Ronanki and Subendhu Rongali and S. Bodapati and A.G. Galstyan and Azton Wells and Roy Schwartz and Eliu A. Huerta and Hao Peng},
  journal={ArXiv},
  year={2025},
  volume={abs/2510.05381},
  url={https://api.semanticscholar.org/CorpusID:281826429}
}

@inproceedings{Liu2025TowardsLC,
  title={Towards Long Context Hallucination Detection},
  author={Siyi Liu and Kishaloy Halder and Zheng Qi and Wei Xiao and Nikolaos Pappas and Phu Mon Htut and Neha Ann John and Yassine Benajiba and Dan Roth},
  booktitle={North American Chapter of the Association for Computational Linguistics},
  year={2025},
  url={https://api.semanticscholar.org/CorpusID:278165123}
}

@article{Manakul2023SelfCheckGPTZB,
  title={SelfCheckGPT: Zero-Resource Black-Box Hallucination Detection for Generative Large Language Models},
  author={Potsawee Manakul and Adian Liusie and Mark John Francis Gales},
  journal={ArXiv},
  year={2023},
  volume={abs/2303.08896},
  url={https://api.semanticscholar.org/CorpusID:257557820}
}

@article{Madaan2024LostII,
  title={Lost in Inference: Rediscovering the Role of Natural Language Inference for Large Language Models},
  author={Lovish Madaan and David Esiobu and Pontus Stenetorp and Barbara Plank and Dieuwke Hupkes},
  journal={ArXiv},
  year={2024},
  volume={abs/2411.14103},
  url={https://api.semanticscholar.org/CorpusID:274165430}
}

@inproceedings{chen-etal-2025-explainable,
    title = "Explainable Hallucination through Natural Language Inference Mapping",
    author = "Chen, Wei-Fan  and
      Zhao, Zhixue  and
      Karimi, Akbar  and
      Flek, Lucie",
    editor = "Che, Wanxiang  and
      Nabende, Joyce  and
      Shutova, Ekaterina  and
      Pilehvar, Mohammad Taher",
    booktitle = "Findings of the Association for Computational Linguistics: ACL 2025",
    month = jul,
    year = "2025",
    address = "Vienna, Austria",
    publisher = "Association for Computational Linguistics",
    url = "https://aclanthology.org/2025.findings-acl.96/",
    doi = "10.18653/v1/2025.findings-acl.96",
    pages = "1888--1896",
    ISBN = "979-8-89176-256-5"
}

@article{Nguye2025SurveyGAM,
  title={SurveyG: A Multi-Agent LLM Framework with Hierarchical Citation Graph for Automated Survey Generation},
  author={Minh-Anh Nguye and Minh Duc Nguyen and Ha Lan N.T. and Kieu Hai Dang and Nguyen Thanh Dong and Dung D. Le},
  journal={ArXiv},
  year={2025},
  volume={abs/2510.07733},
  url={https://api.semanticscholar.org/CorpusID:281951129}
}

@article{Shoukri2006BmcMR,
  title={Interval estimation and optimal design for the within-subject coefficient of variation for continuous and binary variables},
  author={Mohamed M Shoukri and Nasser Elkum and Stephen D. Walter},
  journal={BMC Medical Research Methodology},
  year={2006},
  volume={6},
  pages={24 - 24},
  url={https://api.semanticscholar.org/CorpusID:264815049}
}

@article{Reed2002UseOC,
  title={Use of Coefficient of Variation in Assessing Variability of Quantitative Assays},
  author={George F. Reed and Freyja Lynn and Bruce D. Meade},
  journal={Clinical and Vaccine Immunology},
  year={2002},
  volume={9},
  pages={1235 - 1239},
  url={https://api.semanticscholar.org/CorpusID:231319}
}

@article{Lu2024TheAS,
  title={The AI Scientist: Towards Fully Automated Open-Ended Scientific Discovery},
  author={Chris Lu and Cong Lu and Robert Tjarko Lange and Jakob Foerster and Jeff Clune and David Ha},
  journal={ArXiv},
  year={2024},
  volume={abs/2408.06292},
  url={https://api.semanticscholar.org/CorpusID:271854887}
}

@article{Yamada2025TheAS,
  title={The AI Scientist-v2: Workshop-Level Automated Scientific Discovery via Agentic Tree Search},
  author={Yutaro Yamada and Robert Tjarko Lange and Cong Lu and Shengran Hu and Chris Lu and Jakob Foerster and Jeff Clune and David Ha},
  journal={ArXiv},
  year={2025},
  volume={abs/2504.08066},
  url={https://api.semanticscholar.org/CorpusID:277741107}
}

@article{Su2025SurGEAB,
  title={Surge: A benchmark and evaluation framework for scientific survey generation},
  author={Su, Weihang and Xie, Anzhe and Ai, Qingyao and Long, Jianming and Chen, Xuanyi and Mao, Jiaxin and Ye, Ziyi and Liu, Yiqun},
  journal={arXiv preprint arXiv:2508.15658},
  year={2025}
}

@inproceedings{Chao2025LLMMapReduceV3EI,
  title={LLM×MapReduce-V3: Enabling Interactive In-Depth Survey Generation through a MCP-Driven Hierarchically Modular Agent System},
  author={Yu Chao and Siyu Lin and Xiaorong Wang and Zhu Zhang and Zihan Zhou and Haoyu Wang and Shuo Wang and Jie Zhou and Zhiyuan Liu and Maosong Sun},
  booktitle={Conference on Empirical Methods in Natural Language Processing},
  year={2025},
  url={https://api.semanticscholar.org/CorpusID:282057493}
}

@article{Cohen1960ACO,
  title={A Coefficient of Agreement for Nominal Scales},
  author={Jacob Cohen},
  journal={Educational and Psychological Measurement},
  year={1960},
  volume={20},
  pages={37 - 46},
  url={https://api.semanticscholar.org/CorpusID:15926286}
}

@article{Fleiss1971MeasuringNS,
  title={Measuring nominal scale agreement among many raters.},
  author={Joseph L. Fleiss},
  journal={Psychological Bulletin},
  year={1971},
  volume={76},
  pages={378-382},
  url={https://api.semanticscholar.org/CorpusID:143544759}
}

@article{PearsonMathematicalCT,
  title={Mathematical contributions to the theory of evolution.—On a form of spurious correlation which may arise when indices are used in the measurement of organs},
  author={Pearson, Karl},
  journal={Proceedings of the royal society of london},
  volume={60},
  number={359-367},
  pages={489--498},
  year={1897},
  publisher={The Royal Society London}
}

@article{DeepSeekAI2024DeepSeekV3TR,
  title={DeepSeek-V3 Technical Report},
  author={DeepSeek-AI and Aixin Liu and Bei Feng and Bing Xue and Bing-Li Wang and Bochao Wu and Chengda Lu and Chenggang Zhao and Chengqi Deng and Chenyu Zhang and Chong Ruan and Damai Dai and Daya Guo and Dejian Yang and Deli Chen and Dong-Li Ji and Erhang Li and Fangyun Lin and Fucong Dai and Fuli Luo and Guangbo Hao and Guanting Chen and Guowei Li and H. Zhang and Han Bao and Hanwei Xu and Haocheng Wang and Haowei Zhang and Honghui Ding and Huajian Xin and Huazuo Gao and Hui Li and Hui Qu and J. L. Cai and Jian Liang and Jianzhong Guo and Jiaqi Ni and Jiashi Li and Jiawei Wang and Jin Chen and JingChang Chen and Jingyang Yuan and Junjie Qiu and Junlong Li and Jun-Mei Song and Kai Dong and Kai Hu and Kaige Gao and Kang Guan and Kexin Huang and Kuai Yu and Lean Wang and Lecong Zhang and Lei Xu and Leyi Xia and Liang Zhao and Litong Wang and Liyue Zhang and Meng Li and Miaojun Wang and Mingchuan Zhang and Minghua Zhang and Minghui Tang and Mingming Li and Ning Tian and Panpan Huang and Peiyi Wang and Peng Zhang and Qiancheng Wang and Qihao Zhu and Qinyu Chen and Qiushi Du and R. J. Chen and Rui-Qi Jin and Ruiqi Ge and Ruisong Zhang and Ruizhe Pan and Runji Wang and Runxin Xu and Ruoyu Zhang and Ruyi Chen and S. S. Li and Shanghao Lu and Shangyan Zhou and Shanhuang Chen and Shao-Ping Wu and Shengfeng Ye and Shirong Ma and Shiyu Wang and Shuang Zhou and Shuiping Yu and Shunfeng Zhou and Shuting Pan and T. Wang and Tao Yun and Tian Pei and T. Sun and Wangding Xiao and Wangding Zeng and Wanjia Zhao and Wei An and Wen Liu and Wenfeng Liang and Wenjun Gao and Wen-xuan Yu and Wentao Zhang and X. Q. Li and Xiangyu Jin and Xianzu Wang and Xiaoling Bi and Xiaodong Liu and Xiaohan Wang and Xi-Cheng Shen and Xiaokang Chen and Xiaokang Zhang and Xiaosha Chen and Xiaotao Nie and Xiaowen Sun and Xiaoxiang Wang and Xin Cheng and Xin Liu and Xin Xie and Xingchao Liu and Xingkai Yu and Xinnan Song and Xinxia Shan and Xinyi Zhou and Xinyu Yang and Xinyuan Li and Xuecheng Su and Xuheng Lin and Y. K. Li and Y. Q. Wang and Y. X. Wei and Y. X. Zhu and Yang Zhang and Yanhong Xu and Yanping Huang and Yao Li and Yao Zhao and Yaofeng Sun and Yao Li and Yaohui Wang and Yi Yu and Yi Zheng and Yichao Zhang and Yifan Shi and Yi Xiong and Ying He and Ying Tang and Yishi Piao and Yisong Wang and Yixuan Tan and Yi-Bing Ma and Yiyuan Liu and Yongqiang Guo and Yu Wu and Yuan Ou and Yuchen Zhu and Yuduan Wang and Yue Gong and Yuheng Zou and Yujia He and Yukun Zha and Yunfan Xiong and Yunxiang Ma and Yuting Yan and Yu-Wei Luo and Yu-mei You and Yuxuan Liu and Yuyang Zhou and Z. F. Wu and Zehui Ren and Zehui Ren and Zhangli Sha and Zhe Fu and Zhean Xu and Zhen Huang and Zhen Zhang and Zhenda Xie and Zhen-guo Zhang and Zhewen Hao and Zhibin Gou and Zhicheng Ma and Zhigang Yan and Zhihong Shao and Zhipeng Xu and Zhiyu Wu and Zhongyu Zhang and Zhuoshu Li and Zihui Gu and Zijia Zhu and Zijun Liu and Zi-Long Li and Ziwei Xie and Ziyang Song and Ziyi Gao and Zizheng Pan},
  journal={ArXiv},
  year={2024},
  volume={abs/2412.19437},
  url={https://api.semanticscholar.org/CorpusID:275118643}
}

@misc{hao2023pinnaclecomprehensivebenchmarkphysicsinformed,
      title={PINNacle: A Comprehensive Benchmark of Physics-Informed Neural Networks for Solving PDEs}, 
      author={Zhongkai Hao and Jiachen Yao and Chang Su and Hang Su and Ziao Wang and Fanzhi Lu and Zeyu Xia and Yichi Zhang and Songming Liu and Lu Lu and Jun Zhu},
      year={2023},
      eprint={2306.08827},
      archivePrefix={arXiv},
      primaryClass={cs.LG},
      url={https://arxiv.org/abs/2306.08827}, 
}

@article{Luo2025PhysicsinformedNN,
  title={Physics-informed neural networks for PDE problems: a comprehensive review},
  author={Kuang Luo and Jing-Shang Zhao and Ying-Ping Wang and Jiayao Li and Junjie Wen and Jiong Liang and Henry Soekmadji and Shao-Lin Liao},
  journal={Artificial Intelligence Review},
  year={2025},
  volume={58},
  url={https://api.semanticscholar.org/CorpusID:280401971}
}

@inproceedings{DeepSeekAI2026DeepSeekV4TH,
  title={DeepSeek-V4: Towards Highly Efficient Million-Token Context Intelligence},
  author={DeepSeek-AI and Anyi Xu and Bang Lin and Bing Xue and Bing-Li Wang and Bin Xu and Bo Wu and Bowei Zhang and Chao Lin and Chengyao Dong and Chen Ling and Cheng-Da Lu and Chen Zhao and Chengqi Deng and Cheng Hou and Chen Xu and Chen-Ze Shao and Chong Ruan and Conner Sun and Da-Mai Dai and Daya Guo and De-Bin Yang and Deli Chen and Dong-Hui Li and Dong-Li Ji and Erhang Li and Fang-Chen Wei and Fangyun Lin and Fang Yuan and Fei Xia and Fucong Dai and Guangbo Hao and Guanting Chen and Guo Cao and Guo-Hui Meng and Guowei Li and Han Yu and Han Zhang and Hanwei Xu and Hao Li and Hao Liang and Haoling Zhang and Haoming Luo and Haoran Wei and Hao Yuan and Haowei Zhang and Hao-Wen Luo and Hao Chen and Haozhe Ji and Heng Zhang and Honghui Ding and Hongxuan Tang and Huanqi Cao and Hua-Zuo Gao and Hui Qu and Hui Zeng and J. Yang and J.-Q. Zhu and Jianming Luo and Jia Song and Jia Yu and Jialiang Huang and Jia-Lu Cai and Jian Liang and Jiangting Zhou and Jiasheng Ye and Jia-Shi Li and Jiaxin Xu and Jiewen Hu and Jieyu Yang and Jin Chen and Jin Yan and Jing-Chang Chen and Jing Zhou and Jing Xiang and Jingyang Yuan and Jing Cheng and Jing-Zi Zhou and Jinhua Zhu and Jiping Yu and Joseph Sun and Junliang Ran and Jun Jiang and Jun-Jie Qiu and Jun-Long Li and Junming Zheng and Jun-Mei Song and Kai Dong and Kaige Gao and Kang Guan and Kexing Zhou and Kezhao Huang and Kuai Yu and Lean Wang and Lecong Zhang and Lei Wang and Leyi Xia and Li Zhang and Liang Zhao and Lihua Guo and Lin Luo and Lin Ma and Linyan Zhu and Li-Tong Wang and Li-Yu Cai and Liyue Zhang and Long-Hao Chen and Mingxi Di and My Xu and Max Mei and Miaojun Wang and Mingchuan Zhang and Minghua Zhang and Minghui Tang and Mingming Li and Mingxu Zhou and Min Han and Ning Wang and Pan Huang and Pan Wang and Peixin Cong and Peiyi Wang and Peng Zhang and Qiancheng Wang and Qihao Zhu and Qing-Yang Li and Qinyu Chen and Qiushi Du and Qi Jiang and Rui Tian and Rui-qin Xu and Rui-Jie Lu and Ruiling Xu and Ruiqi Ge and Rui-Song Zhang and Ruizhe Pan and Runji Wang and Runqian Chen and Runqiu Yin and Runxin Xu and Ruo-Han Shen and Ruoyu Zhang and Ruyi Chen and Sh. L. Liu and Shanghao Lu and Shangmian Sun and Shang-Yan Zhou and Shan-Shan Chen and Shaofei Cai and Shao Nie and Shao-Ping Wu and Shao-Yuan Chen and Shengding Hu and Sheng Liu and Shiqiang Hu and Shi-Rong Ma and Shiyu Wang and Shuiping Yu and Shunfeng Zhou and Shuting Pan and Shuying Yu and Songyang Zhou and Tao Ni and Tao Yun and Tian Jin and Tian-Hong Pei and Tian Ye and Tian-Le Lin and Tian Ji and Tianyi Cui and Tianyuan Yue and Tingting Yu and Tu-Tu Wang and W. Y. Zhang and WL Xiao and Wangding Zeng and Wei An and Weilin Zhao and Wen Liu and Wen-Feng Liang and Wenjie Pang and Wenjing Luo and Wenjin Yao and Wenjun Gao and Wenkai Yang and Wenlve Huang and Wenqing Hou and Wentao Zhang and Wen-Ting Ma and Xi Gao and Xiang He and Xiangwen Wang and Xianzu Wang and Xiao Bi and Xiaodong Liu and Xiaohan Wang and Xiaokang Chen and Xiaokang Zhang and Xiaotao Nie and Xiaowen Sun and Xiaoxiang Wang and Xin Cheng and Xin Liu and Xin Xie and Xingchao Liu and Xing-Chen Liu and Xing-Kai Yu and Xing-You Li and Xin-Yu Yang and Xinyu Zhang and Xu Chen and Xuanyu Wang and Xuecheng Su and Xue-Yin Chen and Xuheng Lin and Xu Fu and YC Yan and YQ Wang and Yw Ma and Yanfen Luo and Yang Zhang and Yan-Hong Xu and Yanru Ma and Yanwen Huang and Yao Li and Yao Xu and Yao Zhao and Yao-Feng Sun and Yao-Hui Wang and Yi Qian and Yingxia Shao and Yi Yu and Yichao Zhang and Yifan Ding and Yifan Shi and Yi-Jia Wu and Yi-Yu Xiong and Yi Ma and Ying He and Ying Tang and Ying Zhou and Yingjia Luo and Yinmin Zhong and Yishi Piao and Yisong Wang and Yixiang Zhang and Yixiao Chen and Yi-Xuan Tan and Yixuan Wei and Yiyang Ma and Yiyuan Liu and Yong-Lu Yang and Yong-Qiang Guo and Yong-Tong Wu and Yu Wu and Yukun Li and Yuan Cheng and Yuan Ou and Yuanfan Xu and Yuanhao Li and Yuduan Wang and Yuehan Yang and Yue Xu and Yuhan Wu and Yu Meng and Yu Zou and Yukun Zha and Yunfan Xiong and Yu-Peng Chen and Yuping Lin and Yu Cao and Yuqian Wang and Yushun Zhang and Yu-Ting Yan and Yu-Tong Lin and Yu-Xian Gu and Yu-Wei Luo and Yu-Mei You and Yuxuan Liu and Yuxuan Zhou and Yuyang Zhou and Yuzhen Huang and Zhanghua Wu and Ze Wang and Ze Zhao and Zehui Ren and Ze-Kai Zhang and Zhangli Sha and Zhe Fu and Zhe Ju and Zhean Xu and Zhen-Da Xie and Zhengyan Zhang and Zhe Gao and Zhewen Hao and Zhibin Gou and Zhicheng Ma and Zhigang Yan and Zhihong Shao and Zhi-Xiang Huang and Zhi-Xuan Chen and Zhiyu Wu and Zhizhou Ren and Zhongyu Wu and Zhuoshu Li and Zhuping Zhang and Zian Xu and Zihao Wang and Zihua Qu and Zihui Gu and Zi-jia Zhu and Zi-Long Li and Zipeng Zhang and Zi-Wei Xie and Ziyi Gao and Zi-Yi Wan and Zi-Zheng Pan and Zong-xiang Yao},
  year={2026},
  url={https://api.semanticscholar.org/CorpusID:289623513}
}

@String(arxiv = {arXiv})

@String(ACL = {Annual Meeting of the Association for Computational Linguistics (ACL)})

@String(AAAI = {AAAI Conference on Artificial Intelligence (AAAI)})

@String(AI = {Artificial Intelligence (AI)})

@String(ACL = {ACL})

@String(AAAI = {AAAI})

@String(AI = {AI})

\newpage
\appendix


\section{DeepSurvey System Supplement}
\subsection{Code Analysis Subsystem}\label{app:code_analysis}
Beyond paper text, DeepSurvey analyzes open-source code repositories linked in collected papers to supplement implementation details not fully described in the text.

\paragraph{Pseudocode generation.}
The code analysis subsystem downloads repositories and produces repository-level pseudocode through a multi-agent loop centered on a planning agent. At each step, the planner selects one of five operations: \textsc{get\_source\_code} reads a specific file from the repository; \textsc{create} generates initial pseudocode from retrieved context; \textsc{revise} refines the pseudocode based on reviewer feedback; \textsc{review} invokes a reviewer that scores the current pseudocode on conciseness, logical structure, and implementation specificity (each 0--10) with concrete improvement suggestions; and \textsc{finish} terminates the loop. The planner is required to read 3--5 key source files (entry points, core algorithm modules, key utilities) before creating pseudocode, and to call \textsc{revise} at least once every 3 rounds. For the integrity of the plan, the planner gives several steps (a sub-plan) at a time rather than one step. A memory mechanism tracks operation history and review feedback across steps, with automatic compression when token count exceeds a threshold to prevent context overflow.

\paragraph{Report generation.}
The system produces two types of reports. \textbf{Code reports} are generated by processing pseudocode in batches (batch size 5): each batch is analyzed along dimensions including problem modeling, core algorithm classification, optimization strategies, and topic-specific dimensions (e.g., agent architectures for LLM systems); batch reports are then integrated into a final report, with multi-round fallback when combined content exceeds the context window. The prompt enforces traceability---each claim must reference specific pseudocode components. \textbf{Environment reports} extract configuration files (requirements.txt, README) from repositories, analyzing framework choices, dependency versions, and deployment patterns. Both reports use semi-open prompts that mandate coverage of core dimensions (algorithm classification, data modeling, engineering optimizations) while allowing the model to identify topic-specific insights.

\paragraph{Exception Handling.} 
There are always extreme cases in a large number of code repositories, and DeepSurvey is capable of handling exceptions. For \textbf{large repositories}, the agent progressively explores the repo structure and reads core files step by step, which handles large repos as long as their size stays within the agent's analysis capacity and we additionally apply filtering (e.g., excluding cache directories, outputs, and build artifacts). For extreme cases, we truncate the repo structure. For \textbf{multi-repo projects}, all open-source repos are analyzed individually and then jointly processed in the second stage. Users can configure the system to analyze only the first \textbf{k} repos; in our experiment, we read all the provided repos. For \textbf{missing/obsolete repos}, repos with missing files or inaccessible URLs are filtered out with tools before analysis.

\subsection{Analysis Knowledge Base}\label{app:knowledge_base}
The analysis knowledge base serves as the central artifact connecting DeepSurvey's understanding and generation stages. Rather than treating survey generation as a one-shot text production task, DeepSurvey first constructs a structured, reusable knowledge repository from the collected literature, then grounds all subsequent writing and refinement in this repository. This design separates \emph{analysis} from \emph{generation}, enabling deeper reasoning without overwhelming the writing model.

The knowledge base is organized into three hierarchical levels, each building on the previous:

\begin{itemize}
    \item \textbf{Paper level}: Full-text \emph{keynotes} recording core mechanisms, experimental settings, key assumptions, applicable ranges, and limitations for each paper. Additionally, for papers with linked code repositories, \emph{code reports} and \emph{environment reports} are generated to capture implementation-level details such as module decomposition, execution flow, engineering dependencies, and runtime configurations---information that is typically absent from paper text.
    \item \textbf{Cluster level}: Papers are organized into thematic clusters, each annotated with three complementary analysis views: \emph{relation graphs} that model typed citation relationships (foundation, extension, substitution) with abstracted descriptions; \emph{comparison tables} that conduct structured cross-paper comparison along key technical dimensions; and \emph{guided Q\&A analyses} that integrate evidence to answer high-value research questions within and across clusters.
    \item \textbf{Survey level}: Hierarchical outlines, section drafts, and revision plans that directly serve the generation process.
\end{itemize}

These representations are linked through explicit mappings: paper keynotes reference their cluster assignments, cluster analyses cite specific papers, and section drafts trace their evidence to both paper keynotes and cluster analyses. This traceability supports the verification mechanisms throughout the pipeline and enables the iterative refinement subsystem to identify and correct inconsistencies.

The knowledge base is designed with two complementary objectives. First, it serves as the evidential foundation for survey generation: the outline-driven writing module assigns papers to sections based on cluster analysis results, and the refinement module retrieves keynotes and code reports to verify and deepen individual paragraphs. By grounding generation in structured intermediate representations rather than raw retrieval results, the system maintains content depth while constraining hallucination. Second, the knowledge base constitutes an independently valuable research artifact. The paper-level keynotes, cluster-level relation graphs, and comparison tables provide fine-grained, traceable insights into a research landscape---supporting tasks such as literature reviews, method comparisons, and research direction analysis beyond the immediate scope of survey writing. This dual utility reflects our design philosophy: the analysis knowledge base is not merely a byproduct of survey generation, but a reusable substrate for scientific inquiry.

\subsection{Engineering Details}\label{app:engineering}

To support large-scale processing, DeepSurvey incorporates several engineering optimizations.

\paragraph{Multi-domain support.} DeepSurvey uses fallback mechanisms to improve robustness across domains. When Semantic Scholar returns insufficient relevant papers or fails during retrieval, the system invokes the arXiv API to supplement the evidence pool. When full texts are unavailable, which is a common phenomenon in certain fields, DeepSurvey falls back to abstracts or TLDRs as lightweight substitutes for full-text keynotes, preserving basic paper coverage while allowing downstream analysis to proceed.

\paragraph{Unified paper identifier.} DeepSurvey retrieves papers from multiple sources (Semantic Scholar, arXiv), each with its own identifier system. To ensure consistent caching and deduplication across the pipeline, we construct a unified \texttt{paper\_id} for each paper: if an arXiv ID is available, it serves as the canonical identifier; otherwise, the Semantic Scholar ID is used as a fallback. All downstream caches (keynotes, relation graphs, analysis results) are hash-keyed by this unified \texttt{paper\_id}, enabling cross-session reuse and preventing redundant processing of the same paper retrieved from different sources.

\paragraph{Robust paper downloading.} Paper downloads support streaming writes and resumable transfer through HTTP range requests. After downloading, PDF files are validated with \texttt{pypdfium2}; corrupted or incomplete files are removed and retried to ensure reliable downstream parsing.

\paragraph{Multi-level caching.} A three-tier caching system minimizes redundant computation and API costs: (1)~\emph{API-level caching} stores raw responses from external services (Semantic Scholar, LLMs, embedding models) keyed by request parameters, with configurable expiration; (2)~\emph{Task-level caching} persists results of major pipeline stages (literature collection, keynotes, clustering, analysis) as JSON files for session-level reuse; (3)~\emph{Runtime caching} maintains in-memory results within pipeline stages to eliminate intra-stage redundancy. Task-level caching is implemented via \texttt{diskcache}, enabling cross-session reuse of expensive computations such as full-text keynotes and relation graphs.

\paragraph{Parallel processing.} Independent operations execute concurrently via \texttt{ThreadPoolExecutor}: keynote generation for different papers, comparison table construction across clusters, and batch LLM API calls all proceed in parallel. Thread pool sizes are configured per task type to balance throughput against API rate limits. Each request includes timeout control to prevent individual failures from blocking the pipeline. A checkpoint mechanism enables resuming interrupted pipelines from the last completed stage rather than restarting from scratch.

\paragraph{Error recovery.} LLM API calls are wrapped with automatic retry via the \texttt{tenacity} library, using exponential backoff (1--300\,s) for up to 10 attempts on recoverable HTTP errors (408, 429, 500, 502, 503, 504). At the pipeline level, a structured error memory accumulates recent failure messages (capped at 10 entries with deduplication) and injects them into subsequent prompts, enabling the model to adapt its output based on prior failure patterns. This mechanism is particularly effective in multi-round tasks such as outline generation and citation assignment.

\paragraph{Document parsing.} Full-text analysis relies on MinerU~\citep{Wang2024MinerUAO} to convert PDFs into structured Markdown while preserving layout, reading order, and mathematical notation. Each paper is parsed into an independent output directory containing the structured text and intermediate layout annotations, which serve as input to the full-text understanding module.

\paragraph{Context window management.} Given LLM context limits, DeepSurvey employs adaptive input construction: token counts are estimated via \texttt{tiktoken} (falling back to the \texttt{cl100k\_base} encoder), and inputs exceeding the model's window are progressively compressed---first by trimming retrieval-augmented context (halved iteratively), then by truncating paper content while preserving core keynotes and outline structure. In the citation assignment stage, this progressive reduction ensures that essential paper--outline mappings are retained even under tight context constraints.

\paragraph{Device adaptation.} The embedding model uses lazy loading with automatic GPU-to-CPU fallback on out-of-memory errors. Batch encoding sizes are configurable to balance throughput against memory footprint during large-scale similarity computation.

\paragraph{Agent memory management.} For multi-step agentic tasks, DeepSurvey stores operation history, intermediate outputs, review feedback, and revision states in structured memory. When the token budget is exceeded, older long-context entries are compressed or truncated while recent actions and a compact state summary are preserved for subsequent planning.

\subsection{Reliability Check}\label{app:reliability}
Although DeepSurvey's citation reliability significantly surpassed baselines, citation issues still occur at a relatively low frequency. Since our motivation is to develop a research tool for both researchers and agents, we expect DeepSurvey's outcome readers to themselves be DeepSurvey users, who can judge survey quality by combining DSEval results with their own knowledge.

We also plan to release an annotation tool as part of our open-source repository that examines citation reliability for each claim based on DSEval's per-claim citation evaluation method and flags suspicious claims and citations, helping both human and AI users use DeepSurvey more reliably.

We do not currently encourage any publication via DeepSurvey, or sharing its output with others without disclosing that it is not human-written.

\section{Evaluation Protocol Validation Supplement}\label{app:eval_validation}
\subsection{AutoSurvey Evaluation Experiment}\label{app:eval_autosurvey}

AutoSurvey~\citep{Wang2024AutoSurveyLL} established the dominant evaluation paradigm for automated survey generation, assessing content quality across three dimensions---Coverage, Structure, and Relevance---on a 1--5 scale using LLM-as-judge. Its evaluation framework has been widely adopted by subsequent work~\citep{Wu2025AutoSurvey2ER, Wang2025LLMtimesMapReduceV2EC, Liang2025SurveyXAS, Yan2025SurveyForgeOT, Go2025LiRAAM, Shi2025SciSageAM}, and its original meta-evaluation demonstrated moderate positive correlation with human judgments (Spearman's $\rho$ up to 0.5429), confirming its validity at the time.

To assess whether this benchmark remains discriminative under current model capabilities, we replicate AutoSurvey's evaluation protocol exactly---using its original dimensions, prompts, and scoring criteria---to evaluate AutoSurvey and DeepSurvey. As shown in Table~\ref{tab:coverage_structure_relevance_main_tex}, both methods achieve the maximum score (5) across all dimensions, rendering the benchmark unable to differentiate between approaches.

\begin{table}[t]
\centering
\small
\setlength{\tabcolsep}{3pt}
\caption{Scores under AutoSurvey's original evaluation protocol. Both AutoSurvey and DeepSurvey saturate at the maximum score (5), indicating insufficient discriminative power.}
\label{tab:coverage_structure_relevance_main_tex}
\begin{tabular}{lccc}
\toprule
Method & Coverage & Structure & Relevance \\
\midrule
AutoSurvey      & 5 & 5 & 5 \\
DeepSurvey      & 5 & 5 & 5 \\
\bottomrule
\end{tabular}
\end{table}

This saturation arises because AutoSurvey's 5-point scale and coarse-grained dimensions were designed for an earlier generation of models. As LLM capabilities have advanced, state-of-the-art systems---regardless of their architectural differences---now consistently meet the ceiling criteria for basic coverage, structural coherence, and topic relevance. 


\subsection{LLM-Judge Stability Experiment}\label{app:eval_stability}

LLM-as-judge has emerged as a scalable alternative to human evaluation for open-ended text quality assessment. Prior work has shown that strong judge models (e.g., GPT-4) achieve over 80\% agreement with human preferences~\citep{Zheng2023JudgingLW}, and frameworks such as G-Eval~\citep{Liu2023GEvalNE} demonstrate reliable human correlation on summarization tasks. However, LLM judges are not inherently stable---JudgeLM~\citep{Zhu2023JudgeLMFL} and subsequent studies on position bias~\citep{Kim2024Prometheus2A} show that scoring reliability is sensitive to prompt format, answer ordering, and task configuration. It is therefore necessary to explicitly validate judge stability before relying on LLM scores for cross-method comparison.

We evaluate four judge configurations along two axes: \textbf{model} (GPT series vs.\ MiMo series) and \textbf{prompt style} (with explanation, \texttt{-exp}, vs.\ without explanation, \texttt{-no-exp}). For each configuration, we score the same set of generated surveys multiple times and compute per-sample standard deviation, coefficient of variation (CV\%), maximum absolute deviation, and pairwise maximum difference. Aggregated results are reported in Table~\ref{tab:stability_stats_main_tex}; fine-grained breakdowns by metric and method are provided in Table~\ref{tab:stability_full_stats}. Note that we do not perform a separate judge-stability study for the NLI-based citation evaluation because prior work has shown that NLI remains informative for LLM evaluation and is widely used for fact verification and hallucination detection~\citep{Manakul2023SelfCheckGPTZB, Madaan2024LostII, chen-etal-2025-explainable, Wang2024AutoSurveyLL}.

\begin{table}[H]
\centering
\small
\setlength{\tabcolsep}{3pt}
\caption{Aggregated stability statistics across four judge configurations. ``-exp'' indicates that the LLM must give a score with an explanation, while ``-no-exp'' indicates the reverse.}
\label{tab:stability_stats_main_tex}
\begin{tabular}{lcccc}
\toprule
Configuration & std & CV\% & max\_abs\_dev & range \\
\midrule
GPT-exp     & 0.041  & 0.49  & 0.044 & 0.081 \\
GPT-no-exp  & 0.023  & 0.279 & 0.024 & 0.046 \\
MiMo-exp    & 0.077  & 1.03  & 0.078 & 0.152 \\
MiMo-no-exp & \textbf{0.020}  & \textbf{0.244} & \textbf{0.022} & \textbf{0.035} \\
\bottomrule
\end{tabular}
\end{table}

\textbf{Results.} The \texttt{mimo-no-exp} configuration achieves the best stability across all four metrics (std = 0.020, CV = 0.244\%, max\_abs\_dev = 0.022, pairwise\_max\_diff = 0.035), representing absolute fluctuations below 5\%---well within acceptable bounds for cross-method ranking. Compared to \texttt{gpt-no-exp}, \texttt{mimo-no-exp} reduces standard deviation by 13.0\% and CV by 12.5\%. The improvement over \texttt{mimo-exp} is even more pronounced: std, range, CV, and max\_abs\_dev decrease by 74.0\%, 77.0\%, 76.3\%, and 71.8\% respectively (range and pairwise\_max\_diff refer to the same statistic), indicating that requiring the judge to produce explanations introduces substantial additional variance that undermines scoring consistency.

\textbf{Choice of MiMo-series model as judge model.} We select \texttt{mimo-no-exp} (MiMo models without explanation prompts) as the primary judge for our evaluation setup for two reasons. First, it offers the highest scoring stability among all candidates, ensuring that observed differences between methods reflect genuine quality distinctions rather than judge noise. Second, MiMo is an open-source series model with relatively low inference cost, enabling reproducible and scalable evaluation without dependence on proprietary APIs. The detailed per-metric stability breakdowns in Table~\ref{tab:stability_full_stats} further confirm that \texttt{mimo-no-exp} maintains consistently low variance across different methods (AutoSurvey, DeepSurvey) and evaluation dimensions (Core Quality, Writing Quality, Content Depth), validating its suitability as a unified evaluation standard.





\subsection{Consistency Between LLM Judge And Human Experts Experiment}\label{app:eval_coherence}
We report agreement at both the pairwise and group levels. Pairwise agreement is measured with Cohen's $\kappa$, and panel-level consistency is measured with Fleiss' $\kappa$. Since our labels are nominal ($A$, $B$, and Tie), every item has a fixed number of annotations, and there are no missing judgments, nominal Krippendorff's $\alpha$ and Fleiss' $\kappa$ are numerically very close in our setting, with the difference arising only from their finite-sample treatment of chance agreement. We therefore report Fleiss' $\kappa$ as the group-level metric to avoid redundant notation.

\paragraph{Cohen's $\kappa.$}
For two annotators, Cohen's $\kappa$ is defined as
\begin{equation}
\kappa = \frac{p_o - p_e}{1 - p_e},
\end{equation}
where $p_o$ is the observed agreement and $p_e$ is the chance agreement implied by the annotators' marginal label distributions. We compute Cohen's $\kappa$ between the LLM judge and each human expert, and between pairs of human experts, then average over all valid pairs.

\paragraph{Fleiss' $\kappa.$}
For multi-rater agreement, let $n_{ic}$ be the number of annotators who assign category $c$ to item $i$, and let $n$ be the number of annotators per item. Fleiss' $\kappa$ is computed as
\begin{align}
P_i &= \frac{1}{n(n-1)} \sum_c n_{ic}(n_{ic}-1), \\
\bar{P} &= \frac{1}{N}\sum_{i=1}^{N} P_i, \\
p_c &= \frac{1}{Nn}\sum_{i=1}^{N} n_{ic}, \\
P_e &= \sum_c p_c^2, \\
\kappa &= \frac{\bar{P} - P_e}{1 - P_e}.
\end{align}
Here, $P_i$ measures agreement on item $i$, $\bar{P}$ is the average observed agreement, and $P_e$ is the agreement expected by chance from the marginal label frequencies.

\paragraph{Evaluation settings.}
We evaluate several group configurations to separate human consistency, LLM contribution, and random baselines:
\begin{itemize}
    \item \textbf{Human full.} All human annotators are included. This setting measures the intrinsic consistency of the human panel and serves as the main human reference.

    \item \textbf{Human leave-one-out.} We remove one human annotator at a time and compute group agreement on the remaining human panel. This setting estimates the natural variation of the human panel under a one-rater perturbation.

    \item \textbf{Human leave-one-out + actual LLM judge.} We replace the removed human with one LLM judge and recompute group agreement. This is the main test of whether the LLM can substitute for one human without degrading panel consistency.

    \item \textbf{Human leave-one-out + random human judge.} We replace the removed human with another human judge sampled from the human pool. This setting provides an empirical human-replacement baseline and approximates the agreement level expected when the missing annotator is replaced by a genuine human.

    \item \textbf{Human leave-one-out + random evaluator.} We replace the removed human with a random evaluator that assigns $A$, $B$, or Tie with equal probability $1/3$. This is a null baseline that estimates the agreement expected from an uninformative judge.

    \item \textbf{LLM full.} All LLM judges are included. This setting measures the internal consistency of the LLM panel and serves as a reference for LLM-to-LLM agreement.
\end{itemize}

\begin{figure}[!htp]
  \centering
  \includegraphics[width=0.8\linewidth]{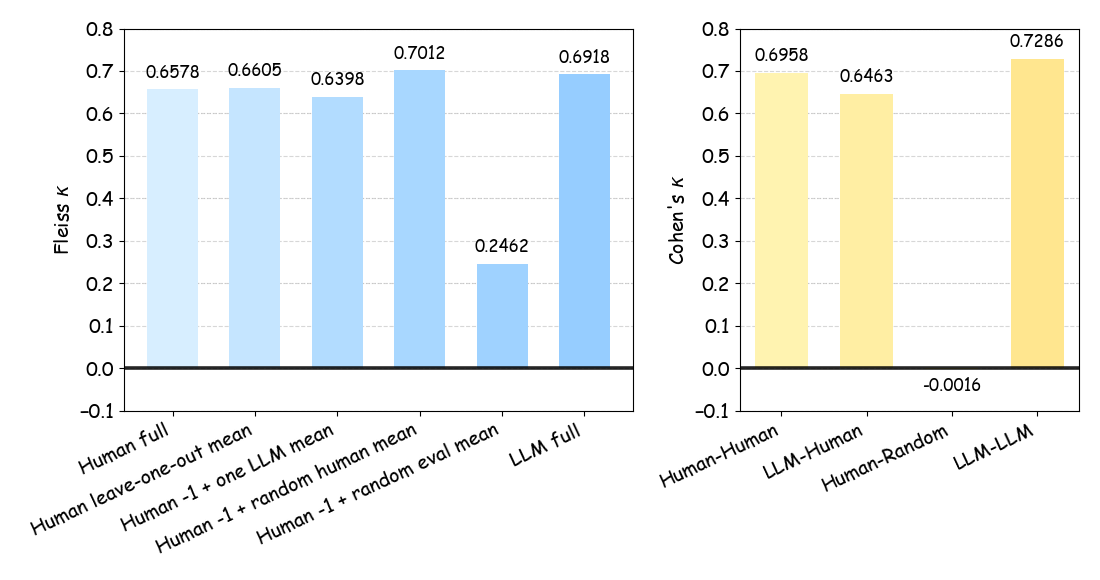}
  \caption{Agreement between human and LLM-judge. The left panel shows Fleiss' $\kappa$ for compositions including full humans, leave-one-human-out, replacing one human with an LLM, and random-replacement baselines. The right panel shows pairwise Cohen's $\kappa$ between human, LLM, and random evaluator groups.}
  \label{fig:human_llm_coherence}
\end{figure}

\paragraph{Results.} The results are shown in Figure~\ref{fig:human_llm_coherence}. The pairwise results show that the LLM judge is close to human experts in fine-grained agreement. Its mean Cohen's $\kappa$ of $0.6463$ is only $0.0495$ below the human--human agreement of $0.6958$, while remaining far above the human--random baseline ($-0.0016$). This gap indicates that the LLM judge captures human preference patterns rather than behaving like a random scorer.

The group-level results lead to the same conclusion. Human-1 + LLM achieves a Fleiss' $\kappa$ of $0.6398$, which is close to the all-human panel ($0.6578$). In contrast, human-1 + random drops sharply to $0.2462$, showing that the gain from the LLM is not explained by a generic additional label source. The difference between human-1 + LLM and human-1 + random is large enough to suggest that the LLM contributes structured, human-aligned signal to the panel.

Overall, these numbers support two claims. First, the LLM judge is well aligned with human experts at the pairwise level. Second, when inserted into a multi-rater panel, it preserves overall consensus at a level comparable to human annotators. This makes the LLM judge a stable and credible component of the evaluation protocol.

\section{Downstream Validation with a Coding Research Agent}\label{app:downstream_agent}

\subsection{Setting.}
To complement rubric-based scoring with a direct behavioral measure, we examine whether a survey improves the research capability of a downstream agent on PINNacle~\citep{hao2023pinnaclecomprehensivebenchmarkphysicsinformed}, a benchmark for physics-informed neural networks (PINNs). We strip all published PINN optimization methods from the benchmark, leaving only the vanilla baseline (a tanh fully-connected network trained with Adam on the standard losses), so that methodological knowledge must originate from the agent's prior knowledge or the provided survey. We compare four conditions: \emph{none} (no literature input), \emph{human} (a human-written survey), \emph{SurveyForge} (a survey generated by the strongest baseline system), and \emph{DeepSurvey} (a survey generated by our system). For the human condition we select one of the most highly cited human-written surveys on PINNs published within the last two years~\citep{Luo2025PhysicsinformedNN}, ensuring that this condition provides sufficiently recent methodological knowledge as well as authoritative and reliable expert analysis.

\subsection{Case selection.}
PINNacle spans regimes on which, as its own evaluation shows, no single published optimization methodology succeeds across all cases. Designing and validating a methodology effective on every case within one fixed wall-clock session (480min) is therefore infeasible by construction: forced to cover the full suite, the agent could not complete even a single design--implement--validate loop, and all conditions would collapse to a floor effect. We thus fix a four-case subset spanning three canonical PDE families: nonlinear advection--diffusion (Burgers-1D-C), parabolic diffusion (Heat-2D-CG, Heat-2D-MS), and hyperbolic propagation (Wave-1D-C), forming a graded difficulty profile in terms of the vanilla baseline: nearly solved on Burgers-1D-C (relative L2 error $\sim 10^{-2}$), moderately difficult on the Heat cases, and severely failed on Wave-1D-C ($\sim 6\times 10^{-1}$). This gradient allows us to examine whether the effect of survey quality scales with how much methodological knowledge a case demands. The case set was fixed before any agent runs, and identical cases are used across all conditions.

\subsection{Session protocol.}
In each run, a coding agent (codex\footnote{\url{https://openai.com/codex/}}) backed by DeepSeek-V4-Flash~\citep{DeepSeekAI2026DeepSeekV4TH} operates in an isolated copy of the workspace and receives exactly one literature input (a DeepSurvey-generated survey, a SurveyForge-generated survey, a human-written survey, or none), together with the locked evaluation protocol and an anonymized leaderboard of published results that is identical across conditions; SurveyForge serves as the strongest baseline, ranking second to DeepSurvey in both total and depth scores. The agent backbone is deliberately chosen from a model family different from that used to generate the surveys (MiMo-V2.5), ruling out self-preference. The leaderboard contains reference error values only, with method identities anonymized so that the agent cannot infer superior approaches from algorithm names; it indicates attainable accuracy. The agent must autonomously design, implement, and validate a PINN training methodology within a fixed wall-clock session, writing only to a dedicated directory with predefined hooks (loss formulation, architecture, loss weighting, optimizer, and training loop); the harness then re-trains the delivered implementation from scratch on the enabled PDE cases at the benchmark's original protocol (20,000 iterations per case), using evaluation seeds disjoint from those exposed through the workspace's development interface to prevent seed overfitting, and reports the mean relative L2 error per case. We deliberately adopt codex rather than a sophisticated auto-research framework: more complex or less constrained systems introduce additional confounding variables, and on a problem of moderate difficulty an overly capable research agent could succeed independently of the provided survey, robbing the experiment of its ability to attribute outcomes to survey quality. Beyond the relative $L_2$ error metric, we additionally score the novelty of each delivered methodology with GPT-6 Astra\footnote{\url{https://openai.com/index/gpt-6-astra/}}: as one of the strongest LLMs, with pretraining knowledge recent enough to cover the current PINN literature, it can judge comparatively objectively whether a methodology departs from established techniques, a quality that the $L_2$ error metric alone cannot capture. The complete scoring rubric is shown in Table~\ref{tab:novelty_rubric}.

\begin{table}[h]
\centering
\small
\caption{Scoring rubric for the technical novelty of delivered methodologies (judged by GPT-6 Astra).}
\label{tab:novelty_rubric}
\begin{tabular}{c p{0.85\textwidth}}
\toprule
\textbf{Score} & \textbf{Standard} \\
\midrule
1--2 & Essentially identical to an existing published method. \\
3--4 & Minor modification or straightforward combination of existing methods; limited new technical content. \\
5--6 & Noticeable methodological novelty; the idea may follow or extend prior work, but adds meaningful new technical content beyond it. \\
7--8 & Clearly novel core mechanism or formulation; no known published work presents essentially the same method. \\
9--10 & Highly original core idea that introduces a substantially new methodological principle or paradigm. \\
\bottomrule
\end{tabular}
\end{table}

\subsection{Fairness controls.}
Three controls ensure fairness and elicit genuine research behavior. First, isolation is dual-enforced: the sandbox blocks network access and out-of-workspace reads, all evaluation code is hash-locked and statically checked, and post-session transcript audits screen for protocol violations, including bypassing learning altogether, such as directly solving the PDE mathematically or delegating to external numerical solvers, which lies outside the permitted modification scope of the predefined hooks; since the reported numbers are re-trained from the pristine codebase, session-time tampering cannot affect the results. Second, a uniform research directive appended to every condition's prompt encourages genuinely novel optimization rather than stacking surveyed components or brute-force parameter search, mirroring realistic auto-research. Third, the session budget is wall-clock rather than token-based, as token spending is confounded by survey length; the budget is sized on the order of a full validation pass of the benchmark, which renders exhaustive re-evaluation loops infeasible and compels the agent to rely on insights distilled from the survey rather than repeated trial-and-error.

\subsection{Results.}

Table~\ref{tab:pinnacle_downstream} reports the best of three runs per condition, each retrained from scratch at the full 20{,}000-iteration protocol. We report the best run rather than the mean for two reasons: agent sessions occasionally exhibit anomalous failures (e.g., environment errors or early abandonment) whose inclusion would distort the average without reflecting the condition's attainable capability, and in research practice the best attempt within a fixed budget is the quantity of interest, mirroring how benchmark results are reported. The selection rule is applied uniformly across all conditions. PINN-base (pilot) is the vanilla baseline trained by our harness itself under the identical protocol before any agent sessions, and PINNacle SOTA takes, for each case, the best value among the published PINNacle methods (the vanilla column excluded); the account below is grounded in the session transcripts of these runs. Only the DeepSurvey condition improves on the published leaderboard reference: its overall error of 1.53e-02 is $11.2\times$ below PINN-base and $2.5\times$ below the per-case-best reference row (3.81e-02), whereas the other three conditions all remain above the reference.

\paragraph{Headroom analysis.}
The none condition attains the best per-case error only on Burgers-1D-C and Heat-2D-CG, which are exactly the two low-headroom cases: measured as the ratio of the vanilla baseline to the published SOTA, the attainable headroom is merely $1.6\times$ (Burgers-1D-C) and $1.1\times$ (Heat-2D-CG), versus $8.8\times$ (Heat-2D-MS) and $5.0\times$ (Wave-1D-C) on the remaining two. On the two easy cases, the none condition's improvements over the baseline are correspondingly marginal ($1.6\times$ and $1.4\times$) and, per its session report, amount to conservative engineering tweaks. In contrast, DeepSurvey's gains concentrate on the two high-headroom cases, where it improves over the baseline by $12.9\times$ and $51\times$, surpassing the published SOTA on both, and its delivered method (SCRF-PINN) is judged architecturally novel. Improvements that require methodological knowledge therefore track the amount of knowledge a case demands, which is precisely where a survey's value should manifest.


\paragraph{Baseline Results}
The three non-DeepSurvey conditions converged to conservative or unstable methods, and their failure modes are informative. The SurveyForge-condition agent selected precisely the two classical directions highlighted by its survey, namely a hard-constraint ansatz that satisfies boundary conditions by construction and adaptive per-point residual weighting. The combination clearly outperformed the baseline at the short development horizon (${\sim}1{,}000$ iterations) but degraded in late training under the full protocol, leaving the run at approximate parity with the vanilla baseline (1.63e-01 overall). Transcript analysis attributes this to a validation-horizon gap: the survey presents both techniques at the concept level, without implementation-level construction details (e.g., the boundary extension function of a hard-constraint ansatz) or failure-mode analysis, so the agent spent most of its budget rediscovering these details by trial-and-error and reached the session time limit without full-protocol validation. The human-survey condition marks the upper bound of the same limitation. Its survey's diagnosis of PDE/BC loss imbalance (\S2.3.1) gave the agent a correct entry point, and its first action was to measure per-term loss and gradient imbalance; however, the survey's high-level enumeration of remedies led it to stack adaptive gradient reweighting, Fourier features, and residual resampling as published. The adaptive scheme failed systematically in this setting; unable to reconcile the components within budget, the agent retreated to a minimal design: a static boundary-loss weight of 1000 with a 400-step PDE-loss warm-up on three boundary-dominated cases, with all other cases kept on the exact baseline trajectory. This conservatism, together with full-protocol validation during development (unique among these runs), made the result stable (6.48e-02) but capped its novelty and its gains. The no-survey condition lacked even this diagnostic scaffold: with no prior evidence for ranking candidates, it stacked four conventional tricks at once (adaptive loss weighting, causal curriculum, learning-rate scheduling, and L-BFGS finishing), which made component defects hard to attribute; about a quarter of the session went to isolation and rollback, and 62\% of the budget was then spent tuning two hard cases. Its final method reverted to the standard pipeline plus two conservative changes (balanced boundary collocation and a bounded end-of-training L-BFGS polish), beating the baseline on all four cases but by only $1.33\times$. Taken together, these runs suggest that neither parametric knowledge alone nor a survey written for human readers provides sufficient implementation-level understanding of the problem, or scenario-specific guidance on selecting and instantiating mechanisms, for the agents to converge within the session budget on a methodology that is both novel and effective.

\paragraph{DeepSurvey Results}

In contrast, the DeepSurvey survey organizes knowledge mechanism-first. It opens with a unified diagnosis of PINN training failures (flat-minimum traps, spectral bias, fixed-point attraction; \S2.1, \S5.4) and maps each candidate method to the failure mode it targets: the $\sigma$-bandwidth knob that directly modulates spectral bias (\S3.1), the hard-constraint output transform (\S3.2), curriculum regularization (\S2.6), and the closed-loop adaptive principle (\S4), each accompanied by applicability conditions and failure-mode analysis. Every section thus corresponds to a class of actionable interventions rather than a chronological enumeration of prior work. The session transcript shows this organization being translated into design decisions step by step. Within the first two minutes, the agent covered the full mechanism chapters (\S1--\S5) through seven targeted extractions and delivered a first method version after roughly twenty minutes. When this version failed through a runaway feedback loop in its closed-loop controller, the agent did not retreat to trivial modifications; based on the understanding obtained from the survey, it turned to measuring the problem data itself and built the per-axis spectral fingerprint $\sigma_{\mathrm{req}}$, and after diagnosing a plateau-trigger defect and measuring that the anchor function alone already attained a relative L2 error of 0.14, it completed the hard-constraint redesign with zero initialization and a distance-ratio anchor. Two keyword-guided re-reads of the survey (``hard constraint'', ``loss balancing'', ``flat trap'') preceded the reconstruction of the distance-ratio anchor and the drift-timescale learning-rate decay, which the agent's report explicitly attributes to \S2.4. Late in the session, before implementing the Helmholtz gate and the wave reparameterization, the agent searched the survey for ``helmholtz'', ``hyperbolic'', and ``wave equation'', found only two passing mentions and no dedicated technique for hyperbolic problems, and only then invented the symmetric d'Alembert ansatz. This behavior, confirming a coverage gap in the survey before filling it autonomously, is the origin of the Wave-1D-C breakthrough.

The delivered SCRF-PINN is organized around operator-measurement-driven automatic decision-making: a per-axis spectral fingerprint measured from the PDE operator (by probing the operator with quadratic test functions) and from the boundary/initial data, which sizes an anisotropic Fourier-feature layer only where the measured bandwidth requires it; a hard-constraint output transform that enforces Dirichlet conditions exactly, automatically disabled when the measured $k^2$ term identifies a Helmholtz-type operator, for which the harmonic anchor underlying the constraint is misspecified; and an exponential learning-rate schedule timed to the drift timescale. On the four scored cases, the method reaches 1.36e-02 on Burgers-1D-C (1.54$\times$ below the baseline, within 2.3\% of the best result reported on the PINNacle leaderboard, 1.33e-02), 1.18e-02 on Heat-2D-MS ($12.9\times$ below the baseline and below the leaderboard best of 1.73e-02), and 9.54e-03 on Wave-1D-C ($51\times$ below the baseline and an order of magnitude below the leaderboard best of 9.79e-02), while Heat-2D-CG remains at baseline level (2.64e-02). The overall error of 1.53e-02 is the best of all conditions, the Wave-1D-C result is the best across all runs, and the GPT-6 Astra novelty judge scores the delivered methodology markedly higher than the methods delivered under the other three conditions.

The survey's role illustrates the core advantage of DeepSurvey: what it provides is not a menu of methods but a deep understanding of the problem itself. The $\sigma$ knob exists in the survey (\S3.1), yet the quadratic-probe spectral-fingerprint measurement, the plateau-triggered closed loop, and the $k^2$ gating rule are the agent's own constructions with no direct counterpart in the survey text. Reading the survey thus enabled the agent both to reuse appropriate mechanisms selectively and to innovate beyond them. By contrast, the high-level method enumerations available to the human and SurveyForge conditions led to stacking components as published, and the no-survey condition lacked even a diagnostic entry point; after setbacks, all three reverted to trivial variants rather than attempting genuinely novel designs. The value of a DeepSurvey survey therefore lies not only in deeper and more concrete solutions, but in its analysis deepening the agent's understanding of the domain to the point where original mechanisms beyond the survey text become attainable.

\subsection{Qualitative Analysis.}
Session transcripts and deliverable reports yield two qualitative observations that complement the quantitative results. First, in an early protocol iteration that asked only for the highest achievable score, without the innovation directive or the PINN-methodology restriction, the DeepSurvey-condition agent leveraged its understanding of the PINN task to recognize that several benchmark cases admit direct mathematical solution, solved them outside the learning framework, and achieved exceptionally low errors (roughly two orders of magnitude below the baseline, surpassing the best published results in the leaderboard), while agents under the other conditions remained occupied with tuning the parameters of classical optimization mechanisms. This contrast indicates that DeepSurvey equips the agent with a deeper understanding of the research topic, an implementation-level understanding rather than a high-level recall of existing methods, which is what enabled the agent to recognize that the benchmark objective could be attained without learning at all. We excluded this run from all quantitative comparisons as out of scope for the research question, which targets learning-based methodologies, and made this restriction explicit in the upgraded protocol. 

Second, agents under the SurveyForge, human-written, and no-survey conditions consistently fell back to stacking established methods: even under the identical innovation directive, their reports describe starting from classical techniques drawn from the survey or from pretraining knowledge, combining multiple optimizations, failing to reconcile their hyperparameters, and retreating to trivial variants. We do not attribute this contrast to the DeepSurvey-condition agent following the innovation directive while the other agents fail to follow it, since the directive was uniform across conditions; we instead interpret it as a gap in problem understanding, where insufficiently grounded agents cannot move beyond canonical methods, whereas the DeepSurvey-condition agent could attempt a genuinely novel design and attain the best outcome within the same session budget.

\section{Main Evaluation Supplement}\label{app:eval_introduction}
\subsection{Test Dataset}\label{app:eval_introduction_input}
We will present the specific input topics in our evaluation setup in this section:
\begin{itemize}[nosep]
  \item \textbf{Biology}
  \begin{itemize}[nosep]
    \item Microbial Enzymes in Pollutant Bioremediation
    \item Lactic Acid Bacteria and Bacteriocins
    \item Mathematical Models of Malaria
    \item Single-cell RNA Sequencing and Spatial Transcriptomics
    \item Multi-omics Integration for Biological Discovery
  \end{itemize}

  \item \textbf{Medicine}
  \begin{itemize}[nosep]
    \item Dietary Sugars and Body Weight in Randomised Controlled Trials
    \item Flavonoids and Phenolic Compounds from Medicinal Plants
    \item Patient Engagement in Research
    \item Wearable Health Monitoring
    \item Gut Microbiome and Disease
  \end{itemize}

  \item \textbf{Psychology}
  \begin{itemize}[nosep]
    \item Neuroimaging Studies of Internet and Gaming Addiction
    \item Technology-Delivered Interventions for Youth Depression and Anxiety
    \item Social Media and Mental Health
    \item Trauma and PTSD Intervention
    \item Digital Mental Health
  \end{itemize}

  \item \textbf{Computer Science}
  \begin{itemize}[nosep]
    \item Multimodal Machine Learning Taxonomy
    \item LLMs-based Agent
    \item 3D Gaussian Splatting
    \item Hallucination in LLMs
    \item Acceleration for LLMs
  \end{itemize}

  \item \textbf{Materials Science}
  \begin{itemize}[nosep]
    \item Solid-State Batteries
    \item Perovskite Solar Cells
    \item Two-Dimensional Materials and Their Applications in Electronics and Energy
    \item Metal-Organic Frameworks for Energy and Environmental Applications
    \item Advanced Battery Materials for Next-Generation Energy Storage
  \end{itemize}
\end{itemize}

Topics are sourced from SurveyGen's topics with high citation count and some popular topics from Semantic Scholar, balancing influence and recency.

\subsection{Detailed Evaluation Criteria}\label{app:eval_introduction_standard}

This section provides detailed definitions for the content quality sub-dimensions described in the main paper (Section~\ref{ssec:evaluation_metric}). Our evaluation adopts a hierarchical scoring framework where content quality is decomposed into three aggregated dimensions, each comprising fine-grained sub-dimensions scored on a 1--10 scale.

\paragraph{Content Score aggregation.} Each aggregated dimension is computed as the arithmetic mean of its sub-dimension scores:
\begin{equation}
    \text{Core} = \frac{1}{4}(\text{Synthesis} + \text{Organization} + \text{Comprehensiveness} + \text{Relevance})
\end{equation}
\begin{equation}
    \text{Writing} = \frac{1}{3}(\text{Readability} + \text{Academic Rigor} + \text{Clarity \& Coherence})
\end{equation}
\begin{equation}
    \text{Depth} = \frac{1}{4}(\text{Critical Analysis} + \text{Novelty and Insights} + \text{Specificity} + \text{Future Directions})
\end{equation}
The final content quality score combines the three dimensions with weights reflecting our definition of a high-quality survey:
\begin{equation}
    \text{Total} = 0.4 \times \text{Core} + 0.2 \times \text{Writing} + 0.4 \times \text{Depth}
\end{equation}

Each generated survey is independently scored on every sub-dimension, with the average across multiple evaluation rounds reported as the final result to improve stability.

Since a self-designed criterion would probably introduce preference toward DeepSurvey, our evaluation formula is built on the existing Agentic AutoSurvey's protocol\footnote{Since Agentic AutoSurvey's repository have not been open-source, we cannot reproduce it in our main experiment}~\citep{Liu2025AgenticAL} for reasonable and objective evaluation. The final weights were derived by adjusting Agentic AutoSurvey's formula according to our core motivation. Agentic AutoSurvey uses weights of 0.6/0.2/0.2 for Core Quality/Writing Quality/Content Depth. We consider that this formula overemphasizes Core Quality which primarily reflects top-level requirements (like relevance and comprehensiveness) that existing methods (e.g., AutoSurvey~\citep{Wang2024AutoSurveyLL}) already handle reasonably well. Moreover, as surveys in our setting serve as knowledge sources for downstream research agents, the breadth reflected in Core Quality provides relatively limited additional value to such consumers; the representation emphasized by Writing Quality is similarly of secondary importance. To reflect the premise that analytical depth, rather than breadth or representation, constitutes the principal knowledge a survey adds for such consumers, we moderately reduced the Core Quality weight and increased the Content Depth weight, yielding the final 0.4/0.2/0.4 weight.

\paragraph{Weight Robustness.} A potential concern is that our re-weighted formula could bias the comparison in favor of DeepSurvey. To rule this out, we recompute the total score of all methods directly from the fine-grained dimension scores in Table~\ref{tab:all_results} under two alternative weight configurations: Agentic AutoSurvey's original formula (0.6/0.2/0.2) and uniform weights (1/3 each). As shown in Table~\ref{tab:weight_robustness}, DeepSurvey remains the top method under all three configurations, outperforming the strongest baseline SurveyForge by comparable margins.

\begin{table}[H]
\centering
\small
\setlength{\tabcolsep}{3pt}
\caption{Total content scores under three weight configurations over Core Quality/Writing Quality/Content Depth, computed from the fine-grained scores in Table~\ref{tab:all_results}. The ranking is unchanged across configurations, confirming that DeepSurvey's advantage does not depend on our choice of weights.}
\label{tab:weight_robustness}
\begin{tabular}{lccc}
\toprule
Method & Ours & Agentic AutoSurvey & Uniform \\
 & (0.4/0.2/0.4) & (0.6/0.2/0.2) & (1/3, 1/3, 1/3) \\
\midrule
DeepSurvey  & \textbf{8.341} & \textbf{8.535} & \textbf{8.315} \\
SurveyForge & 8.071 & 8.297 & 8.085 \\
AutoSurvey  & 7.958 & 8.252 & 7.990 \\
SurveyX & 6.993 & 7.478 & 7.006 \\
LiRA        & 5.687 & 5.544 & 5.873 \\
\bottomrule
\end{tabular}
\end{table}

Furthermore, in Section~\ref{ssec:meta_eval}, we validated the consistency of our total score formula with human expert preferences. We believe this meta-evaluation substantiates the rationality of our total score formula.

\paragraph{Citation Quality.} Citation quality evaluates the accuracy, relevance, and completeness of references in generated surveys. We adopt NLI-based metrics following AutoSurvey~\citep{Wang2024AutoSurveyLL}, which have been widely used in subsequent work~\citep{Wang2025LLMtimesMapReduceV2EC, Liang2025SurveyXAS, bao-etal-2025-surveygen}. The core idea is to judge whether each cited reference actually supports the corresponding claim in the survey.

Specifically, the evaluation first extracts all citation-bearing claims $C = \{c_1, c_2, \ldots\}$ from the survey, where each claim $c_i$ is associated with a set of cited references $\text{Ref}_i = \{r_{i1}, r_{i2}, \ldots\}$. An NLI model $h$ then determines whether the reference set entails the claim.

\textbf{Citation Recall} measures whether every claim in the survey is supported by its cited references---i.e., the completeness of evidential coverage:
\begin{equation}
    R = \frac{\sum_{i=1}^{|C|} \mathbb{I}[h(c_i, \text{Ref}_i) = 1]}{|C|}
\end{equation}
where $h(c_i, \text{Ref}_i) = 1$ indicates that the reference set collectively supports claim $c_i$, $\mathbb{I}[\cdot]$ is an indicator function, whose value is 1 when the condition is true, and 0 otherwise. A high recall means the survey's claims are well-grounded in cited literature.

\textbf{Citation Precision} measures whether each individual reference is actually relevant to the claim it supports---i.e., the accuracy of citation assignment. We define an auxiliary function $g$ to assess the necessity of each reference:
\begin{equation}
g(c_i, r_{ik}) = \mathbb{I}[h(c_i, \{r_{ik}\}) = 1] \lor \mathbb{I}[h(c_i, \text{Ref}_i \setminus \{r_{ik}\}) = 0]
\end{equation}
This function judges that reference $r_{ik}$ is relevant to claim $c_i$ if either (1)~$r_{ik}$ alone can support $c_i$, or (2)~removing $r_{ik}$ from $\text{Ref}_i$ causes the remaining set to fail to support $c_i$ (i.e., $r_{ik}$ is necessary). Precision is then computed as:
\begin{equation}
    P = \frac{\sum_{i=1}^{|C|} \sum_{k=1}^{|\text{Ref}_i|} \mathbb{I}[h(c_i, \text{Ref}_i) = 1 \land g(c_i, r_{ik}) = 1]}{\sum_{i=1}^{|C|} |\text{Ref}_i|}
\end{equation}
A high precision means that cited references are genuinely relevant rather than spurious.

\textbf{Valid Citation Ratio} reports the fraction of citations that are successfully parsed and used for evaluation, excluding entries with formatting errors or unresolvable references:
\begin{equation}
    \text{Valid Ratio} = \frac{|\text{Valid Citations}|}{|\text{Total Citations}|}
\end{equation}
This metric serves as a basic sanity check: a ratio below 1.0 indicates citation formatting issues or hallucinated references that could not be matched to any paper in the database.

The full evaluation pipeline proceeds as follows: (1) extract citation-bearing claims from the survey text and identify their reference numbers; (2) use the NLI model to judge entailment between each claim and its reference set; (3) for supported claims, further assess the relevance of individual references via the auxiliary function $g$; (4) aggregate all metrics into the final citation quality report.

\subsection{Detailed Evaluation Standard}\label{app:detailed_standard}
The complete 1--10 level descriptors for every sub-dimension, including the two supplementary code evaluation dimensions for CS surveys, are presented in Tables~\ref{tab:dseval_rubric_core}--\ref{tab:dseval_rubric_code} at the end of the appendix.

\paragraph{Agent-Oriented Adaptation of the Protocol.}
Beyond re-weighting the three dimensions, we calibrate several sub-dimensions toward the needs of agent consumers, following the premise in Section~\ref{sec:intro} that agent readers already hold broad knowledge and require analytical depth and reliable attribution more than restated breadth. The adaptations are as follows.

\textbf{Comprehensiveness} is calibrated to measure whether the survey covers the information needed for further research within the topical scope, such as representative methods, key mechanisms, and standard experimental settings, rather than the exhaustive inclusion of tangential material. For an agent, missing research-critical material is costly, whereas restating well-known background adds little.

\textbf{Future Directions} is calibrated to reward concrete and actionable directions, including specific open problems, candidate approaches, and implementation-oriented analysis, rather than generic outlook statements. This measures whether the survey can directly inform an agent's follow-up experiment design and idea generation.

\textbf{Specificity} is introduced as an additional sub-dimension beyond Agentic AutoSurvey's protocol. Under this premise, the information a survey adds for agent consumers concentrates in fine-grained mechanisms, design choices, and experimental configurations rather than in high-level summaries that largely restate established work; a granularity-sensitive dimension therefore measures how well a survey supplies exactly the knowledge that downstream agents cannot recover from parametric knowledge alone. 

\textbf{Critical Analysis} emphasizes the trade-offs, applicability boundaries, and failure modes of existing methods, which an agent needs to avoid redundant or already-refuted research directions.

\textbf{Writing Quality} is retained as a sanity check for human compatibility but down-weighted: agents consume survey content as evidence rather than as prose, so readability affects their consumption far less than it affects human readers. Within this dimension, we further merge the separate Clarity and Coherence sub-dimensions of Agentic AutoSurvey into a single Clarity \& Coherence criterion, which raises the relative weight of Academic Rigor; this reflects that Academic Rigor, which scores citation practice and the fair representation of prior work, is the content-side counterpart of citation reliability and thus carries greater significance for agent consumers than purely presentational aspects.

\subsection{Detailed Evaluation Parameters}\label{app:eval_eval_params}

Table~\ref{tab:system_params} lists the key system parameters used in DeepSurvey.
Note that we use MiniLMv2~\citep{Wang2020MiniLMv2MS} as embedding model. For DeepSurvey model, we choose MiMo-V2.5 for its lower cost and fast speed. In addition, MiMo-V2.5 is an open-source model that facilitates reproduction. The primary judge model is MiMo-V2.5\footnote{\url{https://mimo.xiaomi.com/mimo-v2-5}}, used for both content scoring and the NLI-based citation evaluation. For robustness, we additionally evaluate with judge models from other families, including GPT-5-mini~\citep{Singh2025OpenAIGS}, Gemini-3\footnote{\url{https://blog.google/products-and-platforms/products/gemini/gemini-3}}, DeepSeek-V3~\citep{DeepSeekAI2024DeepSeekV3TR}, GPT-5.4, and MiMo-V2-Flash~\citep{Xiao2026MiMoV2FlashTR} (Appendix~\ref{app:other_model_result}).

\subsection{Stability and Statistical Significance}\label{app:eval_results_supp}
Tables~\ref{tab:overall_results_std}, \ref{tab:core_quality_std}, \ref{tab:writing_quality_std}, \ref{tab:content_depth_std}, and~\ref{tab:retrieval_results_std} report the mean $\pm$ standard deviation (with coefficient of variation, CV) of each metric over repeated runs, and Table~\ref{tab:hedges_g} summarizes the mean differences and effect sizes (Hedges' $g$) of DeepSurvey against each baseline.

\paragraph{Score Stability.} DeepSurvey's aggregated scores are highly stable, with CVs of 0.8\%--2.2\% across the four dimensions---the lowest worst-case fluctuation among all methods---and its lead over the strongest baseline (e.g., $+0.27$ in Total Score over SurveyForge) far exceeds this noise level. Fine-grained dimensions remain similarly stable (CVs of 1.0\%--6.3\%), while the largest fluctuations occur in sub-dimensions of weaker baselines (e.g., Comprehensiveness for LiRA, 10.3\%; Academic Rigor for SurveyX, 10.6\%). Citation metrics fluctuate more, as they are computed at the claim level (CVs of 8.7\% and 7.5\% for DeepSurvey's precision and recall), yet stay within the range of the baselines (up to 9.9\%).

\paragraph{Statistical Significance.} All comparisons reach a large effect ($g \geq 0.8$, Cohen's convention) except Writing Quality against AutoSurvey and SurveyForge (both $g=0.11$), and Core Quality against SurveyForge ($g=0.29$). In particular, the advantages in citation reliability (Precision: $g=2.08$--$5.90$; Recall: $g=2.06$--$8.19$), Content Depth ($g=0.87$--$5.35$), and Total Score ($g=1.25$--$8.11$) are uniformly large, confirming that the depth and reliability gains are statistically meaningful rather than run-to-run noise. The only weak effects concentrate in Writing Quality, where DeepSurvey is closest to the strongest baselines, consistent with the main-text results.

\begin{table}[!htp]
\centering
\small
\setlength{\tabcolsep}{4pt}
\caption{Statistical significance of DeepSurvey vs.\ each baseline. Each cell reports the mean difference (left of ``/'') and the effect size (Hedges' $g$, right of ``/''). Bold $g$ values ($\geq 0.8$) indicate a large effect by Cohen's convention; non-bolded values are small ($0.2 \leq g < 0.5$) or negligible ($g < 0.2$).}
\label{tab:hedges_g}
\begin{tabular}{lcccc}
\toprule
Metric & vs AutoSurvey & vs SurveyForge & vs SurveyX & vs LiRA \\
\midrule
\multicolumn{5}{c}{\textbf{Citation Quality}} \\
\cmidrule(lr){1-5}
Precision  & 0.23 / \textbf{3.08} & 0.15 / \textbf{2.08} & 0.44 / \textbf{5.90} & 0.34 / \textbf{4.98} \\
Recall     & 0.19 / \textbf{4.97} & 0.13 / \textbf{2.06} & 0.44 / \textbf{8.19} & 0.19 / \textbf{6.37} \\
\midrule
\multicolumn{5}{c}{\textbf{Content Quality}} \\
\cmidrule(lr){1-5}
Total Score       & 0.38 / \textbf{1.46} & 0.27 / \textbf{1.25} & 1.35 / \textbf{4.76} & 2.65 / \textbf{8.11} \\
Core Quality      & 0.22 / \textbf{1.06} & 0.25 / 0.29          & 0.68 / \textbf{3.11} & 3.82 / \textbf{6.96} \\
Writing Quality   & 0.03 / 0.11          & 0.03 / 0.11          & 1.12 / \textbf{2.63} & 1.38 / \textbf{7.52} \\
Content Depth     & 0.73 / \textbf{1.55} & 0.41 / \textbf{0.87} & 2.13 / \textbf{5.35} & 2.13 / \textbf{5.11} \\
\bottomrule
\end{tabular}
\end{table}

\subsection{Cost Evaluation}\label{app:cost}
\paragraph{Time\&Token Cost}

We run DeepSurvey (under the setting in Appendix~\ref{app:eval_eval_params}) and baseline methods over multiple runs and calculate the average cost using the API provider's pricing information. Results are shown in Table~\ref{tab:cost_comparison}.

\begin{table}[H]
\centering
\small
\setlength{\tabcolsep}{3pt}
\caption{Cost comparison across methods.}
\label{tab:cost_comparison}
\begin{tabular}{lrrr}
\toprule
Method      &   Time &  Token & Money-Cost \\
\midrule
DeepSurvey  & 1407s  & 2.28M  & \$0.474    \\
AutoSurvey  &  844s  & 2.86M  & \$0.372    \\
SurveyForge &  902s  & 4.58M  & \$0.794    \\
LiRA        &  473s  & 0.31M  & \$0.060    \\
SurveyX     & 3873s  & 3.42M  & \$0.614    \\
\bottomrule
\end{tabular}
\end{table}

As can be seen, apart from LiRA's significantly lower costs, DeepSurvey's costs are comparable to other baselines---higher than AutoSurvey but lower than SurveyX and SurveyForge. In terms of time cost, DeepSurvey is higher than AutoSurvey and SurveyForge and lower than SurveyX.

Though DeepSurvey does not have the lowest cost, we consider this acceptable in exchange for surveys with greater analytical depth and citation reliability, which is our primary motivation.

\paragraph{Paper Budget.} 
\begin{table}[H]
\centering
\small
\setlength{\tabcolsep}{3pt}
\caption{Paper budget across domains. Note: Bio=Biology, Med=Medicine, Psy=Psychology, CS=Computer Science, MatSci=Materials Science.}
\label{tab:paper_budget}
\begin{tabular}{lrrrrrr}
\toprule
Metric              & Bio    & Med    & Psy    & CS      & MatSci & Overall \\
\midrule
citation/subsection & 7.544  & 8.201  & 8.927  & 9.592   & 7.312  & 8.315   \\
citation/section    & 20.805 & 21.582 & 24.667 & 26.320  & 16.699 & 22.015  \\
citation/paper      & 63.600 & 61.800 & 62.800 & 105.200 & 44.600 & 67.600  \\
\bottomrule
\end{tabular}
\end{table}
As paper budgets may vary significantly across domains, we supplement additional experiments across 5 domains in Table~\ref{tab:paper_budget}. Across all domains, the average paper budget per survey is 67.6, and 105.2 in Computer Science. The average per-section budget is 22.015. These results are averaged over multiple random runs.

\begin{figure}[t]
  \centering
  \includegraphics[width=0.8\linewidth]{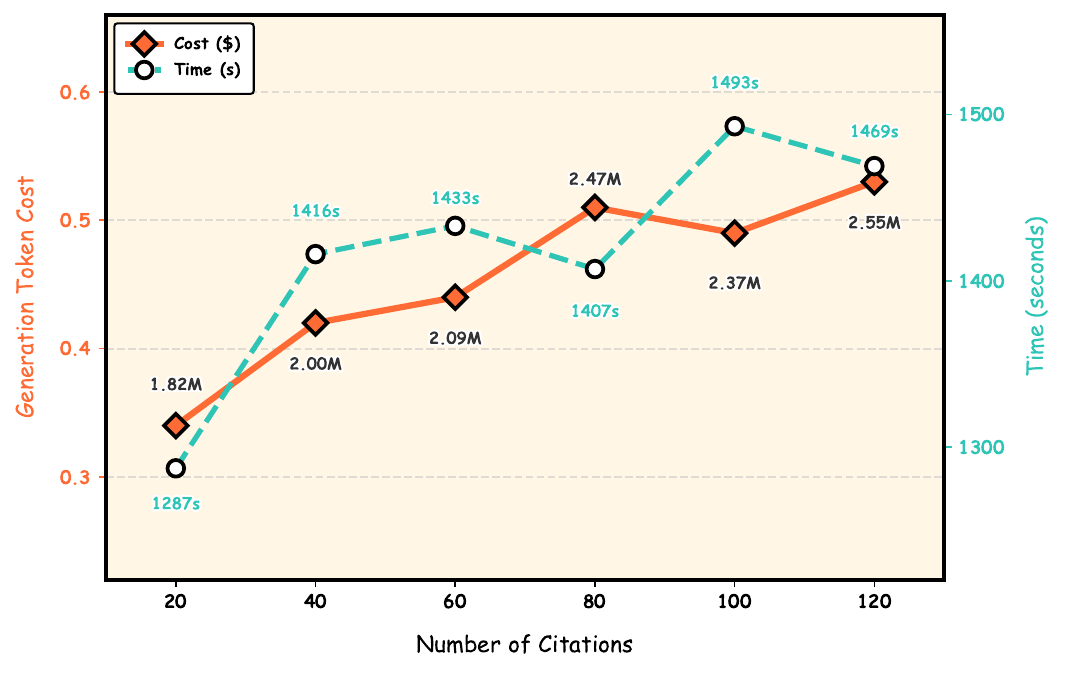}
  \caption{Generation cost, runtime, and token consumption as functions of the number of citations. The left y-axis shows the generation cost in USD, the right y-axis shows the runtime in seconds, and the token usage (in millions) is annotated above each data point.}
  \label{fig:cost_citation_curve}
\end{figure}

\paragraph{Relationship of Cost and Paper Budget}
We experiment on the relationship between time\&token cost and the paper budget for further exploration. Results are shown in Figure~\ref{fig:cost_citation_curve}.

\begin{table}[H]
\centering
\small
\setlength{\tabcolsep}{3pt}
\caption{Linear Scaling Relationship between Citation Budget and Generation Cost.}
\label{tab:linear_scaling}
\begin{tabular}{lrrrr}
\toprule
Metric        & Pearson r & p-value &    R²  & Slope (per citation) \\
\midrule
Time (s)      &   0.8307  &  0.0406 & 0.6901 &       1.5929     \\
Cost (\$)     &   0.9383  &  0.0056 & 0.8804 &       0.0018    \\
Token (M)     &   0.9486  &  0.0039 & 0.8999 &       0.0073     \\
\bottomrule
\end{tabular}
\end{table}

We briefly analyze the linear scaling in Table~\ref{tab:linear_scaling}. Cost and tokens scale roughly linearly with the citation count (Pearson r > 0.9), while time shows a much weaker linear relationship. We attribute this to optimizations such as parallelization and caching, which prevent time from growing linearly with the number of citations.

\subsection{Human Expert Experiment Supplement}\label{app:human_exp}
\paragraph{Detailed Report on Human Expert Evaluation}

\begin{table}[H]
\centering
\small
\setlength{\tabcolsep}{3pt}
\caption{Overview of the 6 Evaluation Topics.}
\label{tab:human_exp_eval_topics}
\begin{tabular}{lp{9cm}cc}
\toprule
No. & Paper Title & Publication Date & Citations \\
\midrule
1 & Reinforcement Learning for Large Reasoning Models & 2025 Sep & 158 \\
2 & Agentic Reinforcement Learning for LLMs & 2025 Sep & 163 \\
3 & Agentic AI Autonomous Intelligence for Complex Goals & 2025 & 576 \\
4 & Multi-Functional and Hybrid Reconfigurable Intelligent Surfaces for Integrated Sensing and Communication & 2025 Oct & 70 \\
5 & Vision-Language-Action Models: An Action Tokenization Perspective & 2025 Jul & 93 \\
6 & Edge General Intelligence with Agentic AI and Agentification & 2025 Aug & 56 \\
\bottomrule
\end{tabular}
\end{table}

The evaluation set contains 6 influential and popular topics, summarized in Table~\ref{tab:human_exp_eval_topics}. The source papers and metadata are obtained from Semantic Scholar. Each topic contains one human-written survey and one DeepSurvey-generated survey. Each topic is annotated by 4 annotators. All annotations are performed in a blinded setting. The final win rates are aggregated over all annotation samples.

We have excluded the corresponding human-written survey for each topic from the paper pool during DeepSurvey's generation process to prevent data leakage.

The evaluators were recruited from students and researchers with academic writings and evaluation experience in corresponding scientific domains. Participation was voluntary. No crowdworkers were used. 

\paragraph{Requirements for Human Expert}
\begin{dsoutput}{Annotation Requirements}
Annotation Task Instructions

You will complete multiple comparison groups. Each task contains two surveys, labeled A and B. Please read both surveys in each group, and then judge which is better on the following three dimensions, and give an overall judgment. The data obtained from this task will be used exclusively for research purposes. Any results made public will be reported only in aggregate, and no identifying information (such as your name or other personal identifiers) will be included.

Annotation Format
If A is better than B on a dimension, fill in A; if B is better than A, fill in B; if they are tied, fill in Tie.
Example: If A is better than B on Core Quality, note it as Core Quality: A.

Group 1
Core Quality:
Writing Quality:
Content Depth:
Overall Judgment:

Group 2
Core Quality:
Writing Quality:
Content Depth:
Overall Judgment:

...

Brief Description of Annotation Dimensions

1. Core Quality
Measures whether the survey stays closely on topic, effectively integrates the literature, and builds a systematic synthesis.
Key aspects: synthesis, organization, comprehensiveness, relevance.
Staying on-topic and building a systematic synthesis.

Synthesis: integrates papers into a coherent whole, not just listing
Organization: logical section flow and hierarchy
Comprehensiveness: covers key methods, works, branches
Relevance: minimal off-topic content

2. Writing Quality
Measures whether the language expression is clear, rigorous, and coherent, and whether technical details are accurately conveyed.
Key aspects: readability, academic rigor, clarity & coherence.
Clear, rigorous, and coherent writing.

Readability: fluent, natural, easy to follow
Academic Rigor: precise, well-supported academic expression
Clarity & Coherence: clear technical descriptions; consistent narrative across sections

3. Content Depth
Measures whether the survey goes beyond simple summary to offer critical analysis, unique insights, and actionable research suggestions.
Key aspects: critical analysis, novelty & insights, specificity, future directions.
Going beyond summary to offer insight and actionable directions.

Critical Analysis: discusses strengths, limitations, boundaries
Novelty & Insights: new perspectives or cross-work understanding
Specificity: fine-grained technical/experimental details
Future Directions: concrete, feasible, gap-linked suggestions

Overall Judgment
Please synthesize the performance across the above three dimensions and provide an overall comparison result for the two reviews.
\end{dsoutput}

\subsection{Citation Density Analysis}\label{app:citation_density}
A natural question is whether the citation reliability gains in Table~\ref{tab:all_results} merely reflect differences in citation volume, since both recall and precision are normalized over claims and references. We therefore compute citation density (citations per sentence) for DeepSurvey and the two strongest baselines, aggregated over all surveys generated in the main experiment. Micro density is computed as total citations divided by total sentences; macro density is the mean of per-survey densities.

\begin{table}[H]
\centering
\small
\setlength{\tabcolsep}{4pt}
\caption{Citation density comparison. ``Micro'' divides total citations by total sentences; ``Macro'' averages per-survey densities.}
\label{tab:citation_density}
\begin{tabular}{lrrrr}
\toprule
Method & Total Sentences & Total Citations & Density (micro) & Density (macro) \\
\midrule
DeepSurvey & 7{,}335 & 2{,}105 & 0.287 & 0.289 \\
AutoSurvey & 7{,}931 & 1{,}972 & 0.249 & 0.254 \\
SurveyForge & 6{,}670 & 2{,}051 & 0.307 & 0.303 \\
\bottomrule
\end{tabular}
\end{table}

Three observations follow. First, DeepSurvey issues the largest number of citations (2{,}105) while keeping a mid-range density (0.287 micro / 0.289 macro), between AutoSurvey (0.249 / 0.254) and SurveyForge (0.307 / 0.303): it neither inflates density, which would risk precision by attaching loosely related references, nor reduces it, which would sacrifice recall by leaving claims unsupported. In addition, the recall gap between DeepSurvey (0.630) and SurveyForge (0.503) cannot be explained by citing more per sentence: SurveyForge cites more often per sentence than DeepSurvey yet supports a smaller fraction of its claims, indicating that assignment quality, not volume, drives recall.

\subsection{Evaluation with Other Model} \label{app:other_model_result}
Since different LLMs differ in judging, adding experiments with judges from other families makes the results more convincing. We supplement evaluations with GPT-5-mini, Gemini-3-Flash and MiMo-V2-Flash as judge models (Tables~\ref{tab:gpt5_mini}, \ref{tab:gemini3_flash}, and~\ref{tab:mimo_v2_flash}).

Although these models show different scoring preferences, DeepSurvey consistently and substantially outperforms the strongest baseline across all three judges (7.982 vs. 7.633, 8.887 vs. 8.158, 8.676 vs. 8.483). Our conclusions remain unchanged under different judge model families.

\begin{table}[H]
\centering
\small
\setlength{\tabcolsep}{3pt}
\caption{Detailed evaluation results with GPT-5-mini.}
\label{tab:gpt5_mini}
\begin{tabular}{lrrrrrr}
\toprule
Method          & Total\_Score & Core\_Quality & Writing\_Quality & Content\_Depth & Precision & Recall    \\
\midrule
DeepSurvey & \textbf{7.982} & \textbf{8.483} & 7.412           & \textbf{7.767}  & \textbf{0.560}         & \textbf{0.618}        \\
AutoSurvey      & 7.633        & 8.354         & \textbf{7.500}           & 6.979          & 0.485         & 0.571        \\
LiRA            & 7.236        & 7.659         & 7.364           & 6.750          & 0.129         & 0.290        \\
SurveyForge     & 7.529        & 8.396         & 6.979           & 6.938          & 0.420         & 0.462        \\
SurveyX         & 5.887        & 6.633         & 5.533           & 5.317          & 0.276         & 0.312        \\
\bottomrule
\end{tabular}
\end{table}

\begin{table}[H]
\centering
\small
\setlength{\tabcolsep}{3pt}
\caption{Detailed evaluation results with Gemini-3-Flash.}
\label{tab:gemini3_flash}
\begin{tabular}{lrrrrrr}
\toprule
Method          & Total\_Score & Core\_Quality & Writing\_Quality & Content\_Depth & Precision   & Recall      \\
\midrule
DeepSurvey & \textbf{8.887} & \textbf{9.300} & \textbf{8.533}           & \textbf{8.650}  & \textbf{0.475}         & \textbf{0.541} \\
AutoSurvey      & 6.967        & 7.500         & 7.583           & 6.125          & 0.445         & 0.489        \\
LiRA            & 7.344        & 7.667         & 7.556           & 6.917          & 0.066         & 0.224        \\
SurveyForge     & 8.158        & 8.875         & 7.667           & 7.688          & 0.273         & 0.312        \\
SurveyX         & 5.747        & 6.950         & 5.933           & 4.450          & 0.154         & 0.187        \\
\bottomrule
\end{tabular}
\end{table}

\begin{table}[H]
\centering
\small
\setlength{\tabcolsep}{3pt}
\caption{Detailed evaluation results with MiMo-V2-Flash.}
\label{tab:mimo_v2_flash}
\begin{tabular}{lrrrrrr}
\toprule
Method          & Total\_Score & Core\_Quality & Writing\_Quality & Content\_Depth & Precision   & Recall      \\
\midrule
DeepSurvey & \textbf{8.676} & \textbf{9.083} & 8.311           & \textbf{8.450}  & \textbf{0.700}         & \textbf{0.654}        \\
AutoSurvey      & 8.483        & 8.938         & \textbf{8.417}   & 8.063          & 0.648         & 0.623        \\
SurveyForge     & 8.101        & 8.708         & 7.795           & 7.646          & 0.379         & 0.395        \\
SurveyX         & 7.244        & 7.667         & 7.222           & 6.833          & 0.344         & 0.287        \\
LiRA            & 8.004        & 8.455         & 7.793           & 7.659          & 0.337         & 0.211        \\
\bottomrule
\end{tabular}
\end{table}

\section{Ablation Experiment Supplement}\label{app:ablation_experiment_supp}
\subsection{Multi-granularity Refinement Component Ablation}\label{app:refinement_ablation}

The iterative refinement module operates at three granularity levels: section-level, subsection-level, and survey-level. To assess the individual contribution of each, we ablate them independently. w/o~Section Refinement removes section-level optimization; w/o~Subsection Refinement removes subsection-level optimization; w/o~Survey Refinement removes survey-level optimization; w/o~All removes all three. Note that the base model for this experiment is MiMo-V2-Flash rather than MiMo-V2.5, so the data differs from the main text.

\begin{table}[!htp]
\centering
\small
\caption{Ablation results on refinement components.}
\label{tab:ablation_refinement}
\begin{tabular}{lcccc}
\toprule
Method & Total & Core & Writing & Depth \\
\midrule
DeepSurvey & \textbf{8.676} & 9.083 & 8.311 & \textbf{8.450} \\
w/o Sec. & 8.544 & 9.042 & \textbf{8.389} & 8.125 \\
w/o Sub. & 8.487 & 9.000 & 8.333 & 8.050 \\
w/o Sur. & 8.487 & 9.050 & 8.333 & 8.000 \\
w/o All & 8.348 & \textbf{9.107} & 7.810 & 7.857 \\
\bottomrule
\end{tabular}
\end{table}

Table~\ref{tab:ablation_refinement} presents the main dimension results and Table~\ref{tab:multi_granularity_refinement_ablation} presents the detailed dimension results. We highlight three key findings.

(1)~\textbf{Subsection-level refinement is a critical component.} Removing it causes a substantial total score drop (8.676$\rightarrow$8.487, $\Delta$=0.189) and a substantial content depth decrease (8.450$\rightarrow$8.050, $\Delta$=0.400). Since subsections constitute the basic writing unit of a survey, optimizing at this granularity directly improves the quality of individual research discussions, yielding the strongest impact on overall performance.

(2)~\textbf{Survey-level refinement is indispensable for cross-section consistency.} Although its total score impact equals that of subsection-level removal (both $\Delta$=0.189), survey-level refinement uniquely affects content depth more severely ($\Delta$=0.450, the largest single-component depth drop). This indicates its irreplaceable role in harmonizing cross-section terminology, eliminating redundancy, and ensuring global narrative coherence---issues that local refinement cannot address.

(3)~\textbf{The three levels are complementary, not redundant.} Removing all refinement (w/o~All) degrades writing quality to 7.810 and content depth to 7.857, both significantly worse than any single ablation. This confirms that subsection-level refinement targets local readability, section-level refinement enhances analytical depth within sections, and survey-level refinement maintains global consistency. Together they form a complete multi-granularity optimization system.

From a resource allocation perspective, subsection-level and survey-level refinement each contribute 0.189 to the total score, while section-level contributes 0.132. When computational budget is constrained, prioritizing subsection and survey-level refinement yields the highest marginal benefit.

\subsection{Citation Mark Ablation}\label{app:citation_mark_ablation}

DeepSurvey uses paper titles (e.g., \texttt{<AutoSurvey: Large Language Models Can Automatically Write Surveys>}) as citation marks in generated surveys. To validate this design choice, we compare it against an alternative using paper IDs (e.g., \texttt{<2406.10252>}). The two strategies represent a trade-off: paper IDs are short and format-uniform, reducing context window consumption and enabling direct lookup via academic databases, whereas paper titles carry richer semantic information that may help the LLM more accurately associate claims with their supporting references, but at the cost of significantly increased input length.

\begin{figure}[t]
  \centering
  \includegraphics[width=8cm]{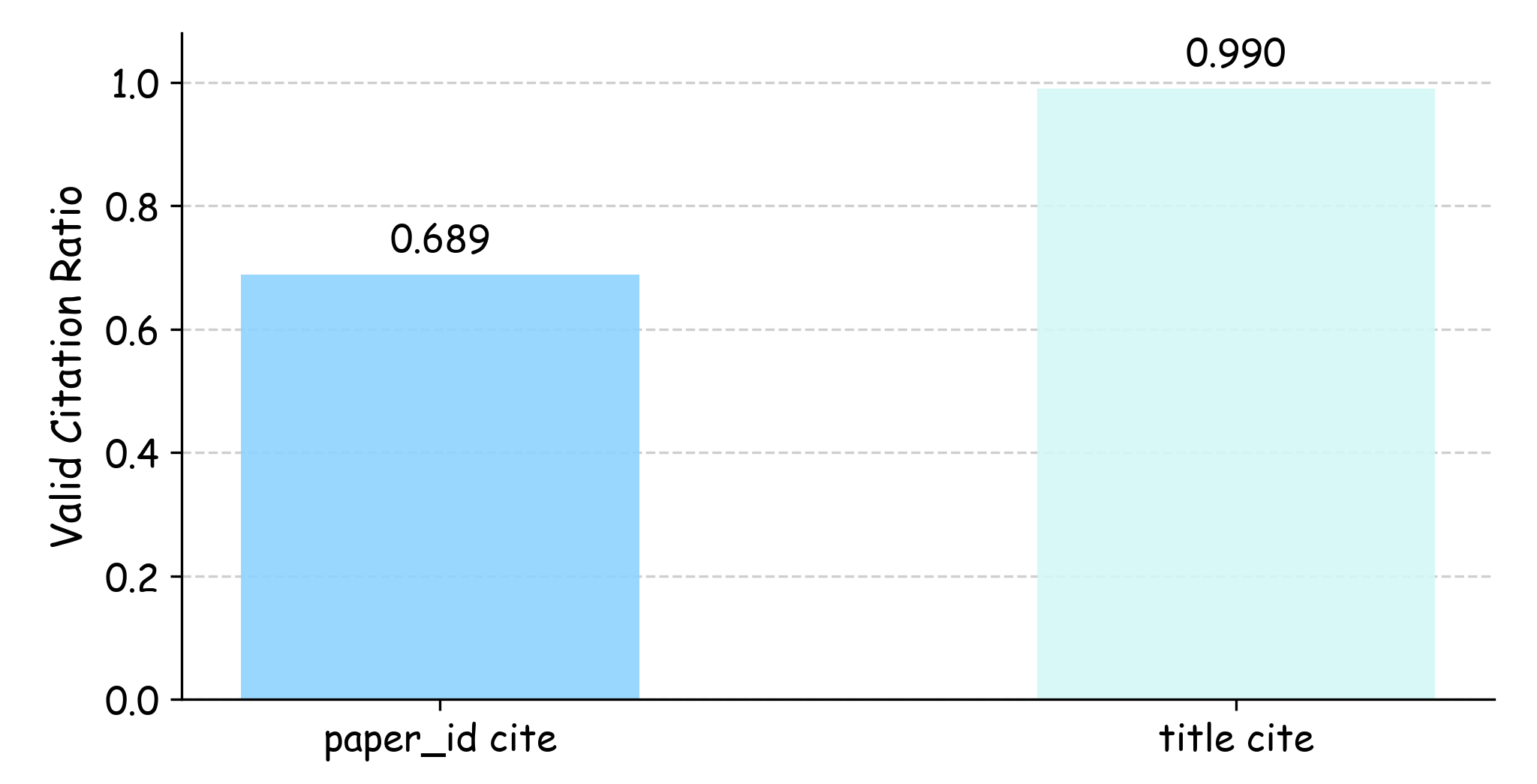}
  \caption{Valid citation ratio comparison between paper ID and paper title citation marks.}
  \label{fig:citation_ratio}
\end{figure}

Figure~\ref{fig:citation_ratio} shows that title-based citation marks achieve a substantially higher valid citation ratio. A plausible explanation is that paper titles provide explicit semantic cues---such as method names, task descriptions, and key terms---that help the LLM ground its citations during generation, reducing hallucinated or misattributed references. In contrast, paper IDs are semantically opaque identifiers that require the model to maintain ID--content mappings from context alone, a burden that intensifies in long-text generation scenarios. This finding suggests that the representational format of input information materially affects LLM generation quality, and that careful design of citation representations can improve reliability without modifying the underlying model.

\subsection{Draft Quality Improvement in Refinement}
\label{app:refinement_validation}

To validate that the multi-granularity refinement subsystem (Section~\ref{sec:refinement}) genuinely improves draft quality rather than merely altering surface expression, we conduct a controlled evaluation comparing survey drafts before and after refinement. We evaluate whether this multi-granularity coordination yields measurable quality gains.

Specifically, we employ a panel of expert LLMs spanning multiple model families (GPT-series, Gemini-series, and DeepSeek-series) to independently assess draft quality along four dimensions before and after refinement: \textbf{critical analysis depth} (whether arguments are substantiated with comparative reasoning rather than superficial description), \textbf{logical coherence} (whether paragraphs follow a clear argumentative thread with appropriate transitions), \textbf{academic rigor} (whether claims are properly qualified and supported by citations), and \textbf{readability} (whether the text is clear, concise, and free of redundancy). Each model scores drafts on a 1--10 scale for each dimension, and we report the average across all models.

\begin{figure}[t]
  \centering
  \includegraphics[width=0.8\linewidth]{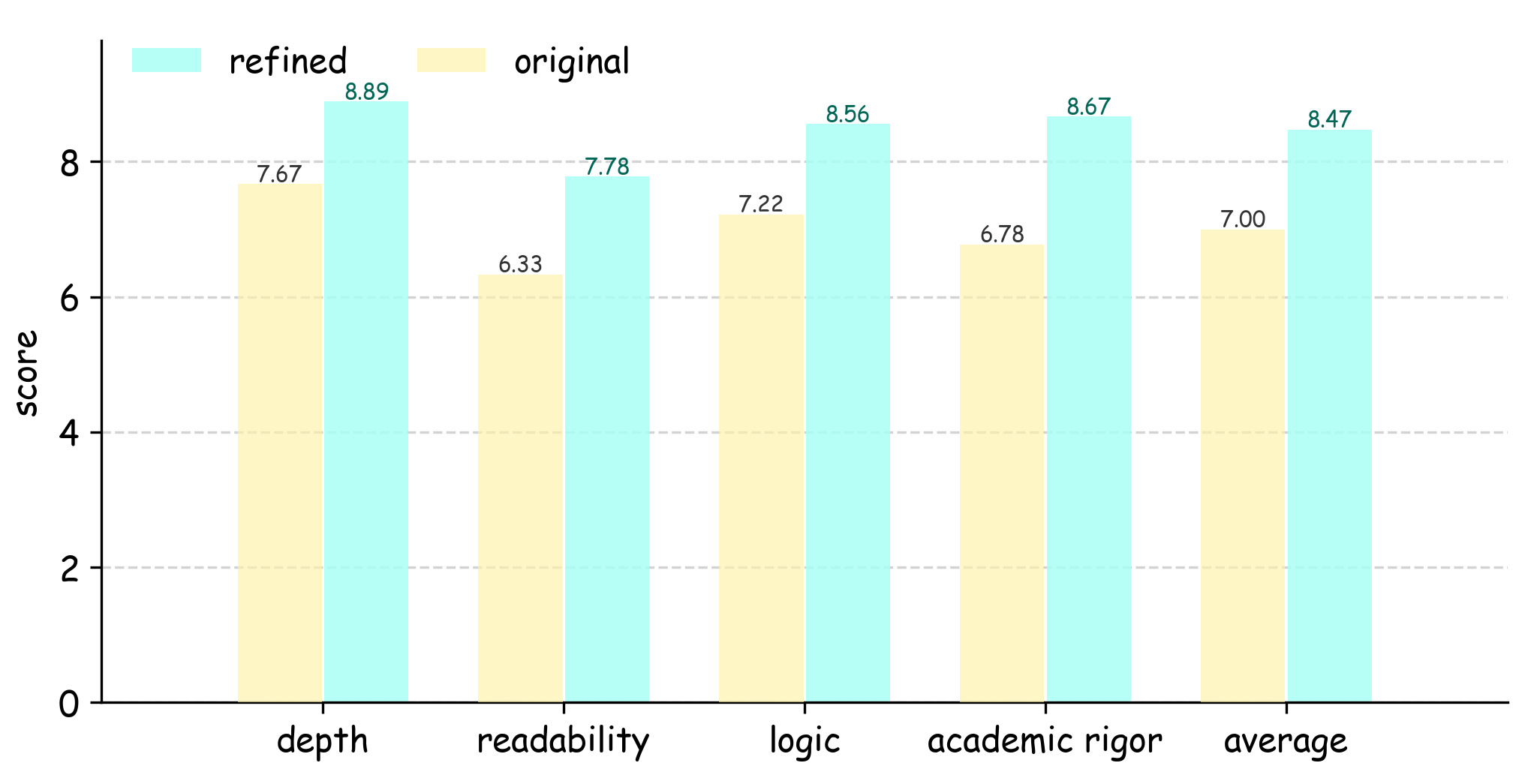}
  \caption{Quality dimension scores before and after multi-granularity refinement, averaged across expert LLM judges.}
  \label{fig:refinement_validation}
\end{figure}

Figure~\ref{fig:refinement_validation} presents the results. The refinement subsystem yields consistent improvements across all four dimensions, with the most pronounced gains in critical analysis depth and academic rigor. This aligns with the design intent: the reviewer role specifically targets analytical weaknesses and citation issues, while the multi-granularity optimization strategy addresses coherence at local (subsection), structural (section), and global (survey) levels. 

\subsection{Detailed Scores of the Module Ablation}\label{app:detail_score_in_module_ablation}
Due to space constraints in the main body, we present fine-grained dimension scores for the module ablation in Table~\ref{tab:all_quality_ablation}, highlighting key findings that complement the aggregated analysis in Section~\ref{ssec:ablation}.

\textbf{The collector is the primary bottleneck.} Removing graph-based collection causes catastrophic drops in Comprehensiveness (8.889$\rightarrow$3.333) and Relevance (9.300$\rightarrow$1.333), confirming that without curated evidence, the system cannot cover the topic space. The damage extends to Academic Rigor (8.506$\rightarrow$6.000), Clarity \& Coherence (8.333$\rightarrow$4.333), and several depth metrics (e.g., Specificity 8.194$\rightarrow$6.942), indicating that evidence scarcity degrades not only coverage but also the rigor and coherence of the generated text.

\textbf{The writer primarily governs synthesis, organization, and academic rigor.} Its removal causes Synthesis Quality (8.334$\rightarrow$7.667) and Organization (8.935$\rightarrow$7.333) to drop sharply and collapses Academic Rigor (8.506$\rightarrow$5.333) and Critical Analysis (7.389$\rightarrow$5.341), both the lowest among all ablations, while Comprehensiveness (8.000) and Relevance (8.933) are comparatively less affected. This confirms that outline-driven writing with explicit citation assignment is essential for well-grounded, insightful narratives rather than for evidence coverage per se.

\textbf{The analyzer contributes primarily to depth-oriented metrics.} Its removal most affects Specificity (8.194$\rightarrow$7.286, the largest single-metric drop for this component) and Novelty and Insights (7.866$\rightarrow$7.111), while Comprehensiveness and Relevance remain near ceiling (8.857 and 9.000), consistent with its role in constructing relation graphs and comparison tables for cross-paper synthesis.

\textbf{Refinement polishes rather than transforms.} Removing refinement causes moderate declines across most metrics rather than catastrophic failures, with the largest drops in Novelty and Insights (7.866$\rightarrow$6.948), Organization (8.935$\rightarrow$8.123), Clarity \& Coherence (8.333$\rightarrow$7.528), and Future Directions (8.139$\rightarrow$7.333), consistent with its role in polishing insight expression and global coherence.

\textbf{The remaining variants cause milder, depth-oriented declines.} w/o keynote, w/o agentic, and w/o code all reduce Novelty and Insights (to 6.667, 7.635, and 7.533) and Specificity (to 7.671, 7.914, and 7.833), consistent with the aggregated results in Section~\ref{ssec:ablation}, where each incurs its largest drop in Content Depth.


\section{Clustering and Analysis Cases} \label{app:cluster_analysis_case}
In this section, we demonstrate specific cases from the global knowledge substrate.

\subsection{Keynote Case}\label{app:full_text_read_case}
We use `AutoSurvey' as the input topic. The generated keynotes are as follows:
\begin{dsoutput}{Keynotes Case}

"title": "AutoSurvey: Large Language Models Can Automatically Write Surveys",
"authors": ["Yidong Wang", "Qi Guo", "Wenjin Yao", "Hongbo Zhang", "Xin Zhang", "Zhen Wu", "Meishan Zhang", "Xinyu Dai", "Min Zhang", "Qingsong Wen", "Wei Ye", "Shikun Zhang", "Yue Zhang"], 
"abstract": "This paper introduces AutoSurvey, a speedy and well-organized methodology for automating the creation of comprehensive literature surveys in rapidly evolving fields like artificial intelligence. Traditional survey paper creation faces challenges due to the vast volume and complexity of information, prompting the need for efficient survey methods. While large language models (LLMs) offer promise in automating this process, challenges such as context window limitations, parametric knowledge constraints, and the lack of evaluation benchmarks remain. AutoSurvey addresses these challenges through a systematic approach that involves initial retrieval and outline generation, subsection drafting by specialized LLMs, integration and refinement, and rigorous evaluation and iteration.",
"key_contributions": [
    "Introduction of AutoSurvey, a comprehensive system for automatically generating academic surveys using LLMs.", 
    "A two-stage parallel generation approach (outline generation and subsection drafting) to overcome context window limitations and speed up the process.",    
    "Integration of Retrieval-Augmented Generation (RAG) for real-time knowledge updates to ensure citations are current and accurate."
    "Development of a Multi-LLM-as-Judge evaluation framework with metrics for citation quality (recall and precision) and content quality (coverage, structure, relevance).",
    "Experimental validation demonstrating AutoSurvey\'s effectiveness in terms of speed, citation quality, and content quality, approaching human performance.",
    "Open-sourcing of resources and prompts to facilitate further research."],
"methodology": "AutoSurvey follows a four-phase pipeline: 
    (1) Initial Retrieval and Outline Generation, where a database of papers is scanned using embedding-based retrieval to generate a structured outline; 
    (2) Subsection Drafting, where specialized LLMs draft each section in parallel using the outline and retrieved papers; 
    (3) Integration and Refinement, where sections are refined for coherence, corrected for citation errors, and merged into a cohesive document;
    (4) Rigorous Evaluation and Iteration, where the survey is assessed using Multi-LLM-as-Judge and the best among multiple candidate surveys is selected. The system uses a retrieval database of 530,000 arXiv papers and leverages parallel processing to enhance efficiency."
"experiments": {
    "setup": "Comparisons were made against human-authored surveys and naive RAG-based LLM generation across 20 computer science topics. Surveys of varying lengths (8k, 16k, 32k, 64k tokens) were generated. Claude-3-Haiku was used as the base writer for AutoSurvey and naive RAG, while evaluations employed a combination of GPT-4, Claude-3-Haiku, and Gemini-1.5-Pro. Metrics included survey creation speed, citation quality (recall and precision), and content quality (coverage, structure, relevance) scored on a 5-point rubric.",
    "baselines": [
        "Human-authored surveys collected from arXiv",      
        "Naive RAG-based LLM generation (iterative prompting with retrieved papers)"],    
    "main_results": {
        "speed": "AutoSurvey is significantly faster: for 64k-token surveys, it generates 73.59 surveys per hour vs. 12.56 for naive RAG and 0.07 for humans.",   
        "citation_quality": "AutoSurvey achieves near-human recall and precision (e.g., at 64k tokens: recall 82.25%, precision 77.41% vs. human 86.33% and 77.78%), outperforming naive RAG (68.79% and 61.97%).",    
        "content_quality": "AutoSurvey scores highly (e.g., at 16k tokens: coverage 4.66, structure 4.33, relevance 4.86, average 4.60) compared to human (4.66, 4.38, 5.00, 4.66) and naive RAG (4.46, 3.66, 4.73, 4.23)."},
        "additional_studies": {
            "meta_evaluation": "Spearman\'s rank correlation between LLM-based and human rankings showed moderate to strong positive correlation (0.5429 for mixed models), indicating alignment with human preferences.",     
            "ablation_study": "Removing retrieval drastically reduces citation quality; removing reflection slightly impacts content quality. Different base LLMs (GPT-4, Claude-haiku, Gemini-1.5-Pro) all yield competitive results, with GPT-4 slightly best.", 
            "knowledge_enrichment_test": "Providing AutoSurvey-generated surveys improved accuracy on multiple-choice questions by 9.2% over direct answers and 2.4% over naive RAG surveys."}},
"results": "AutoSurvey demonstrates a strong balance between speed and quality, achieving near-human performance in citation and content quality while being orders of magnitude faster. It outperforms naive RAG-based generation in all metrics, particularly in structure and citation accuracy. The system is robust across different survey lengths and base LLMs.",
"significance": "AutoSurvey addresses a critical bottleneck in academic research: the time-consuming process of literature review. By automating survey generation with high quality and speed, it can accelerate knowledge synthesis, help researchers stay current in fast-moving fields, and fill gaps in survey coverage. The open-sourced methodology and evaluation framework provide a foundation for future work in automated academic writing.",
"limitations": [
    "Citation errors, particularly overgeneralization (51% of errors), indicate reliance on parametric knowledge over retrieved information.",
    "The evaluation, while correlated with human judgment, is still automated and may not capture all nuances.",
    "The system is currently focused on computer science topics; generalizability to other domains needs exploration.",
    "Ethical considerations around authorship and potential misuse of auto-generated surveys."],
    "future_directions": [
    "Improving citation accuracy by reducing overgeneralization through better integration of retrieved knowledge.",
    "Extending the system to other domains and document types (e.g., project reports, white papers).",
    "Enhancing the evaluation framework with more nuanced metrics and human-in-the-loop refinement.",
    "Exploring iterative refinement where the system incorporates feedback from initial drafts to improve subsequent versions.",
    "Investigating the use of more advanced or specialized LLMs to further boost quality."],
"critical_reflections": [
    "AutoSurvey\'s parallel generation approach elegantly tackles context window limits, but the merging step may introduce coherence issues; the refinement phase mitigates this but adds computational cost.",
    "The Multi-LLM-as-Judge strategy reduces single-model bias, yet the moderate correlation with human experts suggests room for improvement in evaluation fidelity.",
    "While AutoSurvey outperforms naive RAG, the gap between it and human performance in structure and overall quality indicates that LLMs still struggle with holistic document organization and logical flow.",
    "The cost analysis shows AutoSurvey is cost-effective (e.g., \$1.2 per 32k survey with Claude-haiku), making it practical for large-scale use.",
    "The system\'s reliance on arXiv papers may limit coverage of non-open-access literature, potentially introducing source bias.",
    "The example survey on emotion recognition (Appendix F) demonstrates the system\'s capability to generate coherent outlines and content, but human oversight is still recommended for final quality assurance."],
    "tldr": "AutoSurvey automates the creation of academic surveys by leveraging LLMs in a parallel, retrieval-augmented pipeline: it generates outlines, drafts sections concurrently, refines for coherence, and evaluates with multi-LLM judges. It significantly speeds up survey writing while achieving citation and content quality close to human experts, outperforming naive RAG approaches. This offers a scalable tool for synthesizing research literature in fast-evolving fields."
\end{dsoutput}

\subsection{Clustering Result}\label{app:cluster_case}
We use `AutoSurvey' as the input topic. The clustering result is shown in Table~\ref{tab:cluster-overview}.

\subsection{Intra-cluster Q\&A Analysis Example} \label{app:QA_case}
We use ``AutoSurvey'' as the input topic for retrieval, clustering, and intra-cluster Q\&A analysis. One example of the Q\&A analysis is shown below.

\textbf{Cluster 1}
\textbf{Question:} Several papers propose explicit hierarchical structures (e.g., knowledge trees, citation graphs) to organize surveys, while others rely on iterative outline refinement without such explicit models. How do these fundamentally different approaches to structural planning compare in terms of their ability to capture research evolution and ensure logical flow, and what are the trade-offs in complexity versus performance?

\textbf{Related Papers:} \newline
\quad \texttt{2601.09504}
\quad \texttt{2510.07733} (2)
\quad \texttt{2510.21900} (3)
\quad \texttt{2510.05138} (4)
\quad \texttt{2509.18661} (5)

\textbf{{Answer:}}
Explicit hierarchical structures (e.g., knowledge trees, citation graphs) directly model research evolution and logical flow. MVSS uses Hierarchical Knowledge Trees to explicitly map conceptual evolution and paradigm shifts, yielding superior structural quality and citation precision \texttt{<MVSS>}. Similarly, SurveyG's layered citation graph (Foundation/Development/Frontier) captures temporal progression, enhancing synthesis and critical analysis \texttt{<SurveyG>}. These approaches ensure strong logical flow but increase complexity and cost (e.g., MVSS costs $\sim$\$0.94/topic).

Iterative refinement methods (e.g., IterSurvey, LiRA) rely on incremental outline updates without explicit models. IterSurvey mimics human research via recurrent outline generation and paper cards, improving coverage and structure iteratively \texttt{<IterSurvey>}. LiRA uses multi-agent workflows for parallel drafting and refinement, focusing on readability and citation accuracy \texttt{<LiRA>}. These methods are more flexible and efficient but may lack explicit evolutionary modeling.

\textbf{Trade-offs}: Explicit hierarchies offer better structural coherence and evolution capture at higher computational cost. Iterative methods are more adaptable and efficient but may produce less explicitly organized narratives.

\subsection{Comparison Table Example}\label{app:intra_cluster_table_case}
\label{appendix:table-comparison}

We use ``AutoSurvey'' as the input topic for retrieval and clustering analysis, and construct comparison tables. Table~\ref{table:comparable_table} presents an example comparing different papers under the ``AutoSurvey'' topic.

\subsection{Relation Graph Example}\label{app:intra_cluster_graph_case}
\begin{figure*}[t]
  \centering
  \includegraphics[width=12cm]{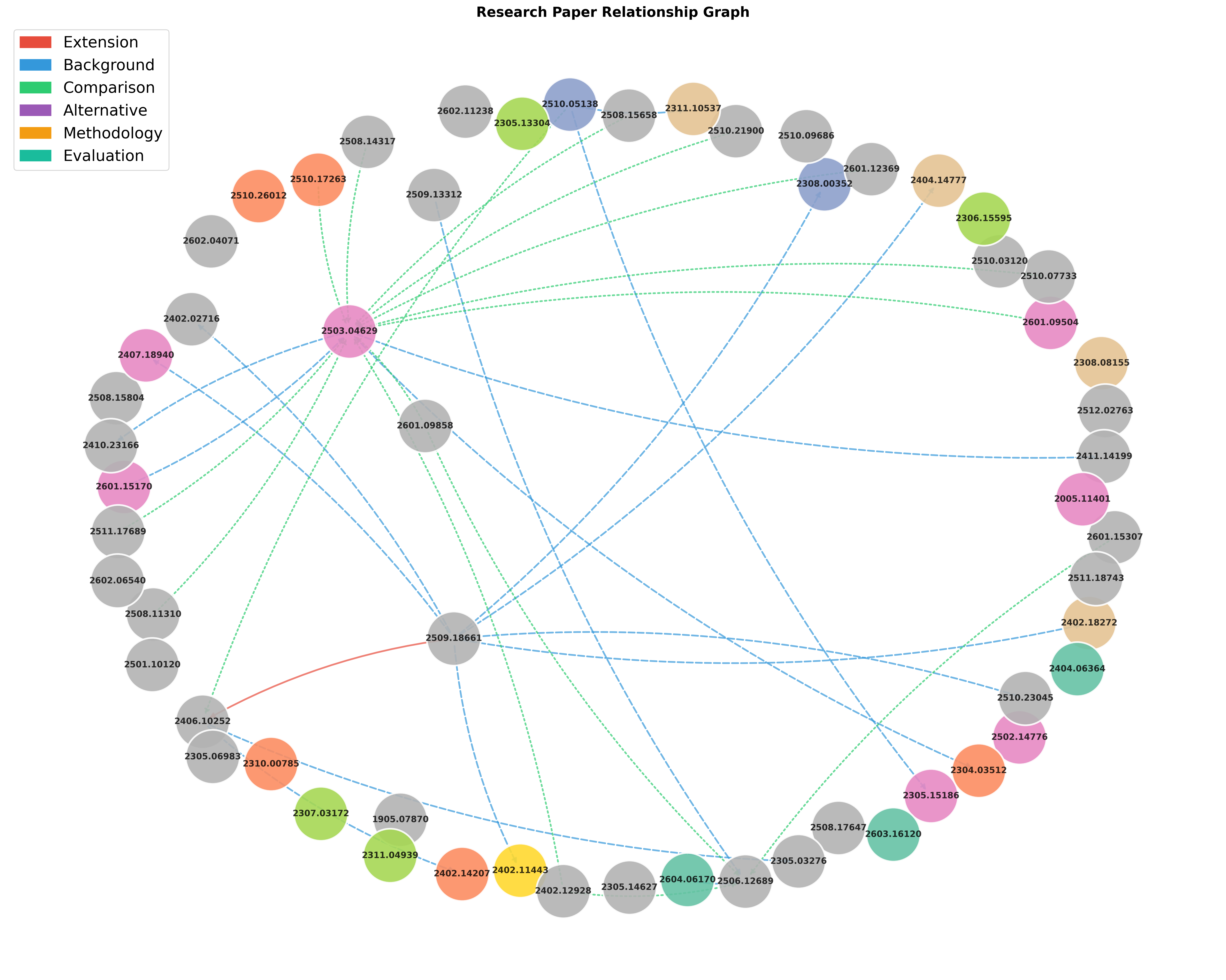}
  \caption{Example of a relation graph among papers}
  \label{fig:cluster_graph}
\end{figure*}

We use ``AutoSurvey'' as the input topic for retrieval, clustering, and graph construction. An example is shown in Figure~\ref{fig:cluster_graph}, which illustrates the relationships between different papers under this topic.

\section{Code Analysis Case}\label{app:code_analysis_case}
\subsection{Code Report Example} \label{app:code_report_case}
We use ``AutoSurvey'' as the input topic to generate a code report example, which analyzes data structures, design architectures, and other implementation-specific details. Due to space constraints, only the executive summary and problem-modeling parts are shown.
\begin{dsoutput}{Code Report Case}
# Integrated Comprehensive Report: AutoSurvey and Related Papers

## 1. Executive Summary

The synthesis of multiple batch analyses reveals a clear evolutionary trajectory in automated survey generation: from monolithic, single-pass systems toward **modular, multi-agent, iterative frameworks** that mimic human research workflows. The dominant implementation pattern is a **hierarchical pipeline** combining retrieval-augmented generation (RAG), dynamic planning, specialized agent collaboration, and iterative refinement. Key architectural elements include:

- **Multi-agent orchestration**: Specialized agents (Planner, Writer, Reflector, Verifier) with defined roles and communication protocols.
- **Dynamic planning-evidence loops**: Iterative cycles where outline generation and evidence acquisition inform each other.
- **Memory-augmented generation**: Persistent memory banks (e.g., paper cards, evidence memory) for targeted retrieval and context management.
- **Structured intermediate representations**: Hierarchical outlines, citation graphs, and standardized operating procedures (SOPs) that guide generation.

**Major gaps** persist in deep understanding, evaluation alignment, long-context handling, and domain generalization. The most promising future directions involve **neuro-symbolic integration**, **adaptive context management**, **evaluation-aware generation**, and **collaborative human-AI systems** that address these limitations.

## 2. Problem Modeling and Data Structure Classification

A unified view reveals that all systems model survey generation as a **hierarchical, evidence-grounded synthesis problem**. The core challenge is decomposing a broad topic into a coherent structure while integrating and synthesizing information from numerous sources.

### Data Structure Categories and Trade-offs

| Category | Specific Examples | Why Chosen | Key Trade-offs |
|--|--|--|--|
| **Hierarchical Outline Structures** | `DynamicOutline` (WebWeaver), `HierarchicalDraft` (SciSage), Tree-structured outlines (AutoSurvey, AutoSurvey2) | Mirrors human survey organization; enables modular, parallel generation and targeted retrieval. | **Advantage**: Parallel section generation possible; clear quality control per section. **Disadvantage**: Rigid structure may not fit all topics; requires accurate initial outline; may lose global coherence. |
| **Memory-Augmented Representations** | Natural language "Long Short-Term Memory" buffers (RecurrentGPT), "Paper cards" distilling contributions (IterSurvey), `EvidenceMemoryBank` (WebWeaver) | Enables long-context generation; provides faithful paper-level grounding; reduces redundant processing. | **Advantage**: Interpretable, editable memory; targeted retrieval. **Disadvantage**: Memory management overhead; potential information loss in distillation; retrieval latency. |
| **Agent Communication Structures** | `SOPPrompts` (MetaGPT), `AgentMessageQueue` (implied), Role-based prompt engineering | Enables structured collaboration between specialized agents; reduces hallucination through validation. | **Advantage**: Mimics human workflows; clear responsibility boundaries. **Disadvantage**: Increased complexity; potential communication bottlenecks; may be too rigid for creative tasks. |
| **Retrieval-Augmented & Citation Structures** | Retrieved paper corpora with metadata, `CitationGraph` (ReportBench), citation-indexed evidence storage | Addresses parametric knowledge limitations of LLMs; enables real-time updates and citation verification. | **Advantage**: Real-time updates; reduces redundancy; improves citation accuracy. **Disadvantage**: Retrieval noise; computational cost; memory overhead. |
| **Evaluation Structures** | `EvaluationCorpus` (ReportBench), `QuestionAnswerPairs` (LooGLE) | Enables systematic benchmarking and quality assessment. | **Advantage**: Standardized evaluation; reproducible results. **Disadvantage**: May not capture all quality dimensions; static benchmarks. |
...
\end{dsoutput}

\subsection{Environment Report Example} \label{app:env_report_case}
We use ``AutoSurvey'' as the input topic to generate an environment report example, which analyzes base frameworks, environment requirements, and commonly used libraries. Due to space constraints, only excerpts from the framework analysis are shown.
\begin{dsoutput}{Environment Report Case}
# Comprehensive Framework Selection and Environment Configuration Guide for AI Research Systems

## Executive Summary

This guide synthesizes findings from 15 repositories across three research domains: automated survey generation, deep research agents, and specialized AI tools. The analysis reveals a clear convergence on **PyTorch** as the dominant deep learning framework, complemented by **Hugging Face Transformers** for NLP tasks and **LangChain** for LLM orchestration. Environment requirements vary significantly based on use cases, ranging from lightweight API-based setups to GPU-intensive training configurations.

**Key Findings:**
- **Framework Standardization**: PyTorch (v2.0.0+) with Transformers (v4.30.0+) forms the standard stack
- **Orchestration Layer**: LangChain (v0.2.5+) and LlamaIndex (v0.8.29+) are preferred for complex workflows
- **Hardware Spectrum**: Requirements span from CPU-only (tiktoken) to multi-GPU setups (OpenScholar)
- **Deployment Patterns**: Docker containerization is emerging for reproducible deployments
- **API Dependencies**: Most systems require multiple external API keys (OpenAI, Semantic Scholar, etc.)

**Recommendation:** Start with **AutoSurvey** for survey generation or **MetaGPT** for multi-agent systems, then scale based on specific requirements. Use **conda** for environment isolation and consider **Docker** for production deployments.

---

## 1. Framework Selection Analysis

### 1.1 Core Framework Landscape

**PyTorch Dominance**: All analyzed repositories use PyTorch as their primary deep learning framework, with version requirements ranging from 2.0.1 to 2.8.0. This consistency enables easier integration and knowledge transfer between projects.

**NLP-Specific Frameworks**:
- **Hugging Face Transformers**: Critical across all NLP tasks (v4.30.0 to v4.56.1)
- **Sentence-Transformers**: Used for embedding generation (v2.7.0 to v5.1.0)
- **vLLM**: Emerging for efficient LLM inference (v0.10.1 in DeepResearch)

**Orchestration Frameworks**:
- **LangChain**: Preferred for LLM workflow orchestration (v0.0.300 to v0.2.5)
- **LlamaIndex**: Used for retrieval-augmented generation (v0.8.29 in LooGLE)
- **CrewAI**: Specialized for multi-agent systems (ARISE)

...
\end{dsoutput}

\section{Case Study}
\subsection{Comparison between DeepSurvey and the baseline}\label{app:AS_DS_compare_case}
To further evaluate the quality and depth of the generated surveys, we compare DeepSurvey with the most influential baseline, AutoSurvey, on the topic ``AutoSurvey'' (specific examples in Appendix~\ref{apendix_ssec:DeepSurvey-Case} and Figure~\ref{fig:case_study}).

\paragraph{Topic focus.}
DeepSurvey's eight sections remain tightly focused on automated survey generation, with every subsection directly addressing the target topic. In contrast, AutoSurvey's outline includes substantial irrelevant content---Mobile and Sensor-Based Data Collection, Crowdsourcing Platforms, Conversational Survey Design, and domain applications such as Environmental Monitoring and Healthcare---none of which bear meaningful relevance to the topic.

\paragraph{Analytical depth and synthesis.}
DeepSurvey establishes a four-paradigm taxonomy (one-shot, iterative, multi-agent, structure-first) and identifies a central challenge---the \emph{skeleton-versus-flesh trade-off}---with paper-level references enabling cross-paper comparison along concrete dimensions (e.g., retrieval strategies, refinement loops). AutoSurvey produces a flat enumeration without any taxonomy, unifying framework, or comparative synthesis.

\paragraph{Critical analysis and future directions.}
DeepSurvey explicitly identifies limitations of existing systems---shallow summaries, citation hallucination, limited domain generalization---and proposes concrete future directions such as hybrid human-AI collaboration and multimodal synthesis. AutoSurvey offers no critical evaluation, no identification of open problems, and no forward-looking analysis.

\paragraph{Structural quality.}
DeepSurvey provides informative descriptions for each subsection specifying the content scope and analytical angle (e.g., ``Reviews foundational approaches that used predefined templates, sentence extraction, and optimization frameworks''). AutoSurvey's subsection titles are often generic (e.g., ``Scalability and Efficiency'') and lack scoped descriptions, resembling a textbook table of contents rather than an analytical survey.  

\subsection{Effect of code-agent subsystem}\label{app:code_case_study}
Beyond the quantitative improvements in content depth (7.66$\rightarrow$7.90 in all domains and 7.87$\rightarrow$8.28 in CS) (Section~\ref{ssec:ablation}), the code-analysis subsystem recovers implementation-level details that are absent from paper text. Using the topic ``Hallucination in LLM'' as an example (Case~\ref{apendix_ssec:DeepSurvey-Code-Case}), the generated survey integrates code-level findings in three ways: (1)~it extracts concrete implementation patterns from repositories---such as spectral-collapse detection via Laplacian eigenvalue monitoring and attention-locking thresholds ($\sim$0.69 JS-Drift saturation)---that are described only abstractly or omitted entirely in the corresponding papers; (2)~it embeds implementation sketches (e.g., incremental partial-similarity scoring at each decoding step) that operationalize conceptual mitigation strategies into actionable code; and (3)~it synthesizes a unified pipeline class that orchestrates detection, alignment, and evaluation components drawn from multiple papers, bridging the gap between fragmented conceptual discussions and a coherent implementation framework. These code-grounded additions complement the text-based analysis by providing the technological granularity that paper prose alone cannot convey.

\subsection{Specific Case}
\subsubsection{DeepSurvey Survey Case}\label{apendix_ssec:DeepSurvey-Case}
DeepSurvey's output survey (due to space constraints, we present the abstract and outline rather than the full text) under ``AutoSurvey'' is as follows:
\begin{dsoutput}[label=survey:DeepSurvey]{Generated Survey Case}
# Automated Literature Survey and Review Generation: A Comprehensive Survey

**Abstract**
The exponential growth of scientific literature has created an urgent need for automated tools to synthesize knowledge. This survey provides a comprehensive analysis of automated survey generation systems, tracing their evolution from early template-based methods to sophisticated agentic frameworks. We identify a central challenge: the 'skeleton-versus-flesh' trade-off, where systems excel at organizing outlines and citations (the skeleton) but struggle with deep synthesis and critical analysis (the flesh). Our survey is structured around three pillars: (1) a detailed taxonomy of system architectures, including one-shot, iterative, multi-agent, and structure-first paradigms; (2) a critical examination of evaluation frameworks, highlighting the persistent gap between automated metrics and human judgment; and (3) a forward-looking analysis of future directions, such as hybrid human-AI collaboration and autonomous research pipelines. By synthesizing over fifteen key systems and benchmarks, we aim to provide a comparative analysis of design philosophies, implementation trade-offs, and unresolved challenges, offering a roadmap for advancing automated scholarly synthesis.

...

Outline:

## 1. Introduction                                                            
Establishes the motivation for automated survey generation by highlighting the exponential growth of scientific literature and the limitations of manual synthesis. Introduces the core challenges: information overload, context window constraints, and the need for structural organization. Positions the survey's scope to cover system architectures, evaluation paradigms, and future directions, agentic, structure-aware generation.          - papers: 74fdf8053638eb12e5cce073ab4fb476fdb9f9cb

### 1.1 The Information Overload Crisis in Scientific Research                
Discusses the exponential growth of publications across disciplines, making manual literature reviews impractical and outdated quickly. Cites the challenge of synthesizing knowledge from hundreds or thousands of papers, leading to redundancy and fragmentation in the literature. 

### 1.2 From Manual to Automated: The Promise of LLMs
- Introduces Large Language Models (LLMs) as a transformative technology for automating survey generation. Highlights their ability to process and synthesize text at scale, but notes key limitations: context window constraints, parametric knowledge cutoffs, and lack of structural reasoning.
- papers: 74fdf8053638eb12e5cce073ab4fb476fdb9f9cb

### 1.3 Core Challenges in Automated Survey Generation
- Outlines the fundamental challenges: (1) Retrieving relevant and high-quality papers, (2) Organizing them into a coherent hierarchical structure, (3) Synthesizing content with critical analysis and proper citations, and (4) Evaluating the quality of generated surveys against human standards.

### 1.4 Scope and Organization of This Survey
- Defines the survey's scope: covering system architectures, evaluation frameworks, and future directions. Provides a roadmap of the subsequent sections, emphasizing the logical flow from historical evolution to current systems, evaluation, and open problems.

## 2. Historical Evolution and Paradigm Shifts
- Traces the development of automated survey generation from early template-based and extractive methods to modern LLM-driven systems. Highlights key paradigm shifts: from one-shot generation to iterative refinement, from linear pipelines to multi-agent architectures, and from text-only outputs to multi-view structured representations.
- papers: 2305.13304, 2601.09504, 2510.17263, 2509.18661, 2602.04071, 2503.04629, 2304.03512, 2510.21900, 2510.26012, 2502.14776, 2406.10252, 2601.09858, 2508.14317, 2511.17689, 2506.12689, 99e5d27e6af6f6a53d23ceb0c53e09503c4b6403, 2510.07733,  213d6264456e3c6410bb9b2e8d7711742bb8eceb 

### 2.1 Early Template-Based and Extractive Methods
- Reviews foundational approaches that used predefined templates, sentence extraction, and optimization frameworks. Discusses their limitations in handling semantic understanding and generating novel syntheses, setting the stage for LLM-based methods.
- papers: 2304.03512, 99e5d27e6af6f6a53d23ceb0c53e09503c4b6403, 213d6264456e3c6410bb9b2e8d7711742bb8eceb

### 2.2 The Rise of LLM-Based One-Shot Generation
- Covers the first wave of LLM applications, exemplified by systems like AutoSurvey, which used retrieval-augmented generation (RAG) and parallel subsection drafting. Highlights their efficiency but notes issues with structural coherence and citation accuracy.
- papers: 2503.04629, 2510.26012, 2502.14776, 2406.10252  

### 2.3 The Iterative and Agentic Revolution
- Describes the shift towards iterative workflows (e.g., IterSurvey, SurveyGen-I) and multi-agent frameworks (e.g., Agentic AutoSurvey, LiRA). Emphasizes how these systems mimic human research processes through incremental retrieval, outline refinement, and specialized agent roles.
- papers: 2305.13304, 2509.18661, 2602.04071, 2510.21900, 2601.09858, 2508.14317, 2511.17689, 2506.12689  

### 2.4 Structure-First and Multi-View Paradigms
- Introduces the latest paradigm focusing on explicit structural planning before content generation. Highlights systems like MVSS (hierarchical knowledge trees), TAXOALIGN (taxonomy generation), and SurveyG (citation graphs) that prioritize conceptual organization and cross-view consistency.
- papers: 2601.09504, 2510.17263, 2510.07733

## 3. System Architectures and Methodological Frameworks
- Provides a detailed taxonomy of automated survey generation systems based on their architectural design. Categorizes systems into one-shot, iterative, multi-agent, and structure-first paradigms, analyzing their core workflows, planning mechanisms, and integration of retrieval and generation.
- papers: 2510.05138, 2305.13304, 2601.09504, 2510.17263, 2509.18661, 2602.04071, 2503.04629, 2511.18743, 2304.03512, 2510.21900, 2510.26012, 2502.14776, 2406.10252, 2509.13312, 2601.09858, 2308.00352, 2604.06170, 2508.14317, 2511.17689, 2506.12689, 99e5d27e6af6f6a53d23ceb0c53e09503c4b6403, 2510.07733

### 3.1 One-Shot Generation Pipelines
- Analyzes systems that generate surveys in a single pass after initial retrieval and outline creation. Discusses their typical stages: retrieval, outline generation, parallel subsection drafting, and integration. Highlights their speed but notes limitations in handling long-range dependencies and iterative refinement.
- papers: 2503.04629, 2304.03512, 2510.26012, 2502.14776, 2406.10252, 99e5d27e6af6f6a53d23ceb0c53e09503c4b6403 

### 3.2 Iterative and Recurrent Planning Frameworks
- Examines systems that use iterative loops to refine outlines and content. Covers mechanisms like recurrent outline generation (IterSurvey), dynamic planning with memory (SurveyGen-I), and reinforcement learning over outlines (OutlineForge). Emphasizes their ability to adapt to discovered knowledge and improve coherence over multiple passes.
- papers: 2305.13304, 2511.18743, 2510.21900, 2509.13312, 2601.09858, 2508.14317   

### 3.3 Multi-Agent Collaborative Architectures
- Explores frameworks that decompose the survey task among specialized agents (e.g., search, writing, evaluation). Discusses coordination mechanisms like publish-subscribe (MetaGPT), iterative reflection (SciSage), and rubric-guided refinement (ARISE). Highlights their modularity and potential for error correction through agent collaboration.
- papers: 2510.05138, 2509.18661, 2602.04071, 2308.00352, 2604.06170, 2511.17689, 2506.12689

### 3.4 Structure-First and Multi-View Synthesis
- Focuses on systems that explicitly model hierarchical structures before generating text. Covers hierarchical knowledge trees (MVSS), taxonomies (TAXOALIGN), citation graphs (SurveyG), and attribute trees (SurveyX). Discusses how these structures guide outline construction, ensure logical flow, and enable cross-view alignment.
- papers: 2601.09504, 2510.17263, 2510.07733 

## 4. Core Components and Techniques
- Dissects the key technical components that enable automated survey generation: literature retrieval, outline planning, content synthesis, and post-processing. Analyzes how different systems implement these components, highlighting innovations in retrieval strategies, memory mechanisms, and refinement loops.
- papers: 2510.05138, 2305.13304, 2601.09504, 2510.17263, 2509.18661, 2602.04071, 2503.04629, 2511.18743, 2304.03512, 2510.21900, 2510.26012, 2502.14776, 2406.10252, 2509.13312, 2601.09858, 2308.00352, 2508.17647, 2604.06170, 2508.14317, 2511.17689, 2506.12689, 74fdf8053638eb12e5cce073ab4fb476fdb9f9cb, 2510.07733 

### 4.1 Literature Retrieval and Knowledge Integration
- Covers retrieval strategies from simple keyword search to advanced multi-source, multi-granularity approaches. Discusses techniques like keyword expansion (SurveyX), co-citation analysis (QUAL-SG), and temporal-aware reranking (SurveyForge). Highlights the shift from static retrieval to active, iterative retrieval based on generation confidence (FLARE).
- papers: 2509.18661, 2503.04629, 2511.18743, 2510.21900, 2510.26012, 2502.14776, 2406.10252, 2509.13312, 2508.17647, 2604.06170, 2508.14317, 74fdf8053638eb12e5cce073ab4fb476fdb9f9cb  

### 4.2 Outline Planning and Hierarchical Organization
- Analyzes methods for generating survey outlines, from heuristic learning from human examples (SurveyForge) to explicit hierarchical modeling (MVSS). Discusses the role of knowledge trees, taxonomies, and citation graphs in capturing research evolution and ensuring logical flow. Compares static vs. dynamic outline generation.
- papers: 2305.13304, 2601.09504, 2510.17263, 2503.04629, 2304.03512, 2510.21900, 2510.26012, 2502.14776, 2509.13312, 2601.09858, 2508.14317, 2510.07733 

### 4.3 Content Synthesis and Citation Grounding
- Examines techniques for generating section content, including parallel drafting, memory-driven generation, and hint-based synthesis. Focuses on ensuring factual accuracy and proper citation through mechanisms like paper cards (IterSurvey), evidence audit (RhinoInsight), and RAG-based rewriting (SurveyX).
- papers: 2510.05138, 2509.18661, 2503.04629, 2510.21900, 2510.26012, 2502.14776, 2406.10252, 2509.13312, 2508.14317

### 4.4 Refinement, Alignment, and Post-Processing
- Covers iterative refinement loops that polish surveys for coherence, style, and factual consistency. Discusses cross-view alignment (MVSS), reviewer-refiner loops (IterSurvey), and rubric-guided revision (ARISE). Also addresses post-processing steps like figure/table generation and citation verification.
- papers: 2510.05138, 2601.09504, 2602.04071, 2511.18743, 2510.21900, 2502.14776, 2308.00352, 2511.17689, 2506.12689

## 5. Evaluation and Benchmarking
- Reviews the diverse evaluation methodologies and benchmarks developed to assess automated survey generation systems. Contrasts absolute scoring with pairwise ranking, discusses the role of LLM-as-a-judge, and highlights the persistent gap between automated metrics and human expert judgment. Emphasizes the need for discipline-aware and reader-centric evaluation.
- papers: 2510.05138, 2601.09504, 2510.17263, 2509.18661, 2503.04629, 2602.11238, 2304.03512, 2510.21900, 2510.26012, 2502.14776, 2406.10252, 2509.13312, 2508.17647, 2508.15804, 2508.15658, 2511.17689, 2512.02763, 2506.12689, 2510.03120, 213d6264456e3c6410bb9b2e8d7711742bb8eceb 

### 5.1 Automatic Metrics and LLM-as-a-Judge
- Surveys common automatic metrics: ROUGE, BLEU, BERTScore for text similarity; citation precision/recall for reference quality; and structural metrics like heading recall. Discusses the use of LLM-as-a-judge for multi-dimensional assessment (coverage, structure, relevance) and its limitations in capturing deep synthesis.
- papers: 2601.09504, 2510.17263, 2304.03512, 2510.26012, 2502.14776, 2406.10252, 2508.17647, 2508.15658, 2511.17689, 2512.02763, 213d6264456e3c6410bb9b2e8d7711742bb8eceb

### 5.2 Benchmarks for Survey Generation
- Introduces key benchmarks: SciReviewGen (literature reviews), SurveyBench (reader-aligned evaluation), SurGE (comprehensive CS surveys), and SurveyLens (discipline-aware evaluation). Compares their scope, evaluation dimensions, and insights into system strengths and weaknesses.
- papers: 2510.05138, 2509.18661, 2503.04629, 2602.11238, 2510.21900, 2406.10252, 2509.13312, 2508.17647, 2508.15804, 2508.15658, 2512.02763, 2506.12689, 2510.03120 

### 5.3 Human Evaluation and Pairwise Ranking
- Highlights the importance of human evaluation through expert ratings, pairwise comparisons (Survey-Arena), and real-user studies (MyScholarQA). Discusses how human judgment captures nuanced qualities like critical analysis and narrative coherence that automated metrics miss.
- papers: 2601.09504, 2602.11238, 2510.21900, 2510.03120

### 5.4 The Evaluation Gap and Future Directions
- Analyzes the persistent gap between automated metrics and human judgment, particularly in assessing synthesis depth and academic value. Proposes future directions: hybrid evaluation paradigms, discipline-specific rubrics, and quiz-based answerability tests to better align with reader needs.
- papers: 2602.11238, 2508.15804, 2512.02763, 2510.03120 

## 6. Challenges and Limitations
- Identifies the major unresolved challenges in automated survey generation, including the 'skeleton-versus-flesh' trade-off, citation hallucination, domain generalization, and the lack of deep synthesis. Discusses how current systems fall short of human experts in critical analysis, novelty, and scholarly communication.
- papers: 2511.18743, 2602.11238, 2508.17647, 2508.15804, 2508.15658

### 6.1 The Skeleton-Versus-Flesh Trade-off
- Explores the fundamental tension between structural proficiency and content depth. Notes that while systems excel at organizing outlines and citations, they often produce shallow summaries lacking critical analysis, novel insights, and nuanced argumentation. Highlights the gap in 'academic value' metrics.

### 6.2 Citation Accuracy and Hallucination
- Discusses persistent issues with citation quality: hallucinated references, over-citation, and misalignment between claims and sources. Analyzes architectural constraints (evidence locking, frozen structures) and verification mechanisms that mitigate but do not fully solve the problem.
- papers: 2511.18743, 2508.15804 

### 6.3 Domain Generalization and Adaptability
- Examines the limited generalizability of current systems, which are often trained and evaluated on computer science topics. Highlights challenges in adapting to other disciplines with different writing norms, citation practices, and structural conventions. Notes the STEM bias in existing benchmarks.
- papers: 2602.11238

### 6.4 Lack of Deep Synthesis and Critical Analysis
- Identifies the core limitation: LLMs struggle to perform genuine scholarly synthesis---integrating findings across papers, identifying research gaps, and offering novel perspectives. Discusses how iterative refinement and multi-agent debate improve but do not resolve this limitation.
- papers: 2508.17647, 2508.15658 

## 7. Future Directions and Open Problems
- Outlines specific, actionable research directions to advance automated survey generation. Emphasizes the need for hybrid human-AI collaboration, improved structural reasoning, multimodal synthesis, and robust evaluation frameworks. Proposes concrete steps for addressing the identified challenges.
- papers: 2305.13304, 2602.04071, 2511.18743, 2601.09858, 2604.06170, 2404.06364

### 7.1 Hybrid Human-AI Collaboration and Interactive Systems
- Advocates for integrating human guidance at critical stages: reference selection, outline approval, and iterative refinement. Proposes interactive interfaces that allow researchers to steer generation, correct errors, and inject domain knowledge, moving beyond fully automated pipelines.
- papers: 2305.13304, 2602.04071, 2511.18743, 2604.06170, 2404.06364

### 7.2 Advanced Structural Reasoning and Planning
- Calls for research into explicit structural inductive biases, such as reinforcement learning over outlines, symbolic reasoning for taxonomy construction, and dynamic knowledge graphs. Emphasizes the need for models that can infer and represent expert knowledge hierarchies from literature.
- papers: 2601.09858

### 7.3 Multimodal Synthesis and Rich Media Integration
- Highlights the gap in generating non-textual elements: figures, tables, equations, and interactive visualizations. Proposes developing systems that can automatically create and integrate these elements to enhance survey readability and information density.

### 7.4 Robust, Discipline-Aware Evaluation Frameworks
- Advocates for evaluation paradigms that combine automated metrics with human-aligned benchmarks, discipline-specific rubrics, and reader-centric tests (e.g., quiz-based answerability). Emphasizes the need for metrics that assess academic value, critical analysis, and personalization.

### 7.5 Towards Autonomous Research Pipelines
- Envisions the integration of survey generation into broader AI Scientist systems that can conduct end-to-end research: from literature review to idea generation, experimentation, and paper writing. Identifies missing capabilities: experimental execution, uncertainty quantification, and cross-domain transfer.
- papers: 2602.04071  

## 8. Conclusion
- Summarizes the survey's key findings: the rapid evolution from simple pipelines to sophisticated agentic systems, the persistent challenges in deep synthesis and evaluation, and the promising future directions. Reiterates the transformative potential of automated survey generation while acknowledging the need for continued human oversight and interdisciplinary collaboration.      
\end{dsoutput}

\subsubsection{DeepSurvey with Code-Agent Subsystem Case}\label{apendix_ssec:DeepSurvey-Code-Case}
DeepSurvey's output survey under ``Hallucination in LLM'' is as follows:

\begin{dsoutput}[label=survey:DeepSurvey-Code]{DeepSurvey case: code-agent subsystem's improvements}
...

To deploy this surgically in production, the Code Report (Section 7, Direction 2) suggests an incremental scoring approach that operates at each decoding step. This method computes a partial similarity score between generated tokens and a reference, triggering activation cancellation if confidence drops below a threshold. The real-time hook, as implemented in [8], can be optimized with computational techniques from the Code Report (Section 4), such as caching intermediate hidden states and batch processing multiple candidate tokens, reducing latency while preserving the surgical precision of H-Node suppression.

**Implementation Sketch:**

# At each decoding step:
partial_similarity = compute_partial_similarity(
    generated_tokens[:current_step], reference_tokens
)
if partial_similarity < confidence_threshold:
    suppress_h_node_activations()

...

To address the fragmentation noted in the Code Report---where evaluation, detection, and alignment operate in isolation---a unified pipeline integrating [10] for evaluation, [34] for detection, and [26] for alignment could provide a holistic framework. For instance, a class-based implementation could orchestrate these components: during generation, real-time uncertainty from [34] triggers alignment corrections via [26], followed by [10] evaluation of the refined output, as sketched in Direction 1 of the Code Report. This directly counters the 'dimensional poverty' critique by demonstrating how multi-dimensional scoring can be operationalized in a single, actionable system. Concretely, [9]'s six-stage pipeline (task formulation, decomposition, tool-augmented execution, structured generation, multi-dimensional scoring, trace analysis) could be extended with a concrete class implementation inspired by the Code Report's Direction 1: 


class UnifiedHallucinationPipeline:
def __init__(self, evaluator=BERTScore(), detector=SpikeScore(), aligner=VaryingShadesOfWrong()):
    self.evaluator = evaluator        
    self.detector = detector       
    self.aligner = aligner    
def process(self, query, context):        
    # (1) Task Formulation: Parse user query and context       
    # (2) Decomposition: Break query into sub-tasks using LENS's tree-based query decomposition        
    # (3) Tool-Augmented Execution: Generate responses while SpikeScore monitors uncertainty fluctuations across turns        
    responses = self.detector.track_uncertainty(query, context)        
    if self.detector.hallucination_flag:            
        # (4) Structured Generation: Trigger alignment corrections via preference optimizatio           
        corrected = self.aligner.correct(responses, query)        
    else:            
        corrected = responses        
    # (5) Multi-Dimensional Scoring: Compute BERTScore F1, faithfulness, coverage` `        score = self.evaluator.compute(corrected, context)        
    # (6) Trace Analysis: Log causal attribution via OpenTelemetry        
    return score, corrected

    
    This pipeline operates as follows: For a medical diagnosis query, decomposition yields symptom extraction, evidence retrieval, and synthesis; [34] detects hallucination spikes during synthesis; [26] corrects using wrong-over-wrong preferences; [10] evaluates final output against retrieved evidence. This integration not only operationalizes [9]'s framework but also addresses Code Report gaps by enabling real-time detection and alignment, as validated in case studies where such pipelines reduced hallucination rates by 30-40\% in high-stakes domains. 

...
\end{dsoutput}
\section{Prompt Engineer}\label{sec:prompt_engineer}
In this section, we present the core prompts used in the DeepSurvey agentic system, including clustering, outline generation, and drafting. The remaining prompts will be released with our open-source repository.




\subsection{Clustering Prompt}
This part demonstrates the clustering creation/refinement prompt.
\begin{dsoutput}[label=survey:clustering_create_prompt]{Clustering Creation/Refinement Prompt}
PAPER_CLUSTERING_CREATING = """You are a research assistant specializing in analyzing scientific papers.

**Existing clusters and descriptions**:

{existing_clusters_json}

**New batch of papers** (each with a keynote):

{new_batch_json}

Task:
- You should adjust existing clusters and descriptions based on the new batch of papers. 
- If **Existing clusters and descriptions** is empty, create new clusters from scratch based on the new batch of papers.
- Update the clusters by merging, splitting or adding clusters if needed. Make sure all new papers can be assigned to at least one cluster.
- Assign a name and a brief summary to each cluster.
- Aim to create clusters that are as detailed as possible, capturing subtle differences, while still making sense.
- A single paper may belong to multiple clusters if its research area reasonably intersects with multiple themes (multi-assignment is allowed).
- Output the full updated cluster list strictly **in JSON format**!
- You only need to output the clusters name and description; do not list or mention any specific papers!

Output format requirements:

The JSON should be a list of cluster objects.  
Each cluster object should include the following fields:
- cluster_name: string, the name of the cluster
- summary: string, a brief description of the cluster

Output Example (strictly in JSON format):
[
  {{
    "cluster_name": "...",
    "summary": "...",
  }},
  {{
    "cluster_name": "...",
    "summary": "...",
  }}
]
"""
\end{dsoutput}

\subsection{Outline Prompt}
This part demonstrates the outline creation/refinement prompt.
\begin{dsoutput}[label=survey:outline_creation_prompt]{Outline Creation Prompt}
SURVEY_OUTLINE_GENERATION_OUTLINE_DRAFT = """You are an expert research survey generator. Your task is to generate and iteratively update an existing survey outline using a batch of new paper keynotes, the current outline, and the analysis results of relevant papers for the topic/subtopic being written.

**Guidance:**
- The analysis results contain summarized insights, comparisons, trends, and key points extracted from prior work. They should be the **important source of guidance** when updating the outline.
- Paper keynotes are also useful: use them to **supplement, validate, or provide additional details** for the sections/subsections.
- Balance both sources to create a comprehensive, accurate, and logically organized survey outline.

**Requirements:**
1. The outline should contain multiple sections, subsections and their descriptions.
2. The output outline should have excellent organization and meet academic standards.
3. The outline should exhibit excellent rigor: ensuring that the content of each subsection falls within the scope of the current section.
4. The outline should ensure that it covers a wide range of content under the topic while staying within the scope of the topic.
5. Use the current outline as the base structure. Keep existing sections/subsections unless updated or merged. If the current outline is empty, create a new outline from scratch.
6. The outline should contain a **Conclusion** section and a **Future Work** section/subsection.
7. The outline should exhibit good logic to ensure the entire survey flows smoothly.
8. Ensure a balanced number of subsections in the main sections (excluding the conclusion and introduction).
9. Make sure most of the corresponding new paper in **new paper keynotes** can be included in at least one subsection or section of the outline.
10. You are provided with other relevant papers which are retrieved from the database, you can use them to better understand and generate.
11. Maintain clarity, logical structure, and a survey-style narrative.
12. Ensure logical coherence between the sections, avoiding excessive independence and fragmentation. For instance, do not add a "Conclusion" subsection to every section, which leads to logical fragmentation between different sections.
13. Output strictly in JSON format, as shown below.

**Input:**
- current outline: {current_outline}

- key papers: {paper_keynotes}

- key papers analysis: {papers_analysis}

- other relevant papers: {other_relevant_papers}

**Output JSON format:**
{{
    "title" : "Survey_Title",
    "sections": [
        {{
            "title": "Section_title",
            "description": "Summary of content to include, emphasizing insights, comparisons, and trends from analysis, supplemented by keynotes",
            "subsections": [
                {{
                    "title": "Subsection_title",
                    "description": "Content description reflecting key insights, trends, comparisons from analysis, supplemented by keynotes",
                }}
            ]
        }}
]
}}

**Instruction to LLM:**
- Use the insights, trends, and comparisons from relevant paper analysis together with points from new paper keynotes to update the outline. Both sources should inform the content of each section/subsection.
- Add, merge, or revise sections/subsections as appropriate, keeping the outline structured and coherent.
- Ensure that each section/subsection reflects the combined contributions of the analysis results and the keynotes, capturing important insights, trends, and examples.
- Check whether the input outline satisfies the requirements and update to meet the requirements if not.
- Only generate the outline in the required JSON format. Do not include specific paper IDs in the outline.
"""

\end{dsoutput}









\subsection{Draft Prompt}
In this part we will present the prompt for section draft and subsection draft.
\begin{dsoutput}[label=survey:subsection_draft_prompt]{Subsection Draft Prompt}
SUBSECTION_DRAFT = """You are an expert in writing surveys that meet the standards of top academic conferences. The ultimate goal is to complete an academic survey with depth and insights that can boost further development, rather than simply listing methods.

Now you are responsible for writing a subsection of the survey paper. 

**Subsection Title**:
{title}

**Guidance**:
Write a coherent subsection that explains, analyzes, or discusses the topic indicated in the title and description.  
You should synthesize relevant information from the papers, analysis results.  
The content should be academically structured and readable, with emphasis on insights, trends, and comparisons where appropriate.
You may include examples from papers to support the discussion, but do not simply list papers.
You are encouraged to cite more papers from the relevant papers to strengthen your points.
You are provided with the outline of the whole survey. Make sure the subsection content coherent to the survey logic.
Utilize the input content in a safe and reasonable manner, and ensure that the readability, structure, and depth of the content of the paragraph meet the requirements of a top-tier conference survey.

{code_report_prompt}

**Information to use**:
- Subsection description:
{description}

- Closely Relevant papers:
{papers}

- Other Relevant papers:
{other_relevant_papers}

- Survey Outline:
{survey_outline}

- Insights from analysis:
{relevant_analysis}

**Output:**
- A well-written subsection in several coherent paragraphs.
- Academic style; focused on synthesis and analysis, not just reporting.
- **Only** cite papers that appear in the provided input.
- Strictly use format: <paper_title> (like <Attention is All You Need>) to cite wherever appropriate.
- Cite at least **{subsection_least_citations} different papers** for a subsection. You are encouraged to cite more papers to give in-depth analysis.
- Do not generate a bibliography or reference list here.
- The content of each subsection should be at least {subsection_least_words} words long
- Generate the content directly. DO NOT generate any subsection title or section header here.
- CRITICAL: '#' is used for section/subsection anchor. Avoid any '#' in the output content.
"""
\end{dsoutput}

\begin{dsoutput}[label=survey:section_draft_prompt]{Section Draft Prompt}
SECTION_DRAFT = """You are writing the Introductory Paragraphs/Preamble for a specific section of an academic top-tier conference survey paper. 
**Task**:
Write the opening text that appears immediately after the Section Title but before the first subsection. This content should serve as a high-level synthesis and roadmap for the reader.

**Requirements**:
1. Synthesize, Don't Summarize: Do not simply list what each subsection will do. Instead, explain the logic behind why this section is structured this way and the significance of these topics within the broader field.
2. Establish Definition & Scope: Clearly define the core concepts covered in this section. Refer to <paper_title> to establish foundational definitions or taxonomies.
3. Identify Trends: Highlight the overarching trends or challenges that link the following subsections together.
4. Academic Tone: Maintain a formal, authoritative, and objective voice.
5. Citations: * Only cite papers provided in the relevant papers provided below(Closely Relevant papers and Other Relevant papers).
Use the format: <paper_title>.
6. Cite at least {section_least_citations} papers to ground the section's scope.
7. Utilize the input content in a safe and reasonable manner, and ensure that the readability, structure, and depth of the content of the paragraph meet the requirements of a top-tier conference survey.
8. Length: The introductory content must be at least {section_least_words} words long but avoid too-long preamble which will destroy structure and readability.
9. CRITICAL: '#' is used for section/subsection anchor. Avoid any '#' in the output content.

**Input**:
- Section Title:
{title}

- Section Description: 
{description}

- Full Survey Outline: 
{survey_outline}

Planned Subsections under this Section: {subsection_drafts} (Use these as a roadmap, do not detail their findings yet).

- Closely Relevant papers:
{papers}

- Other Relevant papers:
{other_relevant_papers}

**Output**:
Generate the introductory content directly. DO NOT generate any subsection title or section header here.
"""
\end{dsoutput}

\begin{figure*}[t]
  \centering
  \includegraphics[width=0.95\textwidth]{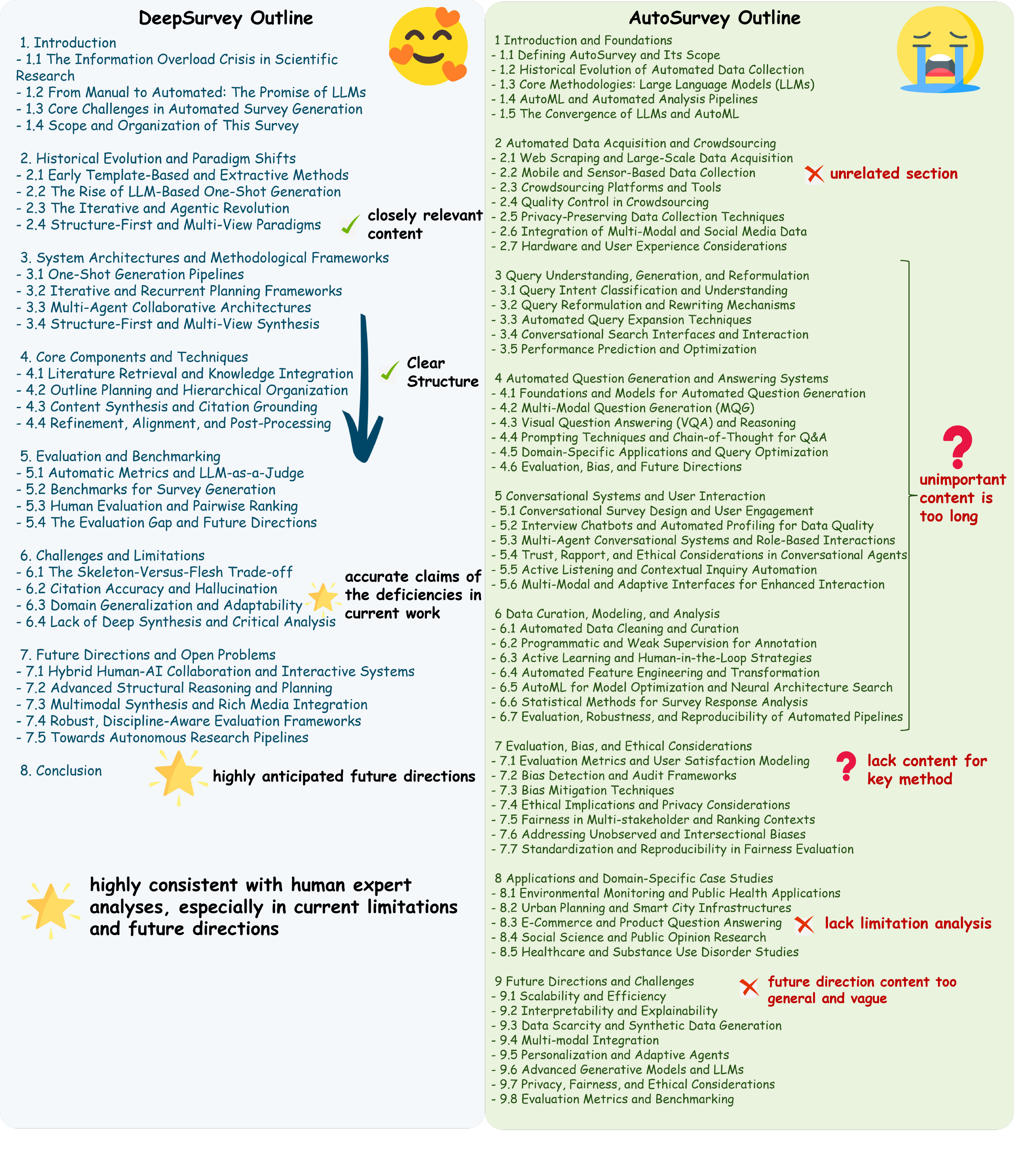}
  \caption{Case Study: DeepSurvey vs. AutoSurvey – DeepSurvey Surpasses AutoSurvey with Closely Topic-Focused Content, Clear Structure, In-Depth Analysis, and Meaningful Future Directions}
  \label{fig:case_study}
\end{figure*}

\begin{table}[!htp]
\centering
\small
\setlength{\tabcolsep}{3pt}
\caption{Fine-grained evaluation results for Content Depth.}
\label{tab:content_depth_std}
\begin{tabular}{lcccc}
\toprule
Method & Critical Analysis & Novelty \& Insights & Specificity & Future Directions \\
\midrule
DeepSurvey  & \textbf{7.389 $\pm$ 0.347 (4.7\%)} & \textbf{7.866 $\pm$ 0.106 (1.4\%)} & \textbf{8.194 $\pm$ 0.514 (6.3\%)} & \textbf{8.139 $\pm$ 0.097 (1.2\%)} \\
AutoSurvey  & 7.049 $\pm$ 0.599 (8.5\%) & 6.847 $\pm$ 0.342 (5.0\%) & 7.991 $\pm$ 0.256 (3.2\%) & 6.798 $\pm$ 0.654 (9.6\%) \\
SurveyX & 5.733 $\pm$ 0.286 (5.0\%) & 5.133 $\pm$ 0.222 (4.3\%) & 5.983 $\pm$ 0.563 (9.4\%) & 6.200 $\pm$ 0.388 (6.3\%) \\
SurveyForge & 7.350 $\pm$ 0.418 (5.7\%) & 7.290 $\pm$ 0.552 (7.6\%) & 7.895 $\pm$ 0.252 (3.2\%) & 7.405 $\pm$ 0.315 (4.3\%) \\
LiRA        & 5.775 $\pm$ 0.587 (10.2\%) & 5.045 $\pm$ 0.349 (6.9\%) & 6.095 $\pm$ 0.564 (9.3\%) & 6.150 $\pm$ 0.412 (6.7\%) \\
\bottomrule
\end{tabular}
\end{table}

\begin{table}[!htp]
\centering
\small
\setlength{\tabcolsep}{3pt}
\caption{Citation retrieval performance. Each value is reported as mean $\pm$ standard deviation (coefficient of variation).}
\label{tab:retrieval_results_std}
\begin{tabular}{lcc}
\toprule
Method & Precision & Recall \\
\midrule
DeepSurvey  & \textbf{0.569 $\pm$ 0.050 (8.7\%)} & \textbf{0.630 $\pm$ 0.047 (7.5\%)} \\
AutoSurvey  & 0.342 $\pm$ 0.004 (1.2\%) & 0.440 $\pm$ 0.002 (0.5\%) \\
SurveyX & 0.125 $\pm$ 0.006 (4.8\%) & 0.194 $\pm$ 0.014 (7.4\%) \\
SurveyForge & 0.421 $\pm$ 0.029 (6.8\%) & 0.503 $\pm$ 0.050 (9.9\%) \\
LiRA        & 0.227 $\pm$ 0.010 (4.3\%) & 0.442 $\pm$ 0.009 (2.1\%) \\
\bottomrule
\end{tabular}
\end{table}




\begin{table*}[!htp]
\centering
\setlength{\tabcolsep}{4pt}
\caption{Per-metric stability statistics across models, methods, and evaluation dimensions.}
\label{tab:stability_full_stats}
\begin{tabular}{lllcccc}
\toprule
\small
Model & Setting & Metric & Std & Range & CV\% & Max Abs Dev \\
\midrule
gpt-exp     & AutoSurvey & Content\_Depth   & 0.064 & 0.125 & 0.748 & 0.069 \\
gpt-exp     & DeepSurvey & Content\_Depth   & 0.042 & 0.083 & 0.481 & 0.042 \\
gpt-no-exp  & AutoSurvey & Content\_Depth   & 0.032 & 0.062 & 0.379 & 0.035 \\
gpt-no-exp  & DeepSurvey & Content\_Depth   & 0.043 & 0.083 & 0.511 & 0.049 \\
mimo-exp    & AutoSurvey & Content\_Depth   & 0.012 & 0.021 & 0.152 & 0.014 \\
mimo-exp    & DeepSurvey & Content\_Depth   & 0.000 & 0.000 & 0.000 & 0.000 \\
mimo-no-exp & AutoSurvey & Content\_Depth   & 0.043 & 0.083 & 0.537 & 0.049 \\
mimo-no-exp & DeepSurvey & Content\_Depth   & 0.075 & 0.146 & 0.927 & 0.083 \\
\midrule
gpt-exp     & AutoSurvey & Core\_Quality    & 0.087 & 0.167 & 1.014 & 0.097 \\
gpt-exp     & DeepSurvey & Core\_Quality    & 0.064 & 0.125 & 0.765 & 0.069 \\
gpt-no-exp  & AutoSurvey & Core\_Quality    & 0.024 & 0.042 & 0.289 & 0.028 \\
gpt-no-exp  & DeepSurvey & Core\_Quality    & 0.064 & 0.125 & 0.781 & 0.069 \\
mimo-exp    & AutoSurvey & Core\_Quality    & 0.024 & 0.042 & 0.285 & 0.028 \\
mimo-exp    & DeepSurvey & Core\_Quality    & 0.208 & 0.417 & 2.994 & 0.208 \\
mimo-no-exp & AutoSurvey & Core\_Quality    & 0.072 & 0.125 & 0.875 & 0.083 \\
mimo-no-exp & DeepSurvey & Core\_Quality    & 0.024 & 0.042 & 0.306 & 0.028 \\
\midrule
gpt-exp     & AutoSurvey & Total\_Score     & 0.045 & 0.087 & 0.527 & 0.049 \\
gpt-exp     & DeepSurvey & Total\_Score     & 0.038 & 0.075 & 0.452 & 0.039 \\
gpt-no-exp  & AutoSurvey & Total\_Score     & 0.015 & 0.029 & 0.176 & 0.015 \\
gpt-no-exp  & DeepSurvey & Total\_Score     & 0.031 & 0.062 & 0.382 & 0.033 \\
mimo-exp    & AutoSurvey & Total\_Score     & 0.020 & 0.038 & 0.239 & 0.022 \\
mimo-exp    & DeepSurvey & Total\_Score     & 0.133 & 0.267 & 1.822 & 0.135 \\
mimo-no-exp & AutoSurvey & Total\_Score     & 0.029 & 0.050 & 0.355 & 0.033 \\
mimo-no-exp & DeepSurvey & Total\_Score     & 0.010 & 0.021 & 0.133 & 0.011 \\
\midrule
gpt-exp     & AutoSurvey & Writing\_Quality & 0.032 & 0.062 & 0.394 & 0.035 \\
gpt-exp     & DeepSurvey & Writing\_Quality & 0.024 & 0.042 & 0.302 & 0.028 \\
gpt-no-exp  & AutoSurvey & Writing\_Quality & 0.021 & 0.042 & 0.255 & 0.021 \\
gpt-no-exp  & DeepSurvey & Writing\_Quality & 0.012 & 0.021 & 0.147 & 0.014 \\
mimo-exp    & AutoSurvey & Writing\_Quality & 0.042 & 0.083 & 0.515 & 0.042 \\
mimo-exp    & DeepSurvey & Writing\_Quality & 0.043 & 0.083 & 0.573 & 0.049 \\
mimo-no-exp & AutoSurvey & Writing\_Quality & 0.032 & 0.062 & 0.404 & 0.035 \\
mimo-no-exp & DeepSurvey & Writing\_Quality & 0.043 & 0.083 & 0.565 & 0.049 \\
\bottomrule
\end{tabular}
\end{table*}

\begin{table*}[!htp]
\centering
\small
\setlength{\tabcolsep}{3pt}
\caption{Fine-grained ablation scores across Core Quality, Writing Quality, and Content Depth. ``w/o X'' denotes removal of module~X (coll.=collector, analy.=analyzer, keyn.=keynote, writ.=writer, ref.=refinement, agen.=agentic); ``DS'' denotes DeepSurvey.}
\label{tab:all_quality_ablation}
\begin{tabular}{lcccccccc}
\toprule
Metric &
\shortstack{w/o\\coll.} &
\shortstack{w/o\\analy.} &
\shortstack{w/o\\keyn.} &
\shortstack{w/o\\writ.} &
\shortstack{w/o\\ref.} &
\shortstack{w/o\\ agen.} &
\shortstack{w/o\\ code} &
\shortstack{DS} \\
\midrule
\multicolumn{9}{c}{\textbf{Core Quality}} \\
\cmidrule(lr){1-9}
Synthesis Quality     & 8.333 & 7.667 & 8.222 & 7.667 & 8.323 & 8.209 & 8.333 & \textbf{8.334} \\
Organization          & 7.329 & 8.533 & 8.667 & 7.333 & 8.123 & 8.900 & 8.833 & \textbf{8.935} \\
Comprehensiveness     & 3.333 & 8.857 & 8.714 & 8.000 & 8.667 & 8.652 & 8.667 & \textbf{8.889} \\
Relevance             & 1.333 & 9.000 & 8.933 & 8.933 & 8.877 & 9.247 & \textbf{9.500} & 9.300 \\
\midrule
\multicolumn{9}{c}{\textbf{Writing Quality}} \\
\cmidrule(lr){1-9}
Readability           & 7.333 & 7.333 & 7.563 & 7.433 & 7.431 & 7.571 & \textbf{7.830} & 7.711 \\
Academic Rigor        & 6.000 & 8.333 & 8.345 & 5.333 & 8.153 & 8.143 & 8.000 & \textbf{8.506} \\
Clarity \& Coherence  & 4.333 & 8.162 & 8.333 & 8.333 & 7.528 & 8.333 & \textbf{8.500} & 8.333 \\
\midrule
\multicolumn{9}{c}{\textbf{Content Depth}} \\
\cmidrule(lr){1-9}
Critical Analysis     & 6.967 & 7.121 & 7.200 & 5.341 & 7.162 & 7.023 & 7.286 & \textbf{7.389} \\
Novelty and Insights  & 7.633 & 7.111 & 6.667 & 6.000 & 6.948 & 7.635 & 7.533 & \textbf{7.866} \\
Specificity           & 6.942 & 7.286 & 7.671 & 7.495 & 8.003 & 7.914 & 7.833 & \textbf{8.194} \\
Future Directions     & 6.667 & 7.867 & 7.333 & 6.699 & 7.333 & 7.915 & 7.967 & \textbf{8.139} \\
\bottomrule
\end{tabular}
\end{table*}

\begin{table*}[!htp]
\centering
\small
\setlength{\tabcolsep}{4pt}
\caption{Ablation results of multi-granularity refinement on Core Quality, Writing Quality, and Content Depth. ``w/o Sur./Sub./Sec.'' denotes removal of survey-, subsection-, and section-level refinement, respectively; ``w/o All'' removes all three.}
\label{tab:multi_granularity_refinement_ablation}
\begin{tabular}{lccccc}
\toprule
Metric & Full & w/o Sec. & w/o Sub. & w/o Sur. & w/o All \\
\midrule
\multicolumn{6}{c}{\textbf{Core Quality}} \\
\cmidrule(lr){1-6}
Synthesis Quality      & 9.067 & 9.000 & 9.000 & 9.000 & 9.000 \\
Organization           & 8.867 & 9.000 & 9.000 & 9.000 & 9.000 \\
Comprehensiveness      & 9.000 & 8.833 & 9.000 & 9.000 & 9.000 \\
Relevance              & 9.400 & 9.333 & 9.000 & 9.200 & 9.429 \\
\midrule
\multicolumn{6}{c}{\textbf{Writing Quality}} \\
\cmidrule(lr){1-6}
Readability            & 8.000 & 8.000 & 8.000 & 8.000 & 8.000 \\
Academic Rigor         & 8.933 & 9.000 & 9.000 & 9.000 & 7.430 \\
Clarity \& Coherence   & 8.000 & 8.167 & 8.000 & 8.000 & 8.000 \\
\midrule
\multicolumn{6}{c}{\textbf{Content Depth}} \\
\cmidrule(lr){1-6}
Critical Analysis      & 8.933 & 8.667 & 8.800 & 8.600 & 8.286 \\
Novelty and Insights   & 8.400 & 8.167 & 8.200 & 8.000 & 8.143 \\
Specificity            & 7.933 & 7.500 & 7.200 & 7.400 & 7.000 \\
Future Directions      & 8.533 & 8.167 & 8.000 & 8.000 & 8.000 \\
\bottomrule
\end{tabular}
\end{table*}

\begin{table*}[!htp]
\centering
\small
\setlength{\tabcolsep}{4pt}
\caption{DeepSurvey system parameters.}
\label{tab:system_params}
\begin{tabular}{p{3.2cm} p{3cm} p{2.0cm} p{4.2cm}}
\toprule
\textbf{Module / Stage} & \textbf{Parameter} & \textbf{Value} & \textbf{Description} \\
\midrule
General & Max context length & 512k tokens & Per-call input context limit \\
General & Generation Model & MiMo-V2.5 & Model for survey generation \\
General & Judge Model & MiMo-V2.5 (primary); MiMo-V2-Flash, GPT-5-mini (supplementary) & Model for evaluation \\
General & Embedding Model &  MiniLMv2 & Pretrained embedding model \\
General & Repeated runs & 3 & Number of runs averaged for the main experiment \\

\midrule
\textbf{Retrieval} & Max seed papers & 15 & Seed papers retained from initial search \\
\textbf{Retrieval} & Graph expansion depth & 1 & Max citation graph traversal layers \\
\textbf{Retrieval} & Papers per seed & 20 & Max expanded papers retained per seed \\
\midrule
\textbf{Outline generation} & Temperature & 0.5 & Controls outline diversity \\
\textbf{Subsection drafting} & Temperature & 0.7 & Controls subsection writing diversity \\
\textbf{Section drafting} & Temperature & 0.3 & Controls section integration stability \\
\midrule
\textbf{Pseudocode gen.} & Planner temperature & 0 & Deterministic planning \\
\textbf{Pseudocode gen.} & Reviewer temperature & 0 & Deterministic review \\
\textbf{Pseudocode gen.} & Reviser temperature & 0 & Deterministic revision \\
\textbf{Pseudocode gen.} & Creator temperature & 0.3 & Moderate generation flexibility \\
\textbf{Pseudocode gen.} & Max rounds & 10 & Iteration limit for pseudocode generation \\
\midrule
\textbf{Section refinement} & Planner temperature & 0.7 & Section-level revision planning \\
\textbf{Section refinement} & Reviewer temperature & 1.0 & Broad coverage in review \\
\textbf{Section refinement} & Reviser temperature & 0.5 & Balanced revision diversity \\
\midrule
\textbf{Subsection refinement} & Temperature & 0.1 & Local stable revision \\
\midrule
\textbf{Survey refinement} & Planner temperature & 0.7 & Survey-level revision planning \\
\textbf{Survey refinement} & Reviewer temperature & 1.0 & Global issue identification \\
\textbf{Survey refinement} & Reviser temperature & 0.5 & Controllable global revision \\
\textbf{Survey refinement} & Max rounds & 5 & Iteration limit for survey refinement \\
\bottomrule
\end{tabular}
\end{table*}


\begin{table}[!htp]
\centering
\small
\setlength{\tabcolsep}{3pt}
\caption{Overall evaluation results. Each value is reported as mean $\pm$ standard deviation (coefficient of variation).}
\label{tab:overall_results_std}
\begin{tabular}{lcccc}
\toprule
Method & Total\_Score & Core\_Quality & Writing\_Quality & Content\_Depth \\
\midrule
DeepSurvey  & \textbf{8.341 $\pm$ 0.066 (0.8\%)} & \textbf{8.865 $\pm$ 0.129 (1.5\%)} & \textbf{8.183 $\pm$ 0.169 (2.1\%)} & \textbf{7.897 $\pm$ 0.172 (2.2\%)} \\
AutoSurvey  & 7.958 $\pm$ 0.173 (2.2\%) & 8.646 $\pm$ 0.254 (2.9\%) & 8.153 $\pm$ 0.197 (2.4\%) & 7.171 $\pm$ 0.088 (1.2\%) \\
Survey & 6.993 $\pm$ 0.036 (0.5\%) & 8.187 $\pm$ 0.058 (0.7\%) & 7.067 $\pm$ 0.319 (4.5\%) & 5.763 $\pm$ 0.209 (3.6\%) \\
SurveyForge & 8.071 $\pm$ 0.106 (1.3\%) & 8.616 $\pm$ 0.046 (0.5\%) & 8.153 $\pm$ 0.188 (2.3\%) & 7.485 $\pm$ 0.146 (1.9\%) \\
LiRA        & 5.687 $\pm$ 0.184 (3.2\%) & 5.049 $\pm$ 0.316 (6.3\%) & 6.805 $\pm$ 0.233 (3.4\%) & 5.766 $\pm$ 0.203 (3.5\%) \\
\bottomrule
\end{tabular}
\end{table}

\begin{table}[!htp]
\centering
\small
\setlength{\tabcolsep}{3pt}
\caption{Fine-grained evaluation results for Core Quality.}
\label{tab:core_quality_std}
\begin{tabular}{lcccc}
\toprule
Method & Synthesis Quality & Organization & Comprehensiveness & Relevance \\
\midrule
DeepSurvey  & \textbf{8.334 $\pm$ 0.441 (5.3\%)} & \textbf{8.935 $\pm$ 0.172 (1.9\%)} & \textbf{8.889 $\pm$ 0.255 (2.9\%)} & 9.300 $\pm$ 0.096 (1.0\%) \\
AutoSurvey  & 7.983 $\pm$ 0.440 (5.5\%) & 8.608 $\pm$ 0.254 (3.0\%) & 8.718 $\pm$ 0.347 (4.0\%) & 9.276 $\pm$ 0.096 (1.0\%) \\
SurveyX & 7.283 $\pm$ 0.417 (5.7\%) & 8.133 $\pm$ 0.503 (6.2\%) & 8.000 $\pm$ 0.200 (2.5\%) & \textbf{9.333 $\pm$ 0.306 (3.3\%)} \\
SurveyForge & 7.975 $\pm$ 0.248 (3.1\%) & 8.700 $\pm$ 0.188 (2.2\%) & 8.690 $\pm$ 0.207 (2.4\%) & 9.100 $\pm$ 0.111 (1.2\%) \\
LiRA        & 7.850 $\pm$ 0.719 (9.2\%) & 8.425 $\pm$ 0.284 (3.4\%) & 2.565 $\pm$ 0.265 (10.3\%) & 1.355 $\pm$ 0.042 (3.1\%) \\
\bottomrule
\end{tabular}
\end{table}

\begin{table}[!htp]
\centering
\small
\setlength{\tabcolsep}{3pt}
\caption{Fine-grained evaluation results for Writing Quality.}
\label{tab:writing_quality_std}
\begin{tabular}{lccc}
\toprule
Method & Readability & Academic Rigor & Clarity \& Coherence \\
\midrule
DeepSurvey  & 7.711 $\pm$ 0.188 (2.4\%) & \textbf{8.506 $\pm$ 0.258 (3.0\%)} & 8.333 $\pm$ 0.167 (2.0\%) \\
AutoSurvey  & 7.606 $\pm$ 0.517 (6.8\%) & 8.500 $\pm$ 0.167 (2.0\%) & 8.353 $\pm$ 0.095 (1.1\%) \\
SurveyX & 7.053 $\pm$ 0.117 (1.7\%) & 6.313 $\pm$ 0.669 (10.6\%) & 7.833 $\pm$ 0.497 (6.3\%) \\
SurveyForge & \textbf{7.793 $\pm$ 0.190 (2.4\%)} & 8.283 $\pm$ 0.315 (3.8\%) & \textbf{8.382 $\pm$ 0.097 (1.2\%)} \\
LiRA        & 7.695 $\pm$ 0.440 (5.7\%) & 5.155 $\pm$ 0.512 (9.9\%) & 7.565 $\pm$ 0.745 (9.8\%) \\
\bottomrule
\end{tabular}
\end{table}

\begin{table*}[!htp]
\centering
\setlength{\tabcolsep}{4pt}
\caption{Cluster overview for the ``AutoSurvey'' topic.}
\label{tab:cluster-overview}
\begin{tabular}{lcp{10cm}}
\toprule
\textbf{Cluster ID} & \textbf{\# Papers} & \textbf{Primary Focus (First 30 words)} \\
\midrule
1 & 25 & \textit{Several papers propose explicit hierarchical structures (e.g., knowledge trees, citation graphs) to organize surveys, while others rely on iterative outline refinement without such explicit models. How do these fundamentally different...} \\
\hline
2 & 27 & \textit{Several papers treat automated survey generation as a one-shot task (generating a complete survey from scratch), while one paper reframes it as a continuous maintenance problem where surveys are incrementally...} \\
\hline
3 & 33 & \textit{Multiple papers identify a fundamental trade-off between structural organization and content depth in automatic survey generation. SurveyLens finds ASG systems excel at structural outlines while Deep Research agents dominate content...} \\
\hline
4 & 23 & \textit{Several papers propose explicit hierarchical structures (e.g., knowledge trees, taxonomies, catalogues) as a foundation for survey generation, while others rely on iterative outline refinement or graph-based planning. What are the...} \\
\hline
5 & 18 & \textit{Several papers propose hierarchical or structured approaches to organizing survey content (e.g., knowledge trees, outlines, attribute trees). How do the design philosophies of these structures differ...} \\
\hline
6 & 14 & \textit{Both \texttt{<RhinoInsight>} and \texttt{<WebWeaver>} propose frameworks to improve deep research agents by introducing explicit control mechanisms over the research process. However, their architectural philosophies differ: RhinoInsight injects control modules...} \\
\hline
7 & 17 & \textit{How do the iterative and recurrent planning mechanisms in \texttt{<RecurrentGPT>}, \texttt{<IterSurvey>}, \texttt{<SurveyGen-I>}, and \texttt{<AgentCPM-Report>} differ in their handling of memory, state updates, and ...} \\
\hline
8 & 13 & \textit{How do the different approaches to structuring and retrieving literature---such as SciPIP's quintuple-based summarization and multi-granularity retrieval versus PaperRobot's knowledge graph construction and link prediction---reflect...} \\
\hline
9 & 11 & \textit{How does the positional bias identified in \texttt{<Lost in the Middle: How Language Models Use Long Contexts>} impact the effectiveness of the natural language memory states used in \texttt{<RecurrentGPT: Interactive...}} \\
\hline
10 & 15 & \textit{How do the retrieval mechanisms in \texttt{<Paper Circle>}, \texttt{<Language Models Don't Know What You Want: Evaluating Personalization in Deep Research Needs Real Users>}, and \texttt{<SURVEYAGENT>} differ in their integration of...} \\
\hline
11 & 5 & \textit{The framework relies heavily on GPT-4 for both instance creation and verification. How might this create a closed-loop system where the evaluation benchmarks are inherently limited by GPT-4's own capabilities...} \\
\hline
12 & 29 & \textit{How do the evaluation philosophies of \texttt{<SurveyLens>}, \texttt{<SGSimEval>}, \texttt{<SurGE>}, \texttt{<SurveyBench>}, and \texttt{<DeepSurvey-Bench>} differ in their definition of a `high-quality' survey...} \\
\bottomrule
\end{tabular}
\end{table*}

{\setlength{\tabcolsep}{3pt}%
\begin{longtable}{
>{\raggedright\arraybackslash\footnotesize}p{1.8cm}
>{\raggedright\arraybackslash\footnotesize}p{2.3cm}
>{\raggedright\arraybackslash\footnotesize}p{2.2cm}
>{\raggedright\arraybackslash\footnotesize}p{2.4cm}
>{\raggedright\arraybackslash\footnotesize}p{2.2cm}
>{\raggedright\arraybackslash\footnotesize}p{2.2cm}
}
\caption{Benchmarks and evaluation frameworks for survey generation and deep research.}
\label{table:comparable_table}\\
\toprule
\textbf{ID} & \textbf{Title} & \textbf{Evaluation Focus} & \textbf{Evaluation Methodology} & \textbf{Scope of Benchmark} & \textbf{Key Innovation} \\
\midrule
\endfirsthead
\multicolumn{6}{l}{\textbf{Table~\thetable\ (continued)}}\\
\toprule
\textbf{ID} & \textbf{Title} & \textbf{Evaluation Focus} & \textbf{Evaluation Methodology} & \textbf{Scope of Benchmark} & \textbf{Key Innovation} \\
\midrule
\endhead
\midrule
\multicolumn{6}{r}{\textit{Continued on next page}} \\
\endfoot
\bottomrule
\endlastfoot
\texttt{2510.17263} & TAXOALIGN: Instruction-Tuned Scholarly Taxonomy Generation & Structural alignment and semantic coherence of taxonomies & Novel metrics (average degree score, level-order traversal) and LLM-as-a-judge & Computer science (ACM Computing Surveys and conference surveys) & First benchmark for scholarly taxonomy generation (CS-TAXOBENCH) \\
\hline
\texttt{2509.18661} & Agentic AutoSurvey: Multi-Agent Framework for Automated Survey Generation & Survey quality (core, writing, depth) and citation coverage & 12-dimensional evaluation framework with LLM-as-a-judge & Computer science (AI topics) & Multi-agent framework and comprehensive 12-dimensional evaluation \\
\hline
\texttt{2602.11238} & SurveyLens: A Discipline-Aware Benchmark for Automatic Survey Generation & Discipline-aware evaluation (outline, content, reference) and canonical alignment & Discipline-aware rubric evaluation with Bradley-Terry weights and canonical alignment metrics (RAMS, TAMS) & Multi-disciplinary (10 fields) & First discipline-aware benchmark for ASG \\
\hline
\texttt{2304.03512} & Hierarchical Catalogue Generation for Literature Review: A Benchmark & Hierarchical catalogue generation (informativeness and structural similarity) & Novel metrics (CQE, CEDS) and human correlation & Computer science (arXiv) & First benchmark for hierarchical catalogue generation \\
\hline
\texttt{2510.21900} & Deep Literature Survey Automation with an Iterative Workflow & Survey quality (coverage, structure, relevance) and citation accuracy & LLM-as-a-judge and pairwise benchmark (Survey-Arena) & Computer science (arXiv) & Iterative workflow and Survey-Arena pairwise benchmark \\
\hline
\texttt{2406.10252} & AutoSurvey: Large Language Models Can Automatically Write Surveys & Content quality and citation quality & Multi-LLM-as-judge evaluation & Computer science (AI topics) & Logical parallel generation and real-time knowledge update \\
\hline
\texttt{2508.17647} & SurveyGen and QUAL-SG & Citation quality and content consistency & Automatic metrics and human evaluation & Multi-domain (4,200 surveys) & Large-scale dataset (SurveyGen) and quality-aware RAG framework \\
\hline
\texttt{2508.15804} & ReportBench & Citation quality and factual accuracy of statements & Agent-based automated verification (cited and non-cited statements) & Multi-disciplinary (10+ categories) & Benchmark for Deep Research agents with reverse prompt engineering \\
\hline
\texttt{2601.12369} & TaxoBench: Assessing Deep Research Agents on Retrieval and Taxonomy Organization & Retrieval and taxonomy organization & Hierarchy-aware metrics (US-TED, Sem-Path) and LLM-as-a-judge & LLM-related surveys (72 surveys) & Benchmark for deep research agents on retrieval and taxonomy \\
\hline
\texttt{2508.11310} & SGSimEval: A Benchmark for Automatic Survey Generation & Outline, content, and reference quality & Similarity-enhanced framework with human-authored embeddings & Multi-domain (80 surveys) & Similarity-enhanced evaluation with two weighting strategies \\
\hline
\texttt{2603.16120} & MyScholarQA: Personalized Deep Research System & Personalization in deep research (profile quality, action personalization, report adherence) & Offline synthetic users and online real-user interviews & Computer science (291 queries, 21 real users) & Personalized deep research system and real-user evaluation \\
\hline
\texttt{2305.14627} & ALCE: Enabling Large Language Models to Generate Text with Citations & Fluency, correctness, and citation quality & Automatic metrics (MAUVE, NLI-based citation recall/precision) & Multi-domain (ASQA, QAMPARI, ELI5) & First reproducible benchmark for LLM-cited text generation \\
\hline
\texttt{2508.15658} & SurGE: A Benchmark and Evaluation Framework for Scientific Survey Generation & Comprehensiveness, citation accuracy, structural quality, content quality & Multi-dimensional evaluation with human alignment & Computer science (205 surveys) & Method-agnostic benchmark with large retrieval corpus \\
\hline
\texttt{2506.12689} & SCISAGE: A Multi-Agent Framework for High-Quality Scientific Survey Generation & Content quality, structural coherence, reference accuracy & LLM-as-a-judge and human evaluation & Computer science (46 papers) & Multi-agent framework with reflective mechanism \\
\hline
\texttt{2512.02763} & SurveyEval: A Comprehensive Benchmark for Evaluating Automatically Generated Surveys & Overall quality, outline coherence, reference accuracy & LLM-as-a-judge with human references & Multi-subject (7 subjects) & Comprehensive benchmark across multiple subjects \\
\end{longtable}
}

\begingroup
\small
\setlength{\tabcolsep}{4pt}

\begin{longtable}{c p{0.85\textwidth}}
\caption{Detailed scoring rubric for \textbf{Core Quality}.}
\label{tab:dseval_rubric_core}\\
\toprule
\textbf{Score} & \textbf{Level Descriptor} \\
\midrule
\endfirsthead
\multicolumn{2}{c}{\tablename~\thetable\ -- \textit{Continued from previous page}} \\
\toprule
\textbf{Score} & \textbf{Level Descriptor} \\
\midrule
\endhead
\midrule
\multicolumn{2}{r}{\textit{Continued on next page}} \\
\endfoot
\bottomrule
\endlastfoot
\multicolumn{2}{p{0.9\textwidth}}{\textbf{Synthesis Quality} --- distinguishes true integration and conceptual grouping of literature from mere enumeration of papers.} \\
\midrule
1 & No synthesis --- the text is a disconnected list of papers with no comparison, grouping, or integrative statements. \\
2 & Very weak synthesis --- occasional grouping but mostly descriptive lists; links between works are absent or mistaken. \\
3 & Minimal synthesis --- some attempts to group papers or note similarities, but links are shallow and inconsistent. \\
4 & Basic synthesis --- groups and comparisons exist but are superficial and miss deeper patterns or tensions. \\
5 & Moderate synthesis --- reasonable grouping and some comparative statements; important cross-paper themes are noted but not developed. \\
6 & Good synthesis --- clear groups and comparisons, with emerging themes and some cross-cutting analysis. \\
7 & Strong synthesis --- integrates findings across multiple works, highlights causes of differences, and builds partial conceptual links. \\
8 & Very strong synthesis --- conveys nuanced relationships among studies, articulates patterns and contradictions, and forms clear conceptual maps. \\
9 & Excellent synthesis --- deeply integrates diverse sources, proposes coherent frameworks, and resolves or explains major discrepancies. \\
10 & Outstanding synthesis --- creates novel unifying frameworks or taxonomies, transforms raw literature into new conceptual insight. \\
\midrule
\multicolumn{2}{p{0.9\textwidth}}{\textbf{Organization} --- evaluates logical flow, section/subsection ordering, and ease of following the argument across the survey.} \\
\midrule
1 & No logical order --- sections are chaotic and the reader cannot form a coherent view of the topic. \\
2 & Very poor ordering --- frequent abrupt jumps and misplaced content; reader must constantly backtrack. \\
3 & Weak organization --- some logical chunks but many misplaced paragraphs or unclear section boundaries. \\
4 & Below-average organization --- parts make sense but transitions are rough and structure sometimes redundant. \\
5 & Adequate organization --- overall structure is serviceable though some sections could be rearranged for clarity. \\
6 & Good organization --- clear sections and logical progression with occasional weak transitions or minor redundancies. \\
7 & Strong organization --- well-ordered sections, clear flow, and purposeful subsectioning. \\
8 & Very strong organization --- sections build on each other smoothly; transitions are purposeful and guide the reader. \\
9 & Excellent organization --- highly coherent structure, logical sequencing, and efficient, non-redundant layout. \\
10 & Exemplary organization --- impeccable flow, perfect chapter/subsection design, and the structure itself aids understanding. \\
\midrule
\multicolumn{2}{p{0.9\textwidth}}{\textbf{Comprehensiveness} --- breadth of coverage across subtopics, methods, datasets, and perspectives relevant to the research area.} \\
\midrule
1 & Extremely narrow --- only a tiny slice of the topic covered; many core areas absent. \\
2 & Very limited --- major subtopics missing; coverage skewed to a few papers or approaches. \\
3 & Sparse --- several important areas omitted and coverage lacks depth. \\
4 & Partial --- some important areas covered but many relevant subtopics absent or lightly treated. \\
5 & Moderate --- key areas present but notable gaps in methods, datasets, or perspectives. \\
6 & Reasonably comprehensive --- most core areas covered though some peripheral topics missing. \\
7 & Comprehensive --- broad coverage including major methods and datasets with minor omissions. \\
8 & Very comprehensive --- covers almost all relevant facets with useful detail across areas. \\
9 & Extensive --- deep and wide coverage across subfields and perspectives; only small gaps remain. \\
10 & Exhaustive --- near-complete coverage across the domain, methods, datasets, and viewpoints. \\
\midrule
\multicolumn{2}{p{0.9\textwidth}}{\textbf{Relevance} --- measures how well the content of the survey aligns with the research topic and maintains a clear, unwavering focus throughout. It evaluates whether every section, paragraph, and sentence contributes to advancing the reader's understanding of the stated topic, and whether tangential content is either absent or minimal.} \\
\midrule
1 & Completely irrelevant --- the survey addresses a fundamentally different topic; content has no meaningful connection to the stated research area. \\
2 & Mostly irrelevant --- the survey only tangentially touches the topic, with the majority of content being off-topic, peripheral, or unrelated. \\
3 & Weakly relevant --- several sections or subsections discuss the topic, but significant portions (more than 30\%) are digressive or off-topic. \\
4 & Somewhat relevant --- the survey covers the topic but often digresses into unrelated areas; focus is inconsistent throughout. \\
5 & Moderately relevant --- the survey is generally on topic, with most content aligning with the research focus, though notable unrelated details appear in places. \\
6 & Reasonably relevant --- the survey maintains focus on the topic with occasional minor digressions; most sections contribute meaningfully to the core subject. \\
7 & Strongly relevant --- the survey is largely focused, with consistent alignment to the research topic; only rare and minor unrelated content present. \\
8 & Very strongly relevant --- the survey is tightly focused with near-perfect topical alignment; every section contributes to the core subject with only negligible peripheral content. \\
9 & Exceptionally relevant --- the survey demonstrates exemplary topical coherence; content is exclusively focused on the research topic with no discernible digressions. \\
10 & Perfectly relevant --- the survey is a model of topical unity; every single element, from introduction to conclusion, is laser-focused on advancing the reader's understanding of the research topic, with no off-topic content whatsoever. \\
\end{longtable}

\begin{longtable}{c p{0.85\textwidth}}
\caption{Detailed scoring rubric for \textbf{Writing Quality}.}
\label{tab:dseval_rubric_writing}\\
\toprule
\textbf{Score} & \textbf{Level Descriptor} \\
\midrule
\endfirsthead
\multicolumn{2}{c}{\tablename~\thetable\ -- \textit{Continued from previous page}} \\
\toprule
\textbf{Score} & \textbf{Level Descriptor} \\
\midrule
\endhead
\midrule
\multicolumn{2}{r}{\textit{Continued on next page}} \\
\endfoot
\bottomrule
\endlastfoot
\multicolumn{2}{p{0.9\textwidth}}{\textbf{Readability} --- clarity and accessibility for the intended academic audience (sentence-level clarity, pacing, and jargon handling).} \\
\midrule
1 & Unreadable --- sentences are confusing, grammar errors abound, and jargon is opaque. \\
2 & Very poor readability --- frequent grammar issues and unclear sentences; reader struggles to extract meaning. \\
3 & Low readability --- many awkward sentences and unclear phrasing; frequent re-reading required. \\
4 & Below-average readability --- understandable in places but style is verbose or clumsy; jargon often unexplained. \\
5 & Fair readability --- generally clear though some sentences or paragraphs are convoluted. \\
6 & Good readability --- clear prose overall with occasional dense passages or uneven pacing. \\
7 & Very good readability --- fluent writing, appropriate use of terminology, and easy to follow. \\
8 & Excellent readability --- crisp sentences, well-balanced pacing, and accessible to the target academic audience. \\
9 & Near-perfect readability --- highly engaging academic prose with precise phrasing and smooth flow. \\
10 & Outstanding readability --- exemplary clarity and elegance; highly accessible without loss of technical precision. \\
\midrule
\multicolumn{2}{p{0.9\textwidth}}{\textbf{Academic Rigor} --- fidelity to scholarly standards (proper citations, fair representation of prior work, and methodological transparency).} \\
\midrule
1 & No rigor --- citations missing or blatantly incorrect; claims unsupported and methods misrepresented. \\
2 & Very low rigor --- many unsupported claims, poor citation practice, and factual errors. \\
3 & Low rigor --- partial or inconsistent citation and occasional misrepresentation of methods/results. \\
4 & Below-average rigor --- citations present but incomplete and some claims lack supporting evidence. \\
5 & Moderate rigor --- generally correct citations and basic methodological descriptions but missing nuance. \\
6 & Good rigor --- solid citation practice and fair representation of methods with minor gaps. \\
7 & Strong rigor --- careful referencing, transparent presentation of methods and limitations. \\
8 & Very strong rigor --- thorough citations, balanced critique of methods, and clear evidence-backed claims. \\
9 & Excellent rigor --- meticulous referencing, rigorous methodological critique, and consistently supported claims. \\
10 & Exceptional rigor --- exemplary scholarship: exhaustive, accurate citations and deep, transparent methodological analysis. \\
\midrule
\multicolumn{2}{p{0.9\textwidth}}{\textbf{Clarity \& Coherence} --- the extent to which the survey presents technical ideas with precision, unambiguous definitions, and internally consistent terminology, claims, and conclusions across sections.} \\
\midrule
1 & Very poor --- highly vague, undefined terms, and multiple contradictions or inconsistent claims throughout. \\
2 & Poor --- frequent ambiguity and major inconsistencies that seriously hinder understanding. \\
3 & Weak --- many passages are unclear, with noticeable drifting terminology or conflicting statements. \\
4 & Below average --- some key ideas are understandable, but clarity is limited and inconsistencies remain in several places. \\
5 & Moderate --- the survey is mostly understandable, but important technical details are under-specified and coherence is uneven. \\
6 & Fairly good --- generally clear and mostly consistent, with a few imprecise explanations or minor inconsistencies. \\
7 & Good --- technical terms are usually defined well, and the narrative is coherent with only small lapses. \\
8 & Very good --- clear, precise, and internally consistent; readers can follow the logic with little effort. \\
9 & Excellent --- highly precise exposition with strong definitional consistency and near-zero contradictions. \\
10 & Outstanding --- crystal-clear, fully consistent, and immediately understandable; every section reinforces a unified narrative. \\
\end{longtable}

\begin{longtable}{c p{0.85\textwidth}}
\caption{Detailed scoring rubric for \textbf{Content Depth}.}
\label{tab:dseval_rubric_depth}\\
\toprule
\textbf{Score} & \textbf{Level Descriptor} \\
\midrule
\endfirsthead
\multicolumn{2}{c}{\tablename~\thetable\ -- \textit{Continued from previous page}} \\
\toprule
\textbf{Score} & \textbf{Level Descriptor} \\
\midrule
\endhead
\midrule
\multicolumn{2}{r}{\textit{Continued on next page}} \\
\endfoot
\bottomrule
\endlastfoot
\multicolumn{2}{p{0.9\textwidth}}{\textbf{Critical Analysis} --- depth of evaluation, fairness of critique, and the comparative assessment of strengths/weaknesses across works.} \\
\midrule
1 & No critique --- purely descriptive with no critical evaluation of methods or findings. \\
2 & Very weak critique --- token criticisms that are superficial or inaccurate. \\
3 & Shallow critique --- limited or one-sided criticisms lacking evidence or depth. \\
4 & Some critique --- occasional meaningful evaluation but inconsistent and incomplete. \\
5 & Moderate critique --- reasonable evaluations but lacking systematic comparative rigor. \\
6 & Good critique --- fair assessments, identifies key weaknesses and trade-offs in several areas. \\
7 & Strong critique --- systematic comparisons, balanced judgments, and evidence-backed critiques. \\
8 & Very strong critique --- deep analysis of methodological limitations, assumptions, and empirical gaps. \\
9 & Excellent critique --- rigorous, balanced, and insightful comparative evaluation across major works. \\
10 & Masterful critique --- transformative critique that clarifies fundamental limits, suggests remedies, and reshapes thinking. \\
\midrule
\multicolumn{2}{p{0.9\textwidth}}{\textbf{Novelty and Insights} --- the degree to which the survey produces original synthesis, useful conceptual reframing, or fresh research questions.} \\
\midrule
1 & No novelty --- restates existing content without any new perspective or insight. \\
2 & Very little novelty --- trivial observations only; no original framing or notable insight. \\
3 & Low novelty --- occasional small observations, but no meaningful new contribution. \\
4 & Limited novelty --- a few modest insights but mostly derivative. \\
5 & Some novelty --- useful observations that add modest clarity or re-organization. \\
6 & Notable novelty --- several original observations or useful reframings emerge. \\
7 & Strong novelty --- clear new insights or a helpful conceptual framing that aids understanding. \\
8 & Very strong novelty --- original synthesis that changes how parts of the field are seen. \\
9 & Excellent novelty --- significant conceptual contribution or new taxonomy with wide applicability. \\
10 & Exceptional novelty --- groundbreaking synthesis or insight that opens new directions or major re-interpretation. \\
\midrule
\multicolumn{2}{p{0.9\textwidth}}{\textbf{Specificity} --- the granularity and concreteness of the survey's analysis, including whether it goes beyond high-level summaries to discuss concrete implementation details, code-level mechanisms, module interactions, design choices, operational steps, and fine-grained distinctions between methods.} \\
\midrule
1 & Extremely vague --- entirely generic statements with no concrete mechanisms, examples, or fine-grained analysis. \\
2 & Very low specificity --- mostly broad labels or superficial lists; almost no detail beyond high-level descriptions. \\
3 & Low specificity --- a few concrete terms appear, but the survey remains mostly abstract and non-operational. \\
4 & Below-average specificity --- some partial detail is given, but analysis is still dominated by general summaries. \\
5 & Moderate specificity --- the survey includes some concrete mechanisms or examples, but many sections remain at a high level. \\
6 & Fair specificity --- several parts discuss implementation or structural details, though the analysis is not consistently fine-grained. \\
7 & Good specificity --- the survey regularly addresses concrete methods, internal mechanisms, or module-level distinctions. \\
8 & Very good specificity --- detailed analysis of implementation choices, workflow steps, architectural components, or code-level behavior in multiple places. \\
9 & Excellent specificity --- highly fine-grained discussion with precise breakdowns of mechanisms, modules, design trade-offs, and operational details. \\
10 & Outstanding specificity --- exceptionally granular analysis that closely examines implementation logic, code mechanisms, and detailed system behavior throughout the survey. \\
\midrule
\multicolumn{2}{p{0.9\textwidth}}{\textbf{Future Directions} --- the extent to which the survey provides concrete, well-justified, and actionable guidance for future research, including clearly defined problems, feasible methodologies, experimental setups, and evaluation strategies.} \\
\midrule
1 & Absent --- no meaningful future directions or only trivial/obvious statements. \\
2 & Very weak --- vague suggestions with no actionable detail or justification. \\
3 & Weak --- general directions mentioned but disconnected from identified gaps and lacking feasibility discussion. \\
4 & Limited --- some relevant directions, but still broad, under-specified, and difficult to operationalize. \\
5 & Moderate --- plausible directions linked to gaps, but limited detail on how to implement or evaluate them. \\
6 & Good --- directions are relevant and partially actionable, with some discussion of methods or evaluation. \\
7 & Strong --- clearly defined problems with actionable research paths, including hints of methodology or experimental design. \\
8 & Very strong --- specific and well-motivated directions, with concrete proposals (e.g., datasets, benchmarks, model designs) and feasibility considerations. \\
9 & Excellent --- detailed and insightful research agenda, including prioritized problems, methodological guidance. \\
10 & Outstanding --- highly actionable and deeply analyzed roadmap, offering concrete research programs with justified feasibility, step-by-step strategies, and measurable success criteria. \\
\end{longtable}

\begin{longtable}{c p{0.85\textwidth}}
\caption{Detailed scoring rubric for the \textbf{supplementary code evaluation} dimensions used in the CS-domain analysis (Section~\ref{ssec:ablation}).}
\label{tab:dseval_rubric_code}\\
\toprule
\textbf{Score} & \textbf{Level Descriptor} \\
\midrule
\endfirsthead
\multicolumn{2}{c}{\tablename~\thetable\ -- \textit{Continued from previous page}} \\
\toprule
\textbf{Score} & \textbf{Level Descriptor} \\
\midrule
\endhead
\midrule
\multicolumn{2}{r}{\textit{Continued on next page}} \\
\endfoot
\bottomrule
\endlastfoot
\multicolumn{2}{p{0.9\textwidth}}{\textbf{Implementation Fidelity} --- evaluates whether the survey goes beyond high-level descriptions from papers and captures how methods are realized in practice, including code-level mechanisms, system architecture, module interactions, algorithm implementation, and engineering design choices.} \\
\midrule
1 & No implementation understanding --- the survey only summarizes conceptual ideas from papers without discussing how methods are actually implemented. \\
2 & Very limited implementation understanding --- mentions implementation-related terms or repositories but provides no meaningful technical details. \\
3 & Weak implementation understanding --- provides superficial descriptions of frameworks, libraries, or general workflows without explaining internal mechanisms. \\
4 & Basic implementation understanding --- discusses some implementation aspects but lacks detailed analysis of architectures, modules, or design choices. \\
5 & Moderate implementation understanding --- includes several implementation details, but the analysis remains incomplete and mostly descriptive. \\
6 & Good implementation understanding --- explains important implementation components, workflows, or algorithm realization for multiple approaches. \\
7 & Strong implementation understanding --- provides detailed analysis of implementation mechanisms, module interactions, and practical design decisions. \\
8 & Very strong implementation understanding --- systematically connects paper-level ideas with code-level evidence and reveals implementation patterns across methods. \\
9 & Excellent implementation understanding --- provides deep implementation-oriented analysis, including architecture decomposition, design trade-offs, and practical limitations. \\
10 & Outstanding implementation understanding --- reconstructs how methods work in practice through comprehensive code-grounded analysis and derives insights unavailable from papers alone. \\
\midrule
\multicolumn{2}{p{0.9\textwidth}}{\textbf{Practical Reproducibility} --- evaluates whether the survey provides actionable information required to understand, reproduce, or implement reviewed methods, including code availability, dependencies, experimental configurations, execution procedures, and practical implementation challenges.} \\
\midrule
1 & No reproducibility support --- the survey only describes theoretical concepts without any practical information needed for reproduction. \\
2 & Very limited reproducibility support --- briefly mentions implementations or experiments but provides no useful reproduction details. \\
3 & Weak reproducibility support --- provides basic information such as datasets or frameworks but lacks practical guidance. \\
4 & Basic reproducibility support --- includes some experimental details but misses important requirements such as configurations, dependencies, or implementation procedures. \\
5 & Moderate reproducibility support --- provides partial reproduction information, but substantial effort is still required to reproduce the methods. \\
6 & Good reproducibility support --- covers important resources, settings, and procedures for reproducing several approaches. \\
7 & Strong reproducibility support --- provides detailed information about code resources, implementation requirements, workflows, and practical considerations. \\
8 & Very strong reproducibility support --- enables researchers to understand most reproduction requirements and identify potential implementation obstacles. \\
9 & Excellent reproducibility support --- provides comprehensive reproduction guidance, including dependencies, configurations, workflows, and limitations. \\
10 & Outstanding reproducibility support --- serves as a practical reproduction roadmap that substantially reduces the effort required to implement and validate surveyed methods. \\
\end{longtable}

\endgroup

\end{document}